\pdfoutput=1 
 \documentclass{article}

\usepackage[preprint]{neurips_data_2024}

\usepackage{subfigure}
\usepackage[utf8]{inputenc} % allow utf-8 input
\usepackage[T1]{fontenc}    % use 8-bit T1 fonts
\usepackage{url}            % simple URL typesetting
\usepackage{booktabs}       % professional-quality tables
\usepackage{amsfonts}       % blackboard math symbols
\usepackage{nicefrac}       % compact symbols for 1/2, etc.
\usepackage{microtype}      % microtypography
\usepackage{array}
\usepackage{comment}
\usepackage{bbding}
\usepackage{catchfile}

\usepackage{amsmath,amsfonts,bm}

\def\eqref#1{Eq.~(\ref{#1})}
\def\1{\bm{1}}

\DeclareMathAlphabet{\mathsfit}{\encodingdefault}{\sfdefault}{m}{sl}
\SetMathAlphabet{\mathsfit}{bold}{\encodingdefault}{\sfdefault}{bx}{n}

\usepackage[utf8]{inputenc} % allow utf-8 input
\usepackage[T1]{fontenc}    % use 8-bit T1 fonts

\usepackage{url}            % simple URL typesetting

\usepackage{booktabs}       % professional-quality tables
\newcommand{\mr}[2]{\multirow{#1}{*}{#2}}
\newcommand{\mc}[3]{\multicolumn{#1}{#2}{#3}}
\usepackage{amsfonts}       % blackboard math symbols
\usepackage{nicefrac}       % compact symbols for 1/2, etc.
\usepackage{microtype}      % microtypography
\usepackage{array}
\newcolumntype{P}[1]{>{\centering\arraybackslash}p{#1}}

\usepackage{amsmath,amsthm,amssymb,bm} % define this before the line numbering.
\usepackage{xspace}
\usepackage{enumerate}
\usepackage{enumitem}
\usepackage{stmaryrd}
\usepackage{caption}
\usepackage{subcaption}
\usepackage{multirow}
\usepackage{times}
\usepackage[dvipdfmx]{graphicx}
\usepackage{graphicx} % more modern
\usepackage{algorithm}
\usepackage{algorithmicx,algpseudocode}
\usepackage{hyperref}
\usepackage{sidecap}
\usepackage{wrapfig}
\usepackage{natbib}
\usepackage{colortbl}
\usepackage{arydshln}
\usepackage{tikz}
\usepackage{listings}
\usepackage{xcolor}
\usepackage{makecell}
\usepackage{booktabs}
\usepackage{graphicx}
\usepackage{subcaption}

\definecolor{dkred}{rgb}{0.5,0,0}
\definecolor{dkgreen}{rgb}{0,0.6,0}
\definecolor{gray}{rgb}{0.5,0.5,0.5}
\definecolor{mauve}{rgb}{0.58,0,0.82}

\numberwithin{equation}{section}
\sidecaptionvpos{figure}{t}

\newcommand{\tbl}[1]{\textsc{#1}\xspace} %relational table
\theoremstyle{plain}

\theoremstyle{definition}

\theoremstyle{remark}

\newcount\COMMENTs  % 0 suppresses notes to selves in text
\usepackage{color}
\definecolor{darkgreen}{rgb}{0,0.5,0}
\definecolor{purple}{rgb}{1,0,1}
\newcommand{\comm}[2]{\ifnum\COMMENTs=1\textcolor{#1}{#2}\fi}
\newcommand{\kexin}[1]{\comm{magenta}     {[Kexin: #1]}}

\newcommand{\xhdr}[1]{{\noindent\bfseries #1}.}

\newcommand{\hide}[1]{} %hide

\newcount\REVISEs  % 0 suppresses notes to selves in text
\usepackage{color}
\definecolor{darkgreen}{rgb}{0,0.5,0}
\definecolor{purple}{rgb}{1,0,1}
\newcommand{\commm}[2]{\ifnum\REVISEs=1\textcolor{#1}{#2}\else#2\fi}
\newcommand{\ie}{\textit{i.e.}}
\newcommand{\eg}{\textit{e.g.}}

\definecolor{customgray}{rgb}{0.3,0.3,0.3}
\definecolor{customgreen}{RGB}{140,211,89}

\newcommand{\relbench}{\textsc{RelBench}\xspace}

\newcommand{\amazon}{\texttt{rel-amazon}\xspace}
\newcommand{\userChurn}{\texttt{user-churn}\xspace}
\newcommand{\userLtv}{\texttt{user-ltv}\xspace}
\newcommand{\itemChurn}{\texttt{item-churn}\xspace}
\newcommand{\itemLtv}{\texttt{item-ltv}\xspace}
\newcommand{\userItemPurchase}{\texttt{user-item-purchase}\xspace}
\newcommand{\userItemRate}{\texttt{user-item-rate}\xspace}
\newcommand{\userItemReview}{\texttt{user-item-review}\xspace}

\newcommand{\fone}{\texttt{rel-f1}\xspace}
\newcommand{\driverPosition}{\texttt{driver-position}\xspace}

\newcommand{\driverDNF}{\texttt{driver-dnf}\xspace}
\newcommand{\driverTopThree}{\texttt{driver-top3}\xspace}

\newcommand{\handm}{\texttt{rel-hm}\xspace}
\newcommand{\itemSales}{\texttt{item-sales}\xspace}

\newcommand{\event}{\texttt{rel-event}\xspace}
\newcommand{\userAttendance}{\texttt{user-attendance}\xspace}
\newcommand{\userIgnore}{\texttt{user-ignore}\xspace}
\newcommand{\userRepeat}{\texttt{user-repeat}\xspace}

\newcommand{\avito}{\texttt{rel-avito}\xspace}
\newcommand{\userClick}{\texttt{user-clicks}\xspace}
\newcommand{\userVisit}{\texttt{user-visits}\xspace}
\newcommand{\userAdVisit}{\texttt{user-ad-visit}\xspace}
\newcommand{\adsCTR}{\texttt{ad-ctr}\xspace}

\newcommand{\stackex}{\texttt{rel-stack}\xspace}
\newcommand{\userEngage}{\texttt{user-engagement}\xspace}
\newcommand{\postVotes}{\texttt{post-votes}\xspace}
\newcommand{\userBadge}{\texttt{user-badge}\xspace}
\newcommand{\userPostComment}{\texttt{user-post-comment}\xspace}
\newcommand{\postPostLinked}{\texttt{post-post-related}\xspace}

\newcommand{\trials}{\texttt{rel-trial}\xspace}
\newcommand{\studyOutcome}{\texttt{study-outcome}\xspace}
\newcommand{\studyAdverse}{\texttt{study-adverse}\xspace}

\newcommand{\facilitySuccess}{\texttt{site-success}\xspace}
\newcommand{\sponsorConditionRec}{\texttt{condition-sponsor-run}\xspace}
\newcommand{\sponsorFacilityRec}{\texttt{site-sponsor-run}\xspace}

\newcommand{\userstudy}{\url{https://github.com/snap-stanford/relbench-user-study}\xspace}
\newcommand{\relbenchgit}{\url{https://github.com/snap-stanford/relbench}\xspace}
\newcommand{\website}{\url{https://relbench.stanford.edu}\xspace}

\definecolor{darkblue}{rgb}{0.0,0.0,0.65}
\definecolor{darkred}{rgb}{0.68,0.05,0.0}
\definecolor{darkgreen}{rgb}{0.0,0.29,0.29}
\definecolor{darkpurple}{rgb}{0.47,0.09,0.29}
\definecolor{lightgray}{rgb}{0.95, 0.95, 0.95}

\definecolor{keywordcolor}{RGB}{0,0,128}
\definecolor{commentcolor}{RGB}{0,128,0}
\definecolor{stringcolor}{RGB}{163,21,21}
\usepackage{hyperref}
\hypersetup{
   colorlinks = true,
   citecolor  = darkblue,
   linkcolor  = darkred,
   filecolor  = darkblue,
   urlcolor   = darkblue,
 }
\newcommand*{\img}[1]{%
    \raisebox{-.1\baselineskip}{%
        \includegraphics[
        height=\baselineskip,
        width=\baselineskip,
        keepaspectratio,
        ]{#1}%
    }%
}
\newcommand*\samethanks[1][\value{footnote}]{\footnotemark[#1]}
\title{\img{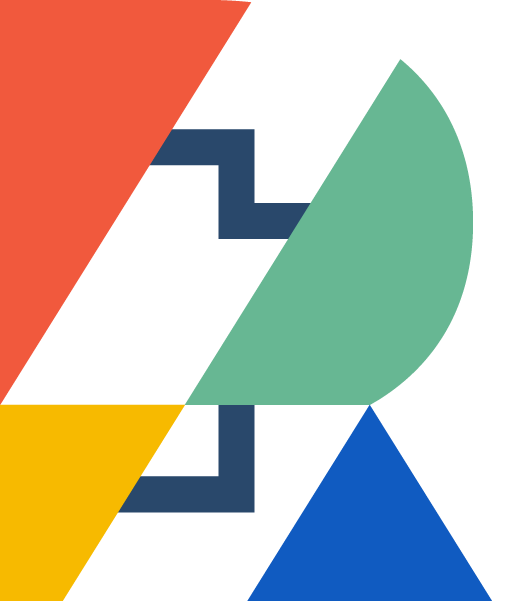}~\relbench: A Benchmark for Deep Learning \\ on Relational Databases}

\author{
Joshua Robinson$^{1}\thanks{Equal contribution, order chosen randomly. First authors may swap the ordering for professional purposes.}$, Rishabh Ranjan$^{1}\samethanks$, Weihua Hu$^{2}\samethanks$, Kexin Huang$^{1}\samethanks$, \\ \bf Jiaqi Han$^1$, Alejandro Dobles$^1$, Matthias Fey$^2$, Jan E. Lenssen$^{2,3}$, \\ \bf Yiwen Yuan$^2$, Zecheng Zhang$^2$, Xinwei He$^2$, Jure Leskovec$^{1,2}$ \\
$^1$Stanford University $^2$Kumo.AI $^3$Max Planck Institute for Informatics\\ \\ 
 \website}

\begin{document}

\maketitle

\begin{abstract}
\vspace{-10pt}
 We present \relbench, a public benchmark for solving predictive tasks over relational databases with graph neural networks. \relbench provides databases and tasks spanning diverse domains and scales, and is intended to be a foundational infrastructure for future research.
%into deep learning on relational databases. 
We use \relbench to conduct the first comprehensive study of Relational Deep Learning (RDL)~\citep{fey2023relational}, which combines graph neural network predictive models with (deep) tabular models that extract initial entity-level representations from raw tables.
%as recently proposed as Relational Deep Learning (RDL)~\citep{fey2023relational}.  
End-to-end learned RDL models fully exploit the predictive signal encoded in primary-foreign key links, marking a significant shift away from the dominant paradigm of manual feature engineering combined with tabular models. To thoroughly evaluate RDL against this prior gold-standard, we conduct an in-depth user study where an experienced data scientist manually engineers features for each task. In this study, RDL learns better models whilst reducing human work needed by more than an order of magnitude. This demonstrates the power of deep learning for solving predictive tasks over relational databases, opening up many new research opportunities enabled by \relbench. 
%\wh{Here we are focusing too much on GNNs, but IMO, we should say we implement relational deep learning (that utilizes GNNs as a core component). This applies throughout the paper.}
\end{abstract}

\vspace{-10pt}
\section{Introduction}
\label{sec:intro}
\vspace{-10pt}

Relational databases are the most widely used database management system, underpinning much of the digital economy. Their popularity stems from their table storage structure, making maintenance relatively easy, and data simple to access using powerful query languages such as SQL. Because of their popularity, AI systems across a wide variety of domains are built using data stored in relational databases, including e-commerce, social media, banking systems, healthcare, manufacturing, and open-source scientific repositories~\citep{johnson2016mimic,pubmed}.

Despite the importance of relational databases, the rich relational information is typically foregone, as no model architecture is capable of handling varied database structures. Instead, data is ``flattened'' into a simpler format such as a single table, often by manual feature engineering, on which standard tabular models can be used~\citep{kaggle-survey}. This results in a significant loss in predictive signal, and creates a need for data extraction pipelines that frequently cause bugs and add to software complexity.

\begin{figure*}[!t]
    \centering
\includegraphics[width=\textwidth]{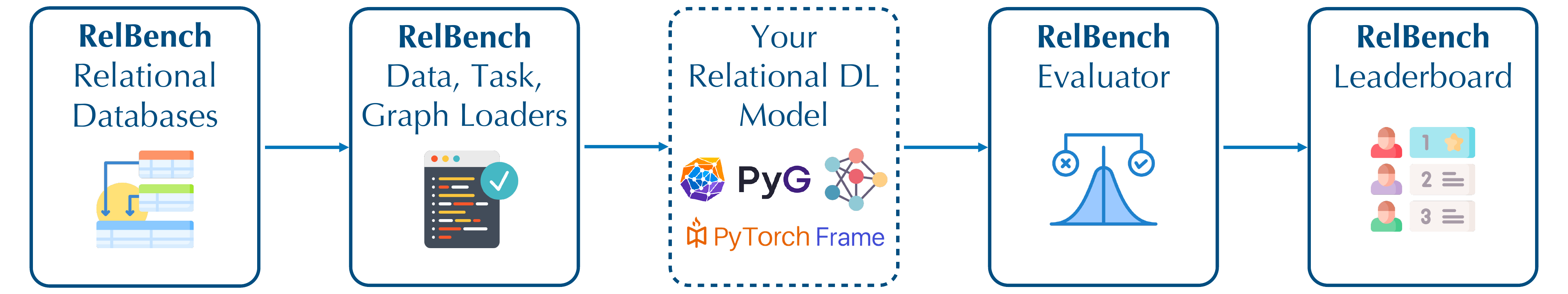}
    \caption{\textbf{\relbench} enables training and evaluation of deep learning models on relational databases. \relbench supports framework agnostic data loading, task specification, standardized data splitting, standardized evaluation metrics, and a leaderboard for tracking progress. \relbench also includes a pilot implementation of the relational deep learning blueprint of \cite{fey2023relational}.}
    \label{fig:rtb-fig}
    \vspace{-5pt}
\end{figure*}

%To fully exploit the predictive signal encoded in the relations between entities, a new proposal is to re-cast relational data as an exact \emph{graph} representation, with a node for each entity in the database, edges indicating primary-foreign key links, and node features extracted using deep tabular models. The graph representation allows Graph Neural Networks (GNNs)~\citep{gilmer2017mpgnn, hamilton2017inductive} to be used as predictive models, an approach termed \emph{Relational Deep Learning} (RDL)~\citep{fey2023relational}.
%\jure{This is inaccurate. RDL is not GNN on a DB graph. RDL is deep learning on a graph. I rewrote the paragraph below --- moved Fey citation a sentence earlier.}

To fully exploit the predictive signal encoded in the relations between entities, a new proposal is to re-cast relational data as an exact \emph{graph} representation, with a node for each entity in the database, edges indicating primary-foreign key links, and node features extracted using deep tabular models, an approach termed \emph{Relational Deep Learning} (RDL)~\citep{fey2023relational}. The graph representation allows Graph Neural Networks (GNNs)~\citep{gilmer2017mpgnn, hamilton2017inductive} to be used as predictive models. RDL is the first approach for an end-to-end learnable neural network model with access to all possible predictive signal in a relational databases, and has the potential to unlock new levels of predictive power.
However, the development of relational deep learning is limited by a complete lack of infrastructure to support research, including: (i) standardized benchmark databases and tasks to compare methods, (ii) initial implementation of RDL, including converting data to graph form and GNN training, and (iii) a pilot study of the effectiveness of relational deep learning. 

Here we present \relbench, the first benchmark for relational deep learning. \relbench is intended to be the foundational infrastructure for future research into relational deep learning, providing a comprehensive set of databases across a variety of domains, including \emph {e-commerce}, \emph{Q\&A platforms}, \emph{medical}, and \emph{sports} databases. \relbench databases span orders of magnitude in size, from 74K entities to 41M entities, and have very different time spans, between 2 weeks and 55 years of training data. They also vary significantly in their relational structure, with the total number of tables varying between 3 and 15, and total number of columns varying from 15 to 140.
Each database comes with multiple predictive tasks, 30 in total, including entity classification/regression and recommendation tasks, each chosen for their  real-world significance.
%\jure{can we say something about the variability: number of tables, total rows across all tables, total columns across all tables, number of predictive tasks. Let's reveal a bit more about the benchmark and get the reader excited!}
%several prediction tasks, including prediction of \emph{sales}, \emph{user value}, and \emph{user churn}, as well as \emph{recommendation} in various settings, and general \emph{entity behaviour} tasks, are defined and exposed as benchmarks. 

In addition to databases and tasks, we release open-source software designed to make relational deep learning widely available. This includes (i) the \relbench Python package for easy database and task loading, (ii) the first open-source implementation of relational deep learning, designed to be easily modified by researchers, and (iii) a public leaderboard for tracking progress. We comprehensively benchmark our initial RDL implementation on all \relbench tasks, comparing to various baselines. 

The most important baseline we compare to is a strong ``data scientist'' approach, for which we recruited an experienced individual to solve each task by manually engineering features and feeding them into tabular models. This approach is the current gold-standard for building predictive models on relational databases. The study, which we open source for reproducibility, finds that RDL models match or outperform the data scientist's models in accuracy, whilst reducing human hours worked by $96\%$, and lines of code by $94\%$ on average. This cons14titutes the first empirical demonstration of the central promise of RDL, and points to 
a long-awaited end-to-end deep learning solution for relational data.
%\jure{Can we be more specific here. Can we give the reader some answers: RDL vc DS: number of lines of code, accuracy, etc. Let's give some exciting numbers/results here.}

Our website\footnote{\website.} is a comprehensive entry point to RDL, describing \relbench databases and tasks, access to code on GitHub, the full relational deep learning blueprint, and tutorials for adding new databases and tasks to \relbench to allow researchers to experiment with their problems of interest.

\vspace{-5pt}
\section{Overview and Design}
\label{sec:design}
\vspace{-5pt}

%The goal of \relbench is to facilitate research on relational deep learning~\citep{fey2023relational}, an emerging field of research aiming to develop machine learning algorithms that can learn directly over relational databases without manual feature engineering.
\relbench provides a collection of diverse real-world \textbf{relational databases} along with a set of realistic \textbf{predictive tasks} associated with each database. Concretely, we provide:

\vspace{-5pt}
\begin{itemize}[leftmargin=0.6cm,itemsep=3pt, parsep=0pt]
    \item \textbf{Relational databases}, consisting of a set of tables connected via primary-foreign key relationships. Each table has columns storing diverse information about each entity. Some tables also come with \emph{time columns}, indicating the time at which the entity is created (\eg, transaction date). %For each relational database, we define \textsc{val\_timestamp} and \textsc{test\_timestamp} to be used to split data, as we describe below. 
    \item \textbf{Predictive tasks over a relational database}, which are defined by a \textbf{training table}~\citep{fey2023relational} with columns for Entity ID, seed time, and target labels.The seed time indicates \emph{at which time} the target is to be predicted, filtering future data. 
\end{itemize}

    %\footnote{It is the responsibility of machine learning algorithms to avoid time leakage during training and validation. For instance, \citet{fey2023relational} presented an elegant solution based on temporal neighbor sampling, which we adopt in our implementation.}

Next we outline key design principles of \relbench with an emphasis on data curation, data splits, research flexibility, and open-source implementation.

\xhdr{Data Curation} Relational databases are widespread, so there are many candidate predictive tasks. 
For the purpose of benchmarking we carefully curate a collection of relational databases and tasks chosen for their rich relational structure and column features. We also adopt the following principles:

%facilitate effective and efficient comparison of different algorithms, in a similar spirit to earlier graph benchmarks~\citep{hu2020open}. We adopt the following principles for selecting databases and designing tasks. 

\vspace{-0.1cm}
\begin{itemize}[leftmargin=0.6cm,itemsep=3pt, parsep=0pt]
    \item \textbf{Diverse domains:} To ensure algorithms developed on \relbench will be useful across a wide range of application domains, we select real-world relational databases from diverse domains. 
    \item \textbf{Diverse task types:} Tasks cover a wide range of real-world use-cases, including three representative task types: entity classification, entity regression, and recommendation.
\end{itemize}
\vspace{-0.1cm}

\begin{table}[t]
  \centering
  \caption{{\bf Statistics of  \relbench datasets.} Datasets vary significantly in the number of tables, total number of rows, and number of columns. In this table, we only count rows available for test inference, i.e., rows upto the test time cutoff.
  %\jure{Great! Let's also add a "total" row summing up all the tables, rows, cols, and tasks.}\kexin{done}
  }
  \label{tab:datasets_stats}
  \renewcommand{\arraystretch}{1.1}
  %\setlength{\tabcolsep}{3pt}
  % \resizebox{\linewidth}{!}{
\scriptsize
\begin{tabular}{lllrrrccc}
    \toprule
      \mr{2}{Name} & \mr{2}{Domain} &  \mr{2}{\#Tasks} & \mc{3}{c}{Tables} &  \mc{3}{c}{Timestamp (year-mon-day)} \\
       \cmidrule(lr){4-6}\cmidrule(lr){7-9}
      & &  & \#Tables & \#Rows & \#Cols & Start & Val & Test  \\
    \midrule
    \amazon & E-commerce & 7 & 3 & 15,000,713 & 15 & 2008-01-01 & 2015-10-01 & 2016-01-01 \\
    \avito & E-commerce & 4 & 8 & 20,679,117 & 42 & 2015-04-25 & 2015-05-08 & 2015-05-14 \\
    \event & Social & 3 & 5 & 41,328,337 & 128 & 1912-01-01 & 2012-11-21 & 2012-11-29\\ 
    \fone & Sports & 3 & 9 & 74,063 & 67 & 1950-05-13 & 2005-01-01 & 2010-01-01 \\
    \handm & E-commerce & 3 & 3 & 16,664,809 & 37 & 2019-09-07 & 2020-09-07 & 2020-09-14 \\
    \stackex & Social & 5 & 7 & 4,247,264 & 52 & 2009-02-02 & 2020-10-01 & 2021-01-01 \\
    \trials & Medical & 5 & 15 & 5,434,924 & 140 & 2000-01-01 & 2020-01-01 & 2021-01-01 \\
    \midrule
   \multicolumn{2}{c}{Total} & 30 & 51 &  103,466,370 & 489 & / & / & / \\
    \bottomrule
  \end{tabular}
  % }
  \vspace{-10pt}
\end{table}

\relbench databases are summarized in Table~\ref{tab:datasets_stats}, covering E-commerce, social, medical, and sports domains. The databases vary significantly in the numbers of rows (\ie, data scale) the number of columns and tables, as well as the time ranges of the databases. Tasks are summarized in Table~\ref{tab:task_stats},  each corresponding to a predictive problem of practical interest such as predicting customer churn,  predicting the number of adverse events in a clinical trial, and recommending  posts to users.

\vspace{-2pt}
\xhdr{Data Splits} Data is split temporally, with models trained on rows up to \textsc{val\_timestamp}, validated on the rows between \textsc{val\_timestamp} and \textsc{test\_timestamp}, and tested on the rows after \textsc{test\_timestamp}. 
    Our implementation carefully hides data after \textsc{test\_timestamp} during inference to systematically avoid test time data leakage~\citep{kapoor2023leakage}, and uses an elegant solution proposed by  \citet{fey2023relational} to avoid time leakage during training and validation through temporal neighbor sampling. In general, it is the designers responsibility to avoid time leakage. We recommend using our carefully tested implementation where possible.

\vspace{-2pt}
\xhdr{Research Flexibility} \relbench is designed to allow significant freedom in future research directions. For example, \relbench tasks share the same (\textsc{val\_timestamp} and \textsc{test\_timestamp}) splits across tasks within the same relational database. This opens up exciting opportunities for multi-task learning and pre-training to simultaneously improve different predictive tasks within the same relational database. We also expose the logic for converting databases into graphs. This allows future work to consider modified graph constructions, or creative uses of the raw data.

%Our open-source implementation of relational deep learning provides an initial logic for converting database into a graph, which can be modified to alter the graph design, and also provides the original set of tables (prior to graph conversion) to be used in any creative ways researchers see fit.

\vspace{-2pt}
\xhdr{Open-source RDL Implementation}
As well as datasets and tasks, we provide the first open-source implementation of relational deep learning. See Figure 2 of \citet{fey2023relational} for a high-level overview. A neural network is learned over a heterogeneous temporal graph that exactly represents the database in order to make prediction over nodes (for entity classification and regression) and links (for recommendation). Our implementation is built on top of PyTorch Frame~\citep{hu2024pytorch} for extracting initial node embeddings from raw table features, and PyTorch Geometric~\citep{fey2019fast} for GNN modeling. See Section \ref{sec:rdl_implementation} for further  details.

The rest of the paper is organized as follows. Section~\ref{sec:datasets} describes the \relbench relational databases.
Section~\ref{sec:tasks} introduces predictive tasks for each \relbench databases covering the three task types in Sections~\ref{sec:node_classification}, \ref{sec:node_regression}, and \ref{sec:link_prediction}, respectively.
Section~\ref{sec:tasks} also extensively benchmarks our RDL implementation against challenging baselines. Most importantly, we compare to a strong ``data scientist'' baseline (Section \ref{sec:ds_baseline}), finding that end-to-end RDL models outperform manual feature engineering, the current gold-standard
%We demonstrate the promising results of the relational deep learning approach compared to the baselines across tasks.
Finally, Section~\ref{sec:related} discusses related work and  Section~\ref{sec:conclusion} draws final conclusions.

\section{Relational Deep Learning Implementation}\label{sec:rdl_implementation}

As part of \relbench, we provide an initial implementation of relational deep learning, based on the blueprint of 
\cite{fey2023relational}.\footnote{Code available at: \relbenchgit.} Our implementation consists four major components: (1) heterogeneous temporal graph, (2) deep learning model, (3) temporal-aware training of the model, and (4) task-specific loss, which we briefly discuss now.

\xhdr{Heterogeneous temporal graph} 
Given a set of tables with primary-foreigh key relations between them we follow \citet{fey2023relational} to automatically construct a heterogeneous temporal graph, where each table represents a node type, each row in a table represents a node, and a primary-foreign-key relation between two table rows (nodes) represent an edge between the respective nodes. Some node types are associated with time attributes, representing the timestamp at which a node appears. The heterogeneous temporal graph is represented as a PyTorch Geometric graph object.
Each node in the heterogeneous graph comes with a rich feature derived from diverse columns of the corresponding table. We use Tensor Frame provided by PyTorch Frame~\citep{hu2024pytorch} to represent rich node features with diverse column types, \eg, numerical, categorical, timestamp, and text.

\xhdr{Deep learning model}
First, we use deep tabular models that encode raw row-level data into initial node embeddings using PyTorch Frame \citep{hu2024pytorch} (specifically, we use the ResNet tabular model~\citep{gorishniy2021revisiting}). These initial node embeddings are then fed into a GNN to iteratively update the node embeddings based on their neighbors.
For the GNN we use the heterogeneous version of the GraphSAGE model~\citep{hamilton2017inductive,fey2019fast} with sum-based neighbor aggregation. Output node embeddings are fed into task-specific prediction heads and are learned end-to-end.

\xhdr{Temporal-aware subgraph sampling}
We perform temporal neighbor sampling, which samples 
a subgraph around each entity node at a given seed time. 
Seed time is the time in history at which the prediction is made. When collecting the information to make a prediction at a given seed time, it is important for the model to only use information from before the seed time and thus not learn from the future (post the seed time). Crucially, when sampling mini-batch subgraphs we make sure that all nodes within the sampled subgraph appear before the seed time~\citep{hamilton2017inductive,fey2023relational}, which systematically avoids time leakage during training.
The sampled subgraph is fed as input to the GNN, and trained to predict the target label.

\xhdr{Task-specific prediction head and loss}
For entity-level classification, we simply apply an MLP on an entity embedding computed by our GNN to make prediction. For the loss function, we use the binary cross entropy loss for entity classification and $L_1$ loss for entity regression.

Recommendation requires computing scores between pairs of source nodes and target nodes.
For this task type, we consider two representative predictive architectures: two-tower GNN~\citep{wang2019neural} and identity-aware GNN (ID-GNN)~\citep{you2021identity}. First, the two-tower GNN computes the pairwise scores via inner product between source and target node embeddings, and the standard Bayesian Personalized Ranking loss~\citep{rendle2012bpr} is used to train the two-tower model~\citep{wang2019neural}.
Second, the ID-GNN computes the pairwise scores by applying an MLP prediction head on target entity embeddings computed by GNN for each source entity. The ID-GNN is trained by the standard binary cross entropy loss.

\vspace{-5pt}
\section{\relbench Datasets}
\vspace{-5pt}
\label{sec:datasets}

\begin{wrapfigure}[11]{r}{0.55\textwidth}
    \vspace{-25pt} % Adjust this value as needed to reduce or increase space above the figure
    \centering
\includegraphics[width=0.4\textwidth]{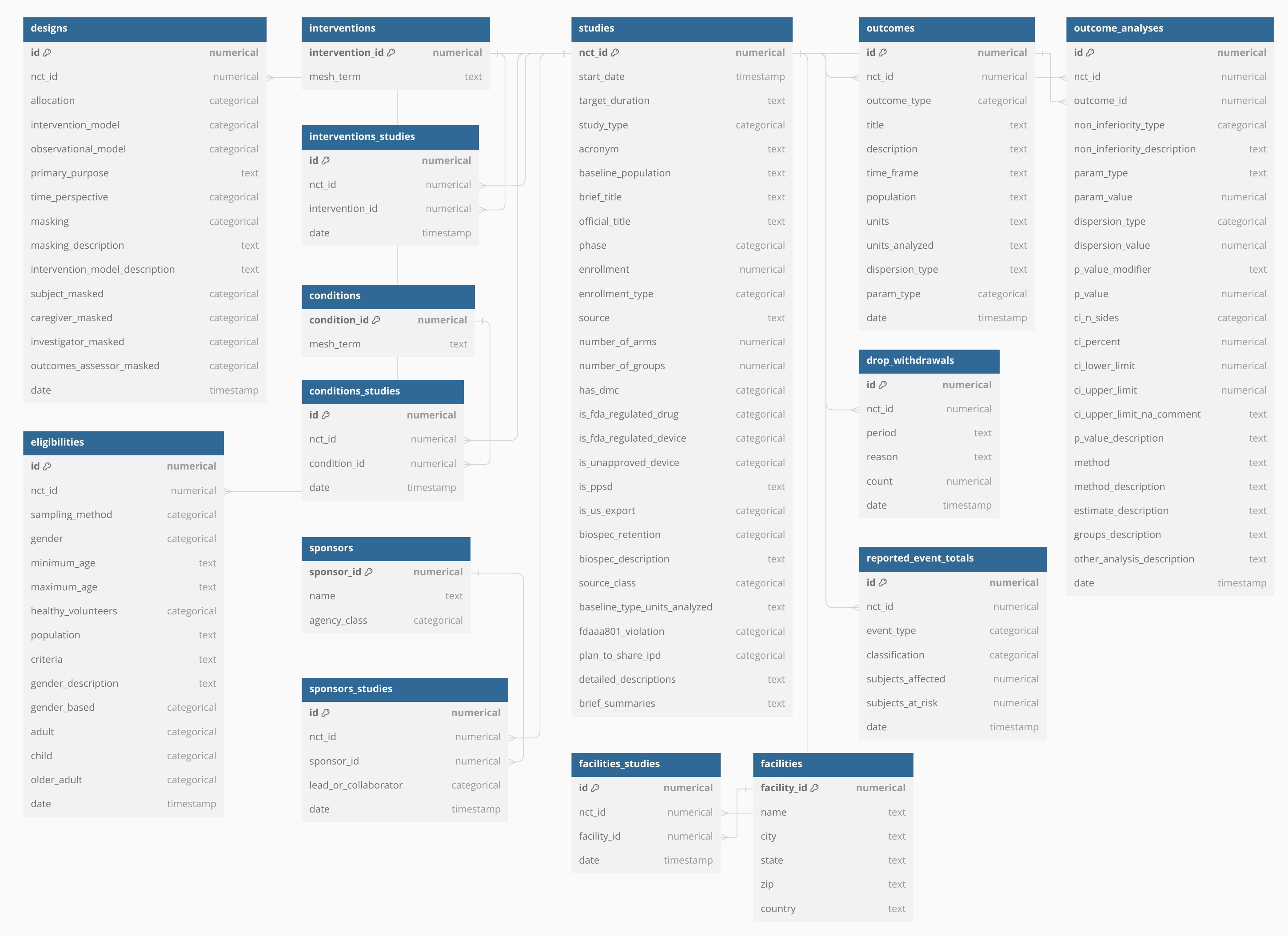}
    \vspace{-10pt} % Adjust this value as needed to reduce or increase space between the figure and the caption
    \caption{Example \relbench schema for \trials. \relbench databases have complex relational structure and rich column features. }
    \vspace{+20pt}
    \label{fig:dbdiagram}
\end{wrapfigure}

\relbench contains 7 datasets each with rich relational structure, providing a challenging environment for developing and comparing relational deep learning methods (see Figure \ref{fig:dbdiagram} for an example). The datasets are carefully processed from real-world relational databases and span diverse domains and sizes. 
Each database is associated with multiple individual predictive tasks defined in Section~\ref{sec:tasks}. 
Detailed statistics of each dataset can be found in Table~\ref{tab:datasets_stats}. We briefly describe each dataset.

\xhdr{\amazon} The Amazon E-commerce database records products, users, and reviews across Amazon's E-commerce platform. It contains rich information about products and reviews. Products include the price and category of each, reviews have the overall rating, whether the user has actually bought the product, and the text of the review itself. We use the subset of book-related products.

\xhdr{\fone} The F1 database tracks all-time Formula 1 racing data and statistics since 1950. It provides detailed information for various stakeholders including drivers, constructors, engine manufacturers, and tyre manufacturers. Highlights include data on all circuits (\eg 
 geographical details), and full historical data from every season. This includes overall standings, race results, and more specific data like practice sessions, qualifying positions, sprints, and pit stops. 

\xhdr{\stackex} Stack Exchange is a network of question-and-answer websites on different topics, where questions, answers, and users are subject to a reputation award process. The reputation system allows the sites to be self-moderating. The database includes detailed records of activity including user biographies, posts and comments (with raw text), edit histories, voting, and related posts.  In our benchmark, we use the stats-exchange site. 

\begin{table}[t]
  \centering
  \caption{{Full list of predictive tasks for each \relbench dataset (introduced in Table~\ref{tab:datasets_stats}).}  }
  \label{tab:task_stats}
  \renewcommand{\arraystretch}{1.1}
  \setlength{\tabcolsep}{4pt}
  % \resizebox{\linewidth}{!}{
  \tiny
  \begin{tabular}{lllccclll}
    \toprule
      \mr{2}{\textbf{Dataset}} & \mr{2}{\textbf{Task name}} & \mr{2}{\textbf{Task type}} & \mc{3}{c}{\textbf{\#Rows of training table}} & \#Unique & \%train/test & \#Dst \\
      & & & Train & Validation & Test & Entities & Entity Overlap  & Entities \\
    \midrule
      \mr{6}{\amazon}
      & \userChurn & entity-cls & 4,732,555 & 409,792 & 351,885 & 1,585,983 & 88.0 & --- \\
      & \itemChurn & entity-cls & 2,559,264 & 177,689 & 166,842 & 416,352 & 93.1 & ---  \\
      & \userLtv & entity-reg  & 4,732,555 & 409,792 & 351,885 & 1,585,983 & 88.0 & --- \\
      & \itemLtv & entity-reg   & 2,707,679 & 166,978 & 178,334 & 427,537 & 93.5 & ---\\
      & \userItemPurchase & recommendation   & 5,112,803 & 351,876 & 393,985 & 1,632,909 & 87.4 & 12,562,384\\
      & \userItemRate & recommendation  & 3,667,157 & 257,939 & 292,609 & 1,481,360 & 81.0 & 7,665,611\\
      & \userItemReview & recommendation  & 2,324,177 & 116,970 & 127,021 & 894,136 & 74.1 & 5,406,835\\
    \midrule
    \mr{4}{\avito}
    & \adsCTR & entity-reg & 5,100 & 1,766 & 1,816 & 4,997 & 59.8 &---\\
    & \userClick & entity-cls & 59,454 & 21,183 & 47,996 & 66,449 & 45.3 &---\\
    & \userVisit & entity-cls & 86,619 & 29,979 & 36,129 & 63,405 & 64.6 &---\\
    & \userAdVisit & recommendation & 86,616 & 29,979 & 36,129 & 63,402 & 64.6 & 3,616,174\\
    \midrule
    \mr{3}{\event}
    & \userAttendance & entity-reg & 19,261 & 2,014 & 2,006 &9,694 & 14.6&---\\
    & \userRepeat & entity-cls & 3,842 & 268 & 246 &1,514 & 11.5&---\\
    & \userIgnore & entity-cls & 19,239 & 4,185 & 4,010 &9,799 & 21.1&---\\
    \midrule
    \mr{3}{\fone}
      & \driverDNF & entity-cls & 11,411 & 566 & 702 & 821 & 50.0 & --- \\
      & \driverTopThree & entity-cls  & 1,353 & 588 & 726 & 134 & 50.0 & --- \\
      & \driverPosition & entity-reg & 7,453 & 499 & 760 & 826 & 44.6 & --- \\
      % & \driverConstructorResult & recommendation & 893 & 73 & 57 & 728 & 5.3 & 2,147 \\
    \midrule
    \mr{3}{\handm}
      & \userChurn & entity-cls  & 3,871,410 & 76,556 & 74,575 & 1,002,984 & 89.7 & --- \\
      & \itemSales & entity-reg & 5,488,184 & 105,542 & 105,542 & 105,542 & 100.0 & --- \\
      & \userItemPurchase & recommendation & 3,878,451 & 74,575 & 67,144 & 1,004,046 & 89.2 & 13,428,473 \\
    \midrule
    \mr{5}{\stackex}
      & \userEngage & entity-cls & 1,360,850 & 85,838 & 88,137 & 88,137 & 97.4 & --- \\
      & \userBadge & entity-cls  & 3,386,276 & 247,398 & 255,360 & 255,360 & 96.9 & --- \\
      & \postVotes & entity-reg  & 2,453,921 & 156,216 & 160,903 & 160,903 & 97.1 & --- \\
      & \userPostComment & recommendation  & 21,239 & 825 & 758 & 11,453 & 59.9 & 44,940 \\
      & \postPostLinked & recommendation  & 5,855 & 226 & 258 & 5,924 & 8.5 & 7,456 \\
    \midrule
    \mr{5}{\trials}
      & \studyOutcome & entity-cls  & 11,994 & 960 & 825 & 13,779 & 0.0 & --- \\
      & \studyAdverse & entity-reg  & 43,335 & 3,596 & 3,098 & 50,029 & 0.0 & --- \\
      & \facilitySuccess & entity-reg  & 151,407 & 19,740 & 22,617 & 129,542 & 42.0 & --- \\
      & \sponsorConditionRec & recommendation  & 36,934 & 2,081 & 2,057 & 3,956 & 98.4 & 533,624 \\
      & \sponsorFacilityRec & recommendation  & 669,310 & 37,003 & 27,428 & 445,513 & 48.3 & 1,565,463 \\
   \bottomrule
  \end{tabular}
  % }
\end{table}

\xhdr{\trials} The clinical trial database is curated from AACT initiative, which consolidates all protocol and results data from studies registered on ClinicalTrials.gov. It offers extensive information about clinical trials, including study designs, participant demographics, intervention details, and outcomes. It is an important resource for health research, policy making, and therapeutic development.

\xhdr{\handm} The H\&M relational database hosts extensive customer and product data for online shopping experiences across its extensive network of brands and stores.  This database includes detailed customer purchase histories and a rich set of metadata, encompassing everything from basic demographic information to extensive details about each product available. 

\xhdr{\event} The Event Recommendation database is obtained from user data on a mobile app called Hangtime. This app allows users to keep track of their friends' social plans. The database contains data on user actions, event metadata, and demographic information, as well as users' social relations, which captures how social relations can affect user behavior. Data is fully anonymized, with no personally identifiable information (such as names or aliases) available.

\xhdr{\avito} Avito is a leading online advertisement platform, providing a marketplace for users to buy and sell a wide variety of products and services, including real estate, vehicles, jobs, and goods. The Avito Context Ad Clicks dataset on Kaggle is part of a competition aimed at predicting whether an ad will be clicked based on contextual information. This dataset includes user searches, ad attributes, and other related data to help build predictive models.

\xhdr{Data Provenance} All data is sourced from publicly available repositories with licenses permitting usage for research purposes. See Appendix \ref{app:data} for details of data sources, licenses, and more.

\section{Predictive Tasks on \relbench Datasets}
\label{sec:tasks}
% Table
% Dataset-task table\\
% row: dataset\\
% col: task type\\
% entry: concrete tasks

\relbench introduces 30 new predictive tasks defined over the databases introduced in Section~\ref{sec:datasets}.
A full list of tasks is given in Table~\ref{tab:task_stats}, with high-level descriptions given in Appendix \ref{app: task info} (and our \href{https://relbench.stanford.edu/}{website}) due to space limitations.
Tasks are grouped into three task types: entity classification (Section~\ref{sec:node_classification}), entity regression (Section~\ref{sec:node_regression}), and entity link prediction (Section~\ref{sec:link_prediction}). Tasks differ significantly in the number of train/val/test entities, number of unique entities (the same entity may appear multiple times at different timestamps), and the proportion of test entities seen during training. Note this is not data leakage, since entity predictions are timestamp dependent, and can change over time. Tasks with no overlap are pure inductive tasks, whilst other tasks are (partially) transductive. 

%\wh{I am wondering why don't we put data scientist baseline in Tables 3, 4, and 5? Otherwise, I would feel all these baselines are too weak. For reviewers, the only ``interesting'' comparison is data scientist (remember one ICML reviewer that asks for a stronger baseline).}

%For each task type, we describe the target label, evaluation metric, and baseline methods with the notable inclusion of a strong data scientist baseline discussed in detail in Section~\ref{sec:ds_baseline}. We also report results comparing our relational deep learning implementation (Appendix~\ref{sec:rdl_implementation}) against these baselines, finding competitive performance.
%and several task-type-specific basic baselines (Sections~\ref{sec:node_classification},~\ref{sec:node_regression}, and~\ref{sec:link_prediction}).

\subsection{Entity Classification}
\label{sec:node_classification}

% \begin{wraptable}[20]{r}{0.5\textwidth}
\begin{table}
    \caption{Entity classification results (AUROC, higher is better) on \relbench. Best values are in bold. See Table~\ref{tab:app_classif} in Appendix~\ref{app:gnn_ablations} for standard deviations.
    % \jure{See Average row (FIX IT)!}
    }
    \centering
    % \setlength{\tabcolsep}{3pt}
    % \scriptsize
    \tiny
    \newcommand{\tabledata}{\multirow[c]{4}{*}{\texttt{rel-amazon}} & \multirow[c]{2}{*}{\texttt{user-churn}} & Val & $52.05$ & $\bm{70.45}$ & $ 35.35$ \% \\
 &  & Test & $52.22$ & $\bm{70.42}$ & $ 34.86$ \% \\
\cmidrule{2-6}
 & \multirow[c]{2}{*}{\texttt{item-churn}} & Val & $62.39$ & $\bm{82.39}$ & $ 32.06$ \% \\
 &  & Test & $62.54$ & $\bm{82.81}$ & $ 32.40$ \% \\
\cmidrule{1-6}
\multirow[c]{4}{*}{\texttt{rel-avito}} & \multirow[c]{2}{*}{\texttt{user-visits}} & Val & $53.31$ & $\bm{69.65}$ & $ 30.66$ \% \\
 &  & Test & $53.05$ & $\bm{66.20}$ & $ 24.78$ \% \\
\cmidrule{2-6}
 & \multirow[c]{2}{*}{\texttt{user-clicks}} & Val & $55.63$ & $\bm{64.73}$ & $ 16.35$ \% \\
 &  & Test & $53.60$ & $\bm{65.90}$ & $ 22.96$ \% \\
\cmidrule{1-6}
\multirow[c]{4}{*}{\texttt{rel-event}} & \multirow[c]{2}{*}{\texttt{user-repeat}} & Val & $67.76$ & $\bm{71.25}$ & $ 5.15$ \% \\
 &  & Test & $68.04$ & $\bm{76.89}$ & $ 13.02$ \% \\
\cmidrule{2-6}
 & \multirow[c]{2}{*}{\texttt{user-ignore}} & Val & $87.96$ & $\bm{91.70}$ & $ 4.25$ \% \\
 &  & Test & $79.93$ & $\bm{81.62}$ & $ 2.12$ \% \\
\cmidrule{1-6}
\multirow[c]{4}{*}{\texttt{rel-f1}} & \multirow[c]{2}{*}{\texttt{driver-dnf}} & Val & $68.42$ & $\bm{71.36}$ & $ 4.31$ \% \\
 &  & Test & $68.56$ & $\bm{72.62}$ & $ 5.93$ \% \\
\cmidrule{2-6}
 & \multirow[c]{2}{*}{\texttt{driver-top3}} & Val & $67.76$ & $\bm{77.64}$ & $ 14.57$ \% \\
 &  & Test & $73.92$ & $\bm{75.54}$ & $ 2.20$ \% \\
\cmidrule{1-6}
\multirow[c]{2}{*}{\texttt{rel-hm}} & \multirow[c]{2}{*}{\texttt{user-churn}} & Val & $56.05$ & $\bm{70.42}$ & $ 25.63$ \% \\
 &  & Test & $55.21$ & $\bm{69.88}$ & $ 26.59$ \% \\
\cmidrule{1-6}
\multirow[c]{4}{*}{\texttt{rel-stack}} & \multirow[c]{2}{*}{\texttt{user-engagement}} & Val & $65.12$ & $\bm{90.21}$ & $ 38.53$ \% \\
 &  & Test & $63.39$ & $\bm{90.59}$ & $ 42.91$ \% \\
\cmidrule{2-6}
 & \multirow[c]{2}{*}{\texttt{user-badge}} & Val & $65.39$ & $\bm{89.86}$ & $ 37.43$ \% \\
 &  & Test & $63.43$ & $\bm{88.86}$ & $ 40.08$ \% \\
\cmidrule{1-6}
\multirow[c]{2}{*}{\texttt{rel-trial}} & \multirow[c]{2}{*}{\texttt{study-outcome}} & Val & $\bm{68.30}$ & $68.18$ & $-0.19$ \% \\
 &  & Test & $\bm{70.09}$ & $68.60$ & $-2.13$ \% \\
\cmidrule{1-6}
\multicolumn{2}{c}{\multirow[c]{2}{*}{Average}} & Val & $64.18$ & $\bm{76.49}$ & $20.34$ \% \\
 &  & Test & $63.66$ & $\bm{75.83}$ & $20.48$ \% \\}
    \begin{tabular}{lllrrr}
    \toprule
    \textbf{Dataset} & \textbf{Task} & \textbf{Split} & \textbf{LightGBM} & \textbf{RDL} & \makecell{\textbf{Rel. Gain}\\\textbf{of RDL}} \\
    \midrule
    \multirow[c]{9}{*}{\amazon} & \multirow[c]{3}{*}{\userItemPurchase} 
 & Train & 11,759,844 & 2.18 & - \\
 &  & Val & 802,540 & 2.28 & 0.18 \\
 &  & Test & 918,919 & 2.33 & 0.15 \\

\cmidrule{2-6}

 & \multirow[c]{3}{*}{\userItemRate} 
 & Train & 7,146,115 & 1.8 & - \\
 &  & Val & 519,496 & 2.01 & 0.19 \\
 &  & Test & 599,867 & 2.05 & 0.15 \\

 \cmidrule{2-6}

 & \multirow[c]{3}{*}{\userItemReview} 
 & Train & 5,138,184 & 2.19 & - \\
 &  & Val & 268,651 & 2.3 & 0.18 \\
 &  & Test & 305,476 & 2.4 & 0.15 \\

\cmidrule{1-6}

\multirow[c]{3}{*}{\handm} & \multirow[c]{3}{*}{\userItemPurchase} & 
Train & 13,191,321 & 3.38 & - \\
 &  & Val & 237,152 & 3.18 & 3.51 \\
 &  & Test & 207,996 & 3.10 & 3.76 \\

% \cmidrule{1-6}
% \multirow[c]{3}{*}{\fone} & \multirow[c]{3}{*}{\driverConstructorResult}
% & Train & 1900 &  & - \\
%  &  & Val & 142 & 1.95 & 9.86 \\
%  &  & Test & 105 & 1.84 & 10.48 \\

\cmidrule{1-6}

\multirow[c]{6}{*}{\stackex} & \multirow[c]{3}{*}{\userPostComment}
 & Train & 43,337 & 2.08 & - \\
 & & Val & 1,603 & 1.94 & 3.43 \\
 &  & Test & 1,517 & 2.0 & 4.09 \\

\cmidrule{2-6}

 & \multirow[c]{3}{*}{\postPostLinked}
 & Train & 7,162 & 1.2 & - \\
 &  & Val & 294 & 1.3 & 0.0 \\
 &  & Test & 359 & 1.39 & 1.39 \\

\cmidrule{1-6}

\multirow[c]{6}{*}{\trials} & \multirow[c]{3}{*}{\sponsorConditionRec}
 & Train & 503,176 & 12.51 & - \\
 &  & Val & 30,448 & 14.63 & 34.48 \\
 &  & Test & 25,694 & 12.49 & 38.37 \\

\cmidrule{2-6}

& \multirow[c]{3}{*}{\sponsorFacilityRec}
 & Train & 1,485,360 & 2.27 & - \\
 &  & Val & 80,103 & 2.16 & 20.91 \\
 &  & Test & 50,635 & 1.85 & 23.29 \\

\cmidrule{1-6}

\multirow[c]{3}{*}{\avito} & \multirow[c]{3}{*}{\userAdVisit} & 
Train & 2,738,733 & 31.53 & - \\
 &  & Val & 877,441 & 29.27 & 6.79 \\
 &  & Test & 712,985 & 19.73 & 4.73 \\
    \bottomrule
    \end{tabular}
    \label{tab:classif results}
\end{table}
% \end{wraptable}

The first task type is entity-level classification. The task is to predict binary labels of a given entity at a given seed time. We use the ROC-AUC~\citep{hanley1983method} metric for evaluation (higher is better). We compare to a LightGBM classifier baseline over the raw entity table features. Note that here only information from the single entity table is used.

\paragraph{Experimental results.} Results are given in Table \ref{tab:classif results}, with RDL outperforming or matching baselines in all cases. Notably, LightGBM achieves similar performance to RDL on the \studyOutcome task from \trials. This task has extremely rich features in the target table (28 columns total), giving the LightGBM many potentially useful features even without feature engineering. It is an interesting research question how to design RDL models better able to extract these features and unify them with cross-table information in order to outperform the LightGBM model on this dataset.

\subsection{Entity Regression}
\label{sec:node_regression}

Entity-level regression tasks involve predicting numerical labels of an entity at a given seed time. We use Mean Absolute Error (MAE) as our metric (lower is better). We consider the following baselines:
\begin{itemize}[leftmargin=0.6cm,itemsep=3pt, parsep=0pt]
    \item \textbf{Entity mean/median} calculates the mean/median label value for each entity in training data and predicts the mean/median value for the entity.
    \item \textbf{Global mean/median} calculates the global mean/median label value over the training data and predicts the same mean/median value across all entities.
    \item \textbf{Global zero} predicts zero for all entities.
    \item \textbf{LightGBM} learns a LightGBM~\citep{ke2017lightgbm} regressor over the raw entity features to predict the numerical targets. Note that only information from the single entity table is used.
\end{itemize}

\xhdr{Experimental results} Results in Table \ref{tab: regression} show our RDL implementation outperforms or matches baselines in all cases. A number of tasks, such as \driverPosition and \studyAdverse, have matching performance up to statistical significance, suggesting some room for improvement. We analyze this further in Appendix \ref{app: user study}, identifying one potential cause, suggesting an opportunity for improved performance for regression tasks.

\begin{table}
\caption{Entity regression results (MAE, lower is better) on \relbench. Best values are in bold. See Table~\ref{tab:app_regression} in Appendix~\ref{app:gnn_ablations} for standard deviations.
% \jure{See Average row (FIX IT)!}
}
\centering
\setlength{\tabcolsep}{5pt}
% \scriptsize
\tiny
\newcommand{\tabledata}{\multirow[c]{4}{*}{\texttt{rel-amazon}} & \multirow[c]{2}{*}{\texttt{user-ltv}} & Val & $14.141$ & $20.740$ & $14.141$ & $17.685$ & $15.978$ & $14.141$ & $\bm{12.132}$ & $ 14.21$ \% \\
 &  & Test & $16.783$ & $22.121$ & $16.783$ & $19.055$ & $17.423$ & $16.783$ & $\bm{14.313}$ & $ 14.72$ \% \\
\cmidrule{2-11}
 & \multirow[c]{2}{*}{\texttt{item-ltv}} & Val & $72.096$ & $78.110$ & $59.471$ & $80.466$ & $68.922$ & $55.741$ & $\bm{45.140}$ & $ 19.02$ \% \\
 &  & Test & $77.126$ & $81.852$ & $64.234$ & $78.423$ & $66.436$ & $60.569$ & $\bm{50.053}$ & $ 17.36$ \% \\
\cmidrule{1-11}
\multirow[c]{2}{*}{\texttt{rel-avito}} & \multirow[c]{2}{*}{\texttt{ad-ctr}} & Val & $0.048$ & $0.048$ & $0.040$ & $0.044$ & $0.044$ & $0.037$ & $\bm{0.037}$ & $ 2.21$ \% \\
 &  & Test & $0.052$ & $0.051$ & $0.043$ & $0.046$ & $0.046$ & $\bm{0.041}$ & $0.041$ & $-0.18$ \% \\
\cmidrule{1-11}
\multirow[c]{2}{*}{\texttt{rel-event}} & \multirow[c]{2}{*}{\texttt{user-attendance}} & Val & $0.262$ & $0.457$ & $0.262$ & $0.296$ & $0.268$ & $0.262$ & $\bm{0.255}$ & $ 2.65$ \% \\
 &  & Test & $0.264$ & $0.470$ & $0.264$ & $0.304$ & $0.269$ & $0.264$ & $\bm{0.258}$ & $ 1.97$ \% \\
\cmidrule{1-11}
\multirow[c]{2}{*}{\texttt{rel-f1}} & \multirow[c]{2}{*}{\texttt{driver-position}} & Val & $11.083$ & $4.334$ & $4.136$ & $7.181$ & $7.114$ & $3.450$ & $\bm{3.193}$ & $ 7.44$ \% \\
 &  & Test & $11.926$ & $4.513$ & $4.399$ & $8.501$ & $8.519$ & $4.170$ & $\bm{4.022}$ & $ 3.56$ \% \\
\cmidrule{1-11}
\multirow[c]{2}{*}{\texttt{rel-hm}} & \multirow[c]{2}{*}{\texttt{item-sales}} & Val & $0.086$ & $0.142$ & $0.086$ & $0.117$ & $0.086$ & $0.086$ & $\bm{0.065}$ & $ 24.50$ \% \\
 &  & Test & $0.076$ & $0.134$ & $0.076$ & $0.111$ & $0.078$ & $0.076$ & $\bm{0.056}$ & $ 26.90$ \% \\
\cmidrule{1-11}
\multirow[c]{2}{*}{\texttt{rel-stack}} & \multirow[c]{2}{*}{\texttt{post-votes}} & Val & $0.062$ & $0.146$ & $0.062$ & $0.102$ & $0.064$ & $0.062$ & $\bm{0.059}$ & $ 4.19$ \% \\
 &  & Test & $0.068$ & $0.149$ & $0.068$ & $0.106$ & $0.069$ & $0.068$ & $\bm{0.065}$ & $ 4.11$ \% \\
\cmidrule{1-11}
\multirow[c]{4}{*}{\texttt{rel-trial}} & \multirow[c]{2}{*}{\texttt{study-adverse}} & Val & $57.083$ & $75.008$ & $56.786$ & $57.083$ & $57.083$ & $\bm{45.774}$ & $46.290$ & $-1.13$ \% \\
 &  & Test & $57.930$ & $73.781$ & $57.533$ & $57.930$ & $57.930$ & $\bm{44.011}$ & $44.473$ & $-1.05$ \% \\
\cmidrule{2-11}
 & \multirow[c]{2}{*}{\texttt{site-success}} & Val & $0.475$ & $0.462$ & $0.475$ & $0.447$ & $0.450$ & $0.417$ & $\bm{0.401}$ & $ 3.87$ \% \\
 &  & Test & $0.462$ & $0.468$ & $0.462$ & $0.448$ & $0.441$ & $0.425$ & $\bm{0.400}$ & $ 5.86$ \% \\
\cmidrule{1-11}
\multicolumn{2}{c}{\multirow[c]{2}{*}{Average}} & Val & $17.260$ & $19.939$ & $15.051$ & $18.158$ & $16.668$ & $13.330$ & $\bm{11.952}$ & $8.55$ \% \\
 &  & Test & $18.299$ & $20.393$ & $15.985$ & $18.325$ & $16.801$ & $14.045$ & $\bm{12.631}$ & $8.14$ \% \\}
\begin{tabular}{lllrrrrrrrr}
\toprule
\textbf{Dataset} & \textbf{Task} & \textbf{Split} & \makecell{\textbf{Global}\\\textbf{Zero}} & \makecell{\textbf{Global}\\\textbf{Mean}} & \makecell{\textbf{Global}\\\textbf{Median}} & \makecell{\textbf{Entity}\\\textbf{Mean}} & \makecell{\textbf{Entity}\\\textbf{Median}} &
\textbf{LightGBM} &
\textbf{RDL} & 
\makecell{\textbf{Rel. Gain}\\\textbf{of RDL}}\\
\midrule
\multirow[c]{9}{*}{\amazon} & \multirow[c]{3}{*}{\userItemPurchase} 
 & Train & 11,759,844 & 2.18 & - \\
 &  & Val & 802,540 & 2.28 & 0.18 \\
 &  & Test & 918,919 & 2.33 & 0.15 \\

\cmidrule{2-6}

 & \multirow[c]{3}{*}{\userItemRate} 
 & Train & 7,146,115 & 1.8 & - \\
 &  & Val & 519,496 & 2.01 & 0.19 \\
 &  & Test & 599,867 & 2.05 & 0.15 \\

 \cmidrule{2-6}

 & \multirow[c]{3}{*}{\userItemReview} 
 & Train & 5,138,184 & 2.19 & - \\
 &  & Val & 268,651 & 2.3 & 0.18 \\
 &  & Test & 305,476 & 2.4 & 0.15 \\

\cmidrule{1-6}

\multirow[c]{3}{*}{\handm} & \multirow[c]{3}{*}{\userItemPurchase} & 
Train & 13,191,321 & 3.38 & - \\
 &  & Val & 237,152 & 3.18 & 3.51 \\
 &  & Test & 207,996 & 3.10 & 3.76 \\

% \cmidrule{1-6}
% \multirow[c]{3}{*}{\fone} & \multirow[c]{3}{*}{\driverConstructorResult}
% & Train & 1900 &  & - \\
%  &  & Val & 142 & 1.95 & 9.86 \\
%  &  & Test & 105 & 1.84 & 10.48 \\

\cmidrule{1-6}

\multirow[c]{6}{*}{\stackex} & \multirow[c]{3}{*}{\userPostComment}
 & Train & 43,337 & 2.08 & - \\
 & & Val & 1,603 & 1.94 & 3.43 \\
 &  & Test & 1,517 & 2.0 & 4.09 \\

\cmidrule{2-6}

 & \multirow[c]{3}{*}{\postPostLinked}
 & Train & 7,162 & 1.2 & - \\
 &  & Val & 294 & 1.3 & 0.0 \\
 &  & Test & 359 & 1.39 & 1.39 \\

\cmidrule{1-6}

\multirow[c]{6}{*}{\trials} & \multirow[c]{3}{*}{\sponsorConditionRec}
 & Train & 503,176 & 12.51 & - \\
 &  & Val & 30,448 & 14.63 & 34.48 \\
 &  & Test & 25,694 & 12.49 & 38.37 \\

\cmidrule{2-6}

& \multirow[c]{3}{*}{\sponsorFacilityRec}
 & Train & 1,485,360 & 2.27 & - \\
 &  & Val & 80,103 & 2.16 & 20.91 \\
 &  & Test & 50,635 & 1.85 & 23.29 \\

\cmidrule{1-6}

\multirow[c]{3}{*}{\avito} & \multirow[c]{3}{*}{\userAdVisit} & 
Train & 2,738,733 & 31.53 & - \\
 &  & Val & 877,441 & 29.27 & 6.79 \\
 &  & Test & 712,985 & 19.73 & 4.73 \\
% manually entered
\bottomrule
\end{tabular}
\label{tab: regression}
\end{table}

\subsection{Recommendation}\label{sec:link_prediction}

Finally, we also introduce recommendation tasks on pairs of entities. The task is to predict a list of top $K$ target entities given a source entity at a given seed time.
The metric we use is Mean Average Precision (MAP) @$K$, where $K$ is set per task (higher is better). We consider the following baselines:
\begin{itemize}[leftmargin=0.6cm,itemsep=3pt, parsep=0pt]
    \item \textbf{Global popularity} computes the top $K$ most popular target entities (by count) across the entire training table and predict the $K$ globally popular target entities across all source entities.
    \item \textbf{Past visit} computes the top $K$ most visited target entities for each source entity within the training table and predict those past-visited target entities for each entity.
    \item \textbf{LightGBM} learns a LightGBM~\citep{ke2017lightgbm} classifier over the raw features of the source and target entities (concatenated) to predict the link. Additionally, global popularity and past visit ranks are also provided as inputs.
\end{itemize}

% Task table\\
% row: tasks\\
% col: stats (number of time frames, timedelta, number of rows, number of entities involved, number of train/val/test data)

For recommendation, it is also important to ensure a certain density of links in the training data in order for there to be sufficient predictive signal. In Appendix \ref{app: task info} we report statistics on the average number of destination entities each source entity links to. For most tasks the density is $\geq 1$, with the exception of \stackex which is more sparse, but is included to test in extreme sparse settings.

\xhdr{Experimental results} Results are given in Table \ref{tab:link results}. We find that either the RDL implementation using GraphSAGE \citep{hamilton2017inductive}, or ID-GNN \citep{you2021identity} as the GNN component performs best, often by a very significant margin. ID-GNN excels in cases were predictions are entity-specific (\ie, Past Visit baseline outperforms Global Popularity), whilst the plain GNN excels in the reverse case. This reflects the inductive biases of each model, with GraphSAGE being able to learn structural features, and ID-GNN able to take into account the specific node ID.
%RDL underperforms in two trial recommendation tasks, potentially because these tasks require historical information which current implementations may fail to capture, suggesting an exciting future RDL research direction.

% Results table\\
% row: methods\\
% cols: tasks (or swap rows/cols)

\begin{table}
\caption{Recommendation results (MAP, higher is better) on \relbench. Best values are in bold. See Table~\ref{tab:app_link} in Appendix~\ref{app:gnn_ablations} for standard deviations.
% \jure{See Average Row. Fix it!}
% \jure{to all tables, can be get avg. relative improvement/difference between RDL and DT?}
}
\centering
% \setlength{\tabcolsep}{4pt}
% \scriptsize
\tiny
\newcommand{\tabledata}{\multirow[c]{6}{*}{\texttt{rel-amazon}} & \multirow[c]{2}{*}{\texttt{user-item-purchase}} & Val & $0.31$ & $0.07$ & $0.18$ & $\bm{1.53}$ & $0.13$ & $ 397.55$ \% \\
 &  & Test & $0.24$ & $0.06$ & $0.16$ & $\bm{0.74}$ & $0.10$ & $ 204.74$ \% \\
\cmidrule{2-9}
 & \multirow[c]{2}{*}{\texttt{user-item-rate}} & Val & $0.16$ & $0.09$ & $0.22$ & $\bm{1.42}$ & $0.15$ & $ 550.12$ \% \\
 &  & Test & $0.15$ & $0.07$ & $0.17$ & $\bm{0.87}$ & $0.12$ & $ 395.92$ \% \\
\cmidrule{2-9}
 & \multirow[c]{2}{*}{\texttt{user-item-review}} & Val & $0.18$ & $0.05$ & $0.14$ & $\bm{1.03}$ & $0.11$ & $ 476.06$ \% \\
 &  & Test & $0.11$ & $0.04$ & $0.09$ & $\bm{0.47}$ & $0.09$ & $ 313.07$ \% \\
\cmidrule{1-9}
\multirow[c]{2}{*}{\texttt{rel-avito}} & \multirow[c]{2}{*}{\texttt{user-ad-visit}} & Val & $0.01$ & $3.66$ & $0.17$ & $0.09$ & $\bm{5.40}$ & $ 47.37$ \% \\
 &  & Test & $0.00$ & $1.95$ & $0.06$ & $0.02$ & $\bm{3.66}$ & $ 87.09$ \% \\
\cmidrule{1-9}
\multirow[c]{2}{*}{\texttt{rel-hm}} & \multirow[c]{2}{*}{\texttt{user-item-purchase}} & Val & $0.36$ & $1.07$ & $0.44$ & $0.92$ & $\bm{2.64}$ & $ 145.60$ \% \\
 &  & Test & $0.30$ & $0.89$ & $0.38$ & $0.80$ & $\bm{2.81}$ & $ 214.49$ \% \\
\cmidrule{1-9}
\multirow[c]{4}{*}{\texttt{rel-stack}} & \multirow[c]{2}{*}{\texttt{user-post-comment}} & Val & $0.03$ & $2.05$ & $0.04$ & $0.43$ & $\bm{15.17}$ & $ 640.05$ \% \\
 &  & Test & $0.02$ & $1.42$ & $0.04$ & $0.11$ & $\bm{12.72}$ & $ 795.15$ \% \\
\cmidrule{2-9}
 & \multirow[c]{2}{*}{\texttt{post-post-related}} & Val & $0.47$ & $0.00$ & $1.62$ & $0.00$ & $\bm{7.76}$ & $ 378.26$ \% \\
 &  & Test & $1.46$ & $1.74$ & $2.00$ & $0.07$ & $\bm{10.83}$ & $ 440.27$ \% \\
\cmidrule{1-9}
\multirow[c]{4}{*}{\texttt{rel-trial}} & \multirow[c]{2}{*}{\texttt{condition-sponsor-run}} & Val & $2.63$ & $8.58$ & $4.88$ & $3.12$ & $\bm{11.33}$ & $ 32.05$ \% \\
 &  & Test & $2.52$ & $8.42$ & $4.82$ & $2.89$ & $\bm{11.36}$ & $ 34.89$ \% \\
\cmidrule{2-9}
 & \multirow[c]{2}{*}{\texttt{site-sponsor-run}} & Val & $4.91$ & $15.90$ & $10.92$ & $14.09$ & $\bm{17.43}$ & $ 9.65$ \% \\
 &  & Test & $3.75$ & $17.31$ & $8.40$ & $10.70$ & $\bm{19.00}$ & $ 9.74$ \% \\
\cmidrule{1-9}
\multicolumn{2}{c}{\multirow[c]{2}{*}{Average}} & Val & $1.01$ & $3.50$ & $2.07$ & $2.51$ & $\bm{6.68}$ & $297.41$ \% \\
 &  & Test & $0.95$ & $3.55$ & $1.79$ & $1.85$ & $\bm{6.74}$ & $277.26$ \% \\}
\begin{tabular}{lllrrrrrr}
\toprule
\textbf{Dataset} & \textbf{Task} & \textbf{Split} &
\makecell{\textbf{Global}\\\textbf{Popularity}} & \makecell{\textbf{Past}\\\textbf{Visit}} &
\textbf{LightGBM} &
\makecell{\textbf{RDL}\\\textbf{(GraphSAGE)}} & \makecell{\textbf{RDL}\\\textbf{(ID-GNN)}} &
\makecell{\textbf{Rel. Gain}\\\textbf{of RDL}}
\\
\midrule
\multirow[c]{9}{*}{\amazon} & \multirow[c]{3}{*}{\userItemPurchase} 
 & Train & 11,759,844 & 2.18 & - \\
 &  & Val & 802,540 & 2.28 & 0.18 \\
 &  & Test & 918,919 & 2.33 & 0.15 \\

\cmidrule{2-6}

 & \multirow[c]{3}{*}{\userItemRate} 
 & Train & 7,146,115 & 1.8 & - \\
 &  & Val & 519,496 & 2.01 & 0.19 \\
 &  & Test & 599,867 & 2.05 & 0.15 \\

 \cmidrule{2-6}

 & \multirow[c]{3}{*}{\userItemReview} 
 & Train & 5,138,184 & 2.19 & - \\
 &  & Val & 268,651 & 2.3 & 0.18 \\
 &  & Test & 305,476 & 2.4 & 0.15 \\

\cmidrule{1-6}

\multirow[c]{3}{*}{\handm} & \multirow[c]{3}{*}{\userItemPurchase} & 
Train & 13,191,321 & 3.38 & - \\
 &  & Val & 237,152 & 3.18 & 3.51 \\
 &  & Test & 207,996 & 3.10 & 3.76 \\

% \cmidrule{1-6}
% \multirow[c]{3}{*}{\fone} & \multirow[c]{3}{*}{\driverConstructorResult}
% & Train & 1900 &  & - \\
%  &  & Val & 142 & 1.95 & 9.86 \\
%  &  & Test & 105 & 1.84 & 10.48 \\

\cmidrule{1-6}

\multirow[c]{6}{*}{\stackex} & \multirow[c]{3}{*}{\userPostComment}
 & Train & 43,337 & 2.08 & - \\
 & & Val & 1,603 & 1.94 & 3.43 \\
 &  & Test & 1,517 & 2.0 & 4.09 \\

\cmidrule{2-6}

 & \multirow[c]{3}{*}{\postPostLinked}
 & Train & 7,162 & 1.2 & - \\
 &  & Val & 294 & 1.3 & 0.0 \\
 &  & Test & 359 & 1.39 & 1.39 \\

\cmidrule{1-6}

\multirow[c]{6}{*}{\trials} & \multirow[c]{3}{*}{\sponsorConditionRec}
 & Train & 503,176 & 12.51 & - \\
 &  & Val & 30,448 & 14.63 & 34.48 \\
 &  & Test & 25,694 & 12.49 & 38.37 \\

\cmidrule{2-6}

& \multirow[c]{3}{*}{\sponsorFacilityRec}
 & Train & 1,485,360 & 2.27 & - \\
 &  & Val & 80,103 & 2.16 & 20.91 \\
 &  & Test & 50,635 & 1.85 & 23.29 \\

\cmidrule{1-6}

\multirow[c]{3}{*}{\avito} & \multirow[c]{3}{*}{\userAdVisit} & 
Train & 2,738,733 & 31.53 & - \\
 &  & Val & 877,441 & 29.27 & 6.79 \\
 &  & Test & 712,985 & 19.73 & 4.73 \\
\bottomrule
\end{tabular}
\label{tab:link results}
\end{table}

% GNN is better.

\section{Expert Data Scientist User Study}
\label{sec:ds_baseline}

To test RDL in the most challenging circumstances possible, we undertake a human trial wherein a data scientist solves each task by manually designing features and feeds them into tabular methods such at LightGBM or XGBoost~\citep{chen2016xgboost,ke2017lightgbm}. This represents the prior gold-standard for building predictive models on relational databases \citep{heaton2016empirical}, and the key point of comparison for RDL.

We structure our user study along the five main data science workflow steps:
\begin{enumerate}[leftmargin=0.6cm,itemsep=3pt, parsep=0pt]
    \item \textbf{Exploratory data analysis (EDA):} Explore the dataset and task to understand its characteristics, including what column features there are, and if there is any missing data.
    \item \textbf{Feature ideation:} Based on EDA and intuition from prior experiences, propose a set of entity-level features that the data scientist believes may contain predictive signal for the task.
    \item \textbf{Feature enginnering:} Using query languages such as SQL to compute the proposed features, and add them as extra columns to the target table of interest. 
    \item \textbf{Tabular ML:} Run tabular methods such as LightGBM or XGBoost on the table with extra features to produce a predictive model, and record the test performance.
    \item \textbf{Post-hoc analysis of feature importance (Optional):} Common tools include SHAP and LIME, which aim to explain the contribution of each input feature to the final performance. %\kexin{not sure if we need to mention this since we haven't discussed it?} \jr{Was planning to include in appendix}
\end{enumerate}

Consider for example the \handm dataset (schema in Appendix \ref{app:data}) and the task of predicting customer churn. Here the \tbl{customer} table only contains simple biographical information such as username and joining date. To capture more predictive information, additional features, such as  \emph{time since last purchase}, can be computed using the other tables, and added to the \tbl{customer} table. We give a detailed walk-through of the data scientist's work process for solving this specific task in Appendix \ref{app: user study}. We strongly encourage the interested reader to review this, as it highlights the significant amount of task-specific effort that this workflow necessitates.

\xhdr{Limitations of Manual Feature Engineering} This workflow suffers from several fundamental limitations. Most obviously, since features are hand designed they only capture part of the predictive signal in the database, useful signal is easily missed. Additionally, feature complexity is limited by human reasoning abilities, meaning that higher-order interactions between entities are often overlooked. Beyond predictive signal, the other crucial limitation of feature engineering is its extremely manual nature---every time a new model is built a data scientist has to repeat this process, requiring many hours of human labor, and significant quantities of new SQL code to design features \citep{zheng2018feature}. Our RDL models avoid these limitations (see Section \ref{subsec: user study results}).
%\wh{This paragraph feels repetitive of our intro}

\xhdr{Data Scientist} To conduct a thorough comparison to this process, we recruit a high-end data scientist with Stanford CS MSc degree, 4.0 GPA, and 5 years of experience of building machine learning models in the financial industry. This experience includes a significant amount of time building machine learning models in exactly above five steps, as well as broader data science expertise.

\xhdr{User Study Protocol} 
%EDA time is restricted to a maximum of 4 hours per dataset to limit complexity. Feature ideation is performed manually with pen and paper,  and is limited to 1 hour. In practice, the data scientist found that 1 hour was plenty of time to enumerate all promising features at that time. The time taken to write SQL code to generate the features is \emph{unconstrained} in order to eliminate code writing speed as a factor in the study. We do, however, record code writing time for our timing benchmarking. For tabular ML training, we provide a standardized LightGBM training script including comprehensive hyperparameter tuning. The data scientist needs only to feed the table full of engineered features into this training script, which returns test performance results. 
%

\begin{figure}[t]

\centering
\includegraphics[width=\textwidth]{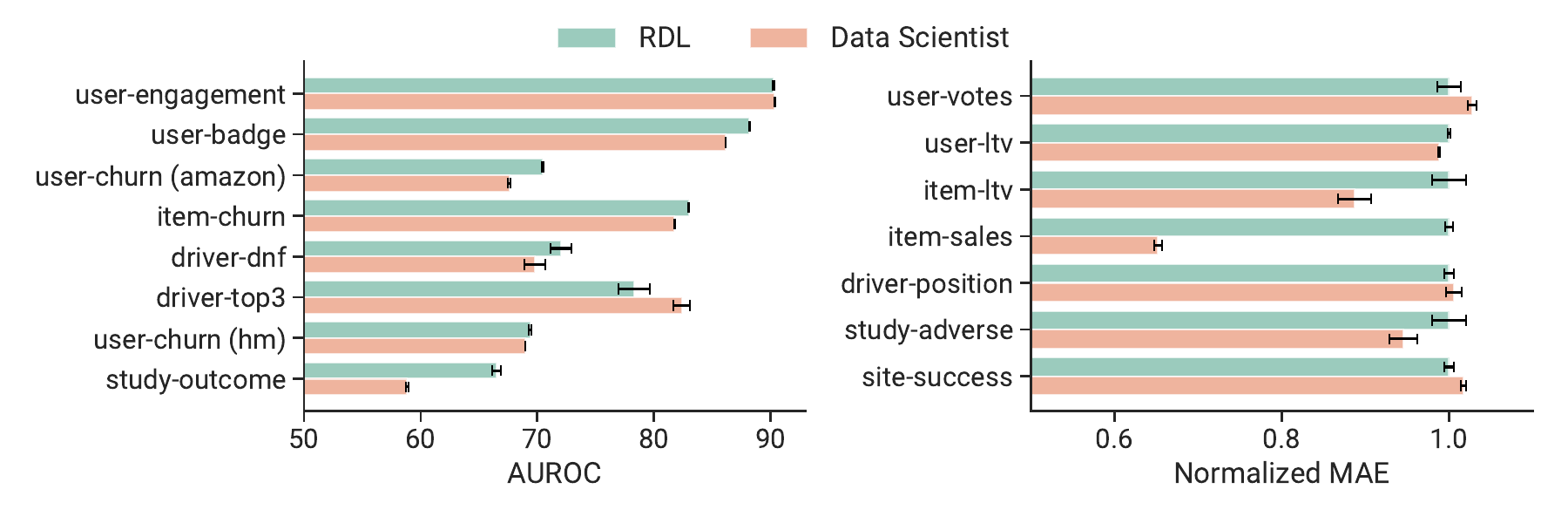}
\vspace{-20pt}
\caption{\textbf{RDL vs. Data Scientist.} Relational Deep Learning matches or outperforms the data scientist in 11 of 15 tasks. Left shows entity classification AUROC, right shows entity regression, reporting MAE normalized so that the RDL MAE is always 1.
}
\vspace{-5pt}
\label{fig:ds main fig}
\end{figure}

Because of the open-ended nature of feature engineering and model development, we follow a specific protocol for the user study in order to standardize the amount of effort dedicated to each dataset and task. Tracking the 5 steps outlined above, we impose the following rules:

\begin{enumerate}
\item \textbf{EDA:} The time allotted for data exploration is capped at 4 hours. This threshold was chosen to give the data scientist enough time to familiarize themselves with the schema, visualize key relationships and distributions, and take stock of any outliers in the dataset, while providing a reasonable limit to the effort applied.
\item \textbf{Feature ideation:} Feature ideation is performed manually with pen and paper, and is limited to 1 hour. In practice, the data scientist found that 1 hour was plenty of time to enumerate all promising features at that time, especially since many ideas naturally arise during the EDA process already.
\item \textbf{Feature engineering:} The features described during the ideation phase are then computed using SQL queries. The time taken to write SQL code to generate the features is unconstrained in order to eliminate code writing speed as a factor in the study. We do, however, record code writing time for our timing benchmarking. This stage presented the most variability in terms of time commitment, partly because it is unconstrained, but mostly because the implementation complexity of the features itself is highly variable.
\item \textbf{Tabular ML:} For tabular ML training, we provide a standardized LightGBM training script including comprehensive hyperparameter tuning. The data scientist needs only to feed the table full of engineered features into this training script, which returns test performance results. However, there is some non-trivial amount of work required to transform the output of the SQL queries from the previous section into the Python objects (arrays) required for training LightGBM. Again, the time taken for this additional pre-preocessing is recorded.
\item \textbf{Post-hoc analysis of feature importance:} Finally, after successfully training a model, an evaluation of model predictions and feature importance is carried out. This mostly serves as a general sanity check and an interesting corollary of the data scientist’s work that provides task-specific insights (see Appendix \ref{app: user study}). In practice, this took no more than a few minutes per task and this time was not counted toward the total time commitment.
\end{enumerate}

\xhdr{Reproducibility}  All of the data scientist's workings are released\footnote{See \userstudy.} to ensure reproducibility and demonstrate the significant lengths gone through to build as accurate models as possible. In Appendix \ref{app: user study} we walk through a complete example for a single dataset and task, showing the data-centric insights it yields. An important by-product is a close analysis of which features contribute to model performance, which we believe will help inspire future well-motivated RDL research directions.

\begin{figure}[t]

\centering
\includegraphics[width=\textwidth]{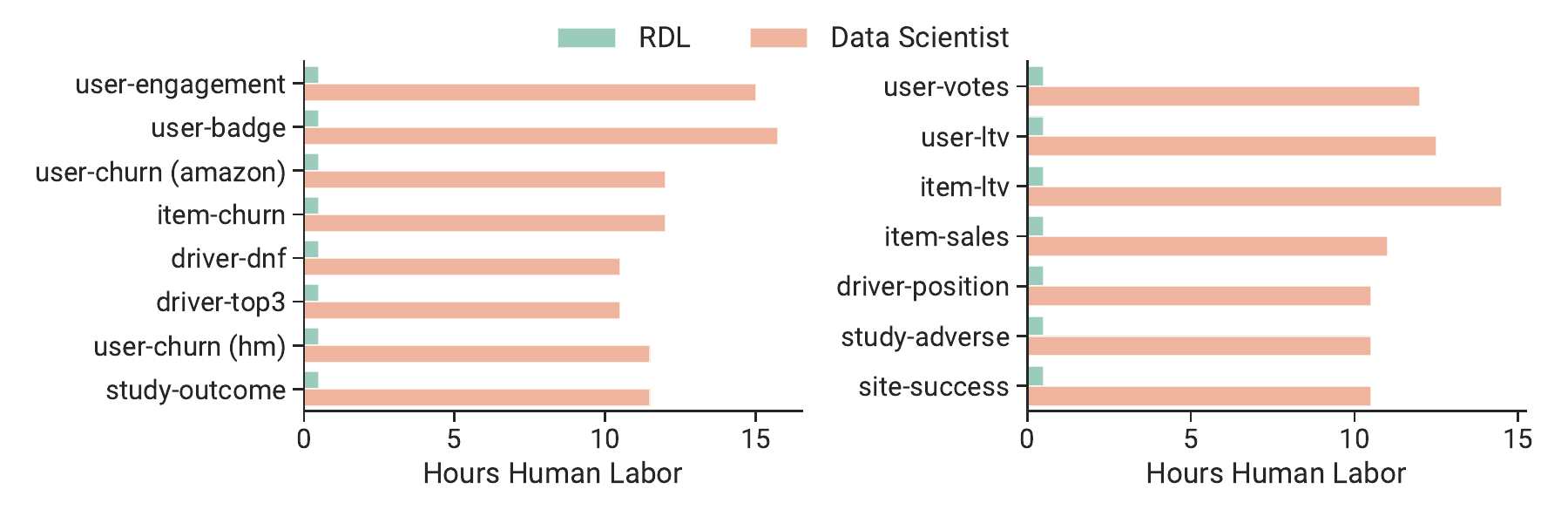}
\vspace{-20pt}
\caption{\textbf{RDL vs. Data Scientist.} Relational Deep Learning reduces the hours of human work required to solve a new task by 96\% on average (from 12.3 to 0.5 hours). Left shows node-level classification, right shows node-level regression.
}
\vspace{-5pt}
\label{fig:ds main fig hours}
\end{figure}

\subsection{Results}\label{subsec: user study results}

As well as (i) raw predictive power, we compare the data scientist to our RDL models in terms of (ii) hours of human  work, and (iii) number of new lines of code required to solve each task. We measure the \emph{marginal} effort, meaning that we do not include code infrastructure that is reused across tasks, including for example data loading logic and training scripts for RDL or LightGBM models. 

\begin{figure}[t]

\centering
\includegraphics[width=\textwidth]{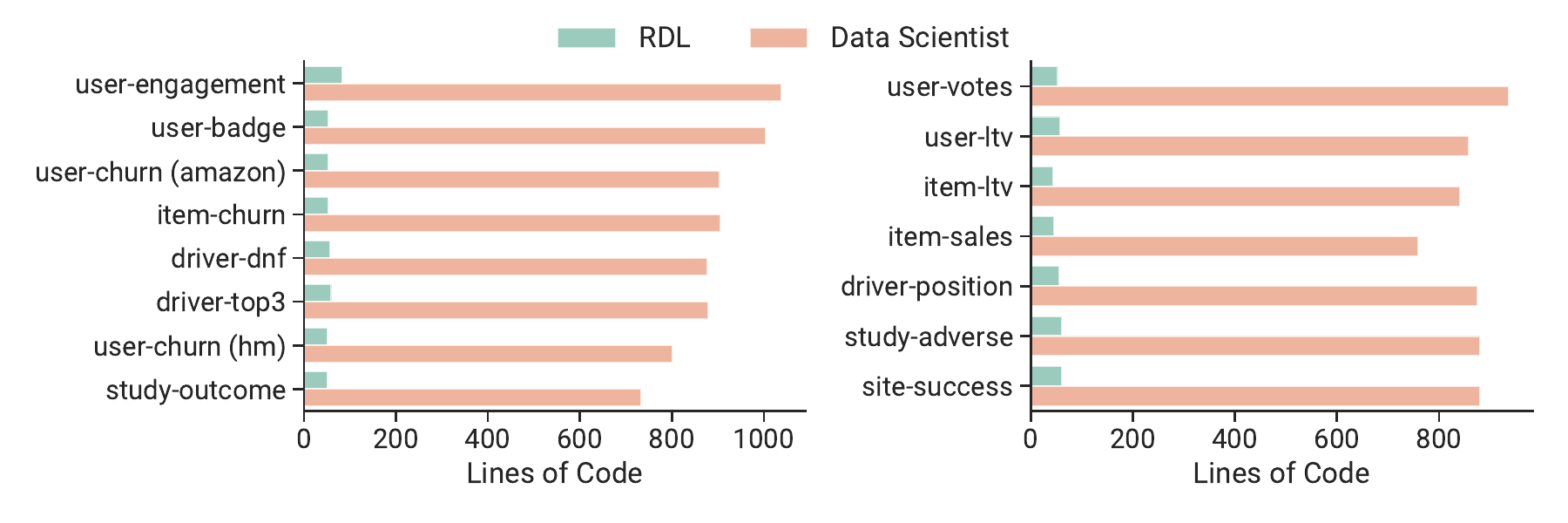}
\vspace{-20pt}
\caption{\textbf{RDL vs. Data Scientist.} Relational Deep Learning reduces the new lines of code needed to solve a new task by 94\%. Left shows entity classification, right shows entity regression.
}
\vspace{-5pt}
\label{fig:ds main fig lines}
\end{figure}

\xhdr{Summary} Figures \ref{fig:ds main fig}, \ref{fig:ds main fig hours}, and \ref{fig:ds main fig lines} show that RDL learns highly predictive models, outperforming the data scientist in 11 of 15 tasks, whilst reducing hours worked by $96\%$ on average, and lines of code by $94\%$ on average. On average, it took the data scientist 12.3 hours to solve each task using traditional feature engineering. By contrast it takes roughly 30 minutes to solve a task with RDL.
% \jure{At the same time we also observe that RLD models are X\% more accurate on the average (X\% for entity classification, Y\% for entity regression, Z\% for link prediction).} \alejandro{Provisionally: X = 0.1\% and Y = -10.7\% and there is no Z (no user-study on link pred). Doesn't align with the message. IDK if we still want to include.}

This observation is \emph{the} central value proposition of relational deep learning, pointing the way to unlocking new levels of predictive power, and potentially a new economic model for solving predictive tasks on relational databases. Replacing hand-crafted solutions with end-to-end learnable models has been a key takeaway from the last 15 years of AI research. It is therefore remarkable how little impact deep learning has had on ML on relational databases, one of the most widespread applied ML use cases. To the best of our knowledge, RDL represents the first proposal for a deep learning approach for relational databases that has demonstrated efficacy compared with established data science workflows. 

We highlight that all \relbench tasks were solved with a single set of default hyperparameters (with 2 exceptions requiring small modifications to learning rate, number of epochs, and GNN aggregation function). This demonstrates the robustness of RDL, and that the performance of RDL in Figure \ref{fig:ds main fig} is not due to extensive hyperparamter search. Indeed, the single set of RDL hyperparameters is compared to a carefully tuned LightGBM, which was allowed to search over 10 sets of hyperparameters.

\xhdr{Predictive Power} Results shown in Figures \ref{fig:ds main fig}. Whilst  outperforming the data scientist in 11 of 15 tasks, we note that RDL best outperforms the data scientist on classification tasks, struggling more on regression. Indeed it was necessary for us to apply a ``boosting'' to the RDL model to improve performance (see Appendix \ref{app: user study} for details). Even with boosting, the data scientist model outperforms RDL in several cases. One cause we identify is that the MLP output head of the GNN is poorly suited to regression tasks (see Appendix \ref{app: user study} for our analysis). This suggests an opportunity for improved output heads for regression tasks. We stress that our RDL implementation is an \emph{initial} demonstration. We believe there is significant scope for new research leading to large improvements in performance. In particular, ideas from graph ML, deep tabular ML, and time-series modeling are well suited to advance RDL.

\xhdr{Human Work} Results shown in Figure \ref{fig:ds main fig hours}. In our user study RDL required 96\% less hours work to solve a new task, compared to the data scientist work flow. The RDL solutions always took less than an hour to write, whilst the data scientist took $12$ hours on average, with a standard deviation of $1.6$ hours. We emphasize that this measures \emph{marginal} effort, i.e., it does not include reusable code that can be amortized over many tasks. RDL compares favorably to data scientist because a large majority of RDL code is reusable for new tasks (a GNN architecture and training loop needs only to be defined once) whereas a large portion of the data scientist's code is task specific and must be re-done afresh for every new task that needs to be solved. 

\xhdr{Lines of Code} Results shown in Figure \ref{fig:ds main fig lines}. For the RDL model, the only new addition needed to solve a new task is the code describing how to compute the training supervision for the RDL, which is stored in the training table. This requires a similar number of lines of code for each task, with 56 lines of code on average, with standard deviation $8.8$, with the data scientist requiring with $878 \pm 77$. The minimum lines of code required by RDL is 44, compared to 734 for the data scientist, and maximum is 84 compared to 1039 for the data scientist. Examples of the RDL code required to solve \amazon tasks can be viewed \href{https://github.com/snap-stanford/relbench/blob/main/relbench/tasks/amazon.py#L19}{here}. For the data scientist pipeline, we record the number of lines of code for EDA and SQL files, and the manipulations needed to format data to be fed into the pre-prepared LightGBM script. 
\section{Related Work}
\label{sec:related}

\xhdr{Graph Machine Learning Benchmarks} Challenging and realistic benchmarks drive innovation in methodology. A classic example is the ImageNet~\citep{deng2009imagenet},  introduced prior to the rise of deep learning, which was a key catalyst for the seminal work of \cite{krizhevsky2017imagenet}. In graph machine learning, benchmarks such as the Open Graph Benchmark \citep{hu2020open}, TUDataset \citep{morris2020tudataset}, and more recently, the Temporal Graph Benchmark \citep{huang2024temporal} have sustained the growth and maturation of graph machine learning as a field.
%used common graph data sources including small molecules, bioinformatics, and social networks. 
\relbench differs since instead of collecting together tasks are already recognized as graph machine learning tasks, \relbench presents existing tasks typically solved using other methods, as graph ML tasks. As a consequence, \relbench significantly expands the space of problems solvable using graph ML. Whilst graph ML is a key part of this benchmark, relational deep learning is a \emph{new problem}, requiring only need good GNNs, but also innovation on tabular learning to fuse multimodal input data with the GNN, temporal learning, and even graph construction. We believe that advancing the state-of-the-art on \relbench will involve progress in all of these directions.

\xhdr{Relational Deep Learning}
Several works have proposed to use graph neural networks for learning on relational data~\citep{schlichtkrull2018relational, cvitkovic2020supervisedrd, vsir2021deep, zahradnik2023deep}. They explored different graph neural network architectures on (heterogeneous) graphs, leveraging relational structure. Recently, \cite{fey2023relational} proposed a general end-to-end learnable framework for solving predictive tasks on relational databases, treating temporality as a core concept. 
\relbench provides a comprehensive testbed to develop these ideas further. 

\section{Conclusion}
\label{sec:conclusion}

This work introduces \relbench, a benchmark to facilitate research on relational deep learning~\citep{fey2023relational}. 
\relbench provides diverse and realistic relational databases and define practical predictive tasks that cover both entity-level prediction and entity link prediction.
In addition, we provide the first open-source implementation of relational deep learning and validated its effectiveness over the common practice of manual feature engineering by an experienced data scientist.
We hope \relbench will catalyze further research on relational deep learning to achieve highly-accurate prediction over complex multi-tabular datasets without manual feature engineering.

%\jure{Add ACKs section!}
\begin{ack}

We thank Shirley Wu, Kaidi Cao, Rok Sosic, Yu He, Qian Huang, Bruno Ribeiro and Michi Yasunaga for discussions and for providing feedback on our manuscript.
We also gratefully acknowledge the support of
NSF under Nos. OAC-1835598 (CINES), CCF-1918940 (Expeditions), DMS-2327709 (IHBEM);
Stanford Data Applications Initiative,
Wu Tsai Neurosciences Institute,
Stanford Institute for Human-Centered AI,
Chan Zuckerberg Initiative,
Amazon, Genentech, GSK, Hitachi, SAP, and UCB.
The content is solely the responsibility of the authors and does not necessarily represent the official views of the funding entities.

\end{ack}

\bibliography{refs}
\bibliographystyle{mynat}

% \newpage
% \input{neurips/checklist.tex}

\newpage
\appendix

\section{Additional Task Information}
\label{app: task info}

For reference, the following list documents all the predictive tasks in \relbench.
\begin{enumerate}
\item \amazon

    Node-level tasks:
    \begin{enumerate}
    \item \userChurn: For each user, predict 1 if the customer does not review any product in the next 3 months, and 0 otherwise.
    \item \userLtv: For each user, predict the $\$$ value of the total number of products they buy and review in the next 3 months.
    \item \itemChurn: For each product, predict 1 if the  product does not receive any reviews in the next 3 months.
    \item \itemLtv: For each product, predict the $\$$ value of the total number purchases and reviews it recieves in the next 3 months.
    \end{enumerate}

    Link-level tasks:
    \begin{enumerate}
    \item \userItemPurchase: Predict the list of distinct items each customer will purchase in the
    next 3 months. 
    \item \userItemRate: Predict the list of distinct items each customer will purchase and give a 5 star review in the
    next 3 months.
    \item \userItemReview: Predict the list of distinct items each customer will purchase and give a detailed review in the
    next 3 months.
    \end{enumerate}
\item \avito

    Node-level tasks:
    \begin{enumerate}
    \item \userVisit: Predict whether each customer will visit more than one Ad in the next 4 days.
    \item \userClick: Predict whether each customer will click on more than one Ads in the next 4 day.
    \item \adsCTR: Assuming the Ad will be clicked in the next 4 days, predict the Click-Through-Rate (CTR) for each Ad.
    \end{enumerate}
    Link-level tasks:
    \begin{enumerate}
    \item \userAdVisit: Predict the list of ads a user will visit in the next 4 days.
    \end{enumerate}
\item \fone

    Node-level tasks:
    \begin{enumerate}
    \item \driverPosition: Predict the average finishing position of each driver
    all races in the next 2 months.
    \item \driverDNF: For each driver predict the if they will DNF (did not finish) a race in the next 1 month.
    \item \driverTopThree: For each driver predict if they will qualify in the top-3 for a race in the next 1 month.
    \end{enumerate}

   % Link-level tasks:
   % \begin{enumerate}
  %  \item \driverConstructorResult: For each driver predict which constructors they will join in the next 10 years. \jr{TODO: delete}
    %\end{enumerate}

\item \handm

    Node-level tasks:
    \begin{enumerate}
    \item \userChurn: Predict the churn for a customer (no transactions) in the next week.
    \item \itemSales: Predict the total sales for an article (the sum of prices of the
    associated transactions) in the next week.
    \end{enumerate}

    Link-level tasks:
    \begin{enumerate}
    \item \userItemPurchase: Predict the list of articles each customer will purchase in the next
    seven days.
    \end{enumerate}

\item \stackex

    Node-level tasks:
    \begin{enumerate}
    \item \userEngage: For each user predict if a user will make any votes, posts, or comments in the next 3 months.
    \item \postVotes: For each user post predict how many votes it will receive in the next 3 months
    \item \userBadge: For each user predict if each user will receive in a new badge the next 3 months.
    \end{enumerate}

    Link-level tasks:
    \begin{enumerate}
    \item \userPostComment: Predict a list of existing posts that a user will comment in the next
    two years.
    \item \postPostLinked: Predict a list of existing posts that users will link a given post to in the next
    two years.
    \end{enumerate}

\item \trials

    Node-level tasks:
    \begin{enumerate}
    \item \studyOutcome: Predict if the trials in the next 1 year will achieve its primary outcome.
    \item \studyAdverse: Predict the number of affected patients with severe advsere events/death for the trial in the next 1 year.
    \item \facilitySuccess: Predict the success rate of a trial site in the next 1 year.
    \end{enumerate}

    Link-level tasks:
    \begin{enumerate}
    \item \sponsorConditionRec: Predict whether this condition will have which sponsors.
    \item \sponsorFacilityRec: Predict whether this sponsor will have a trial in a facility.
    \end{enumerate}
    
\item \event

    Node-level tasks:
    \begin{enumerate}
    \item \userAttendance: Predict how many events each user will respond yes or maybe in the next seven days.
    \item \userRepeat: Predict whether a user will attend an event(by responding yes or maybe) in the next 7 days if they have already attended an event in the last 14 days.
    \item \userIgnore: Predict whether a user will ignore more than 2 event invitations
    in the next 7 days.
    \end{enumerate}
%\item \avito

%    Node-level tasks:
%    \begin{enumerate}
%    \item \userClicks: For each user of Avito, predict how many ads they will click in the next four days.
%    \end{enumerate}
    
    % Link-level tasks:
    % \begin{enumerate}
    %    \item \userAdClick: For each user of Avito, predict which ad they will click in the next four days.
    % \end{enumerate}
\end{enumerate}

\section{Experiment Details and Additional Results}\label{app:gnn_ablations}

\begin{table}
    % \vspace{-40pt}
    \caption{Entity classification results (AUROC mean$_{\pm \text{std}}$ over $5$ runs, higher is better) on \relbench. Best values are in bold along with those not statistically different from it.
    % \jure{What are these subscripts. Can you explain in the caption. Why are sometimes both methods bolded.} \jure{Remove subscripts, bold the max value; create another table in appendix with standard errors.}
    }
    \centering
    % \setlength{\tabcolsep}{3pt}
    % \scriptsize
    \tiny
    \newcommand{\tabledata}{\multirow[c]{4}{*}{\texttt{rel-amazon}} & \multirow[c]{2}{*}{\texttt{user-churn}} & Val & $52.05_{\pm 0.06}$ & $\bm{70.45}_{\pm 0.06}$ \\
 &  & Test & $52.22_{\pm 0.06}$ & $\bm{70.42}_{\pm 0.05}$ \\
\cmidrule{2-5}
 & \multirow[c]{2}{*}{\texttt{item-churn}} & Val & $62.39_{\pm 0.20}$ & $\bm{82.39}_{\pm 0.02}$ \\
 &  & Test & $62.54_{\pm 0.18}$ & $\bm{82.81}_{\pm 0.03}$ \\
\cmidrule{1-5}
\multirow[c]{4}{*}{\texttt{rel-avito}} & \multirow[c]{2}{*}{\texttt{user-visits}} & Val & $53.31_{\pm 0.09}$ & $\bm{69.65}_{\pm 0.04}$ \\
 &  & Test & $53.05_{\pm 0.32}$ & $\bm{66.20}_{\pm 0.10}$ \\
\cmidrule{2-5}
 & \multirow[c]{2}{*}{\texttt{user-clicks}} & Val & $55.63_{\pm 0.31}$ & $\bm{64.73}_{\pm 0.32}$ \\
 &  & Test & $53.60_{\pm 0.59}$ & $\bm{65.90}_{\pm 1.95}$ \\
\cmidrule{1-5}
\multirow[c]{4}{*}{\texttt{rel-event}} & \multirow[c]{2}{*}{\texttt{user-repeat}} & Val & $\bm{67.76}_{\pm 0.97}$ & $\bm{71.25}_{\pm 2.53}$ \\
 &  & Test & $68.04_{\pm 1.82}$ & $\bm{76.89}_{\pm 1.59}$ \\
\cmidrule{2-5}
 & \multirow[c]{2}{*}{\texttt{user-ignore}} & Val & $87.96_{\pm 0.28}$ & $\bm{91.70}_{\pm 0.33}$ \\
 &  & Test & $79.93_{\pm 0.49}$ & $\bm{81.62}_{\pm 1.11}$ \\
\cmidrule{1-5}
\multirow[c]{4}{*}{\texttt{rel-f1}} & \multirow[c]{2}{*}{\texttt{driver-dnf}} & Val & $68.42_{\pm 1.14}$ & $\bm{71.36}_{\pm 1.54}$ \\
 &  & Test & $\bm{68.56}_{\pm 3.89}$ & $\bm{72.62}_{\pm 0.27}$ \\
\cmidrule{2-5}
 & \multirow[c]{2}{*}{\texttt{driver-top3}} & Val & $67.76_{\pm 2.75}$ & $\bm{77.64}_{\pm 3.16}$ \\
 &  & Test & $\bm{73.92}_{\pm 5.75}$ & $\bm{75.54}_{\pm 0.63}$ \\
\cmidrule{1-5}
\multirow[c]{2}{*}{\texttt{rel-hm}} & \multirow[c]{2}{*}{\texttt{user-churn}} & Val & $56.05_{\pm 0.05}$ & $\bm{70.42}_{\pm 0.09}$ \\
 &  & Test & $55.21_{\pm 0.12}$ & $\bm{69.88}_{\pm 0.21}$ \\
\cmidrule{1-5}
\multirow[c]{4}{*}{\texttt{rel-stack}} & \multirow[c]{2}{*}{\texttt{user-engagement}} & Val & $65.12_{\pm 0.25}$ & $\bm{90.21}_{\pm 0.07}$ \\
 &  & Test & $63.39_{\pm 0.26}$ & $\bm{90.59}_{\pm 0.09}$ \\
\cmidrule{2-5}
 & \multirow[c]{2}{*}{\texttt{user-badge}} & Val & $65.39_{\pm 0.05}$ & $\bm{89.86}_{\pm 0.08}$ \\
 &  & Test & $63.43_{\pm 0.12}$ & $\bm{88.86}_{\pm 0.08}$ \\
\cmidrule{1-5}
\multirow[c]{2}{*}{\texttt{rel-trial}} & \multirow[c]{2}{*}{\texttt{study-outcome}} & Val & $\bm{68.30}_{\pm 0.53}$ & $\bm{68.18}_{\pm 0.49}$ \\
 &  & Test & $\bm{70.09}_{\pm 1.41}$ & $\bm{68.60}_{\pm 1.01}$ \\}
    \begin{tabular}{lllrr}
    \toprule
    \textbf{Dataset} & \textbf{Task} & \textbf{Split} & \textbf{LightGBM} & \textbf{RDL} \\
    \midrule
    \multirow[c]{9}{*}{\amazon} & \multirow[c]{3}{*}{\userItemPurchase} 
 & Train & 11,759,844 & 2.18 & - \\
 &  & Val & 802,540 & 2.28 & 0.18 \\
 &  & Test & 918,919 & 2.33 & 0.15 \\

\cmidrule{2-6}

 & \multirow[c]{3}{*}{\userItemRate} 
 & Train & 7,146,115 & 1.8 & - \\
 &  & Val & 519,496 & 2.01 & 0.19 \\
 &  & Test & 599,867 & 2.05 & 0.15 \\

 \cmidrule{2-6}

 & \multirow[c]{3}{*}{\userItemReview} 
 & Train & 5,138,184 & 2.19 & - \\
 &  & Val & 268,651 & 2.3 & 0.18 \\
 &  & Test & 305,476 & 2.4 & 0.15 \\

\cmidrule{1-6}

\multirow[c]{3}{*}{\handm} & \multirow[c]{3}{*}{\userItemPurchase} & 
Train & 13,191,321 & 3.38 & - \\
 &  & Val & 237,152 & 3.18 & 3.51 \\
 &  & Test & 207,996 & 3.10 & 3.76 \\

% \cmidrule{1-6}
% \multirow[c]{3}{*}{\fone} & \multirow[c]{3}{*}{\driverConstructorResult}
% & Train & 1900 &  & - \\
%  &  & Val & 142 & 1.95 & 9.86 \\
%  &  & Test & 105 & 1.84 & 10.48 \\

\cmidrule{1-6}

\multirow[c]{6}{*}{\stackex} & \multirow[c]{3}{*}{\userPostComment}
 & Train & 43,337 & 2.08 & - \\
 & & Val & 1,603 & 1.94 & 3.43 \\
 &  & Test & 1,517 & 2.0 & 4.09 \\

\cmidrule{2-6}

 & \multirow[c]{3}{*}{\postPostLinked}
 & Train & 7,162 & 1.2 & - \\
 &  & Val & 294 & 1.3 & 0.0 \\
 &  & Test & 359 & 1.39 & 1.39 \\

\cmidrule{1-6}

\multirow[c]{6}{*}{\trials} & \multirow[c]{3}{*}{\sponsorConditionRec}
 & Train & 503,176 & 12.51 & - \\
 &  & Val & 30,448 & 14.63 & 34.48 \\
 &  & Test & 25,694 & 12.49 & 38.37 \\

\cmidrule{2-6}

& \multirow[c]{3}{*}{\sponsorFacilityRec}
 & Train & 1,485,360 & 2.27 & - \\
 &  & Val & 80,103 & 2.16 & 20.91 \\
 &  & Test & 50,635 & 1.85 & 23.29 \\

\cmidrule{1-6}

\multirow[c]{3}{*}{\avito} & \multirow[c]{3}{*}{\userAdVisit} & 
Train & 2,738,733 & 31.53 & - \\
 &  & Val & 877,441 & 29.27 & 6.79 \\
 &  & Test & 712,985 & 19.73 & 4.73 \\
    \bottomrule
    \end{tabular}
    \label{tab:app_classif}
\end{table}

\begin{table}
\caption{Entity regression results (MAE mean$_{\pm \text{std}}$ over $5$ runs, lower is better) on \relbench. Best values are in bold along with those not statistically different from it.
% \jure{same question here}
}
\centering
\setlength{\tabcolsep}{4pt}
% \scriptsize
\tiny
\newcommand{\tabledata}{\multirow[c]{4}{*}{\texttt{rel-amazon}} & \multirow[c]{2}{*}{\texttt{user-ltv}} & Val & $14.141$ & $20.740$ & $14.141$ & $17.685$ & $15.978$ & $14.141_{\pm 0.000}$ & $\bm{12.132}_{\pm 0.007}$ \\
 &  & Test & $16.783$ & $22.121$ & $16.783$ & $19.055$ & $17.423$ & $16.783_{\pm 0.000}$ & $\bm{14.313}_{\pm 0.013}$ \\
\cmidrule{2-10}
 & \multirow[c]{2}{*}{\texttt{item-ltv}} & Val & $72.096$ & $78.110$ & $59.471$ & $80.466$ & $68.922$ & $55.741_{\pm 0.049}$ & $\bm{45.140}_{\pm 0.068}$ \\
 &  & Test & $77.126$ & $81.852$ & $64.234$ & $78.423$ & $66.436$ & $60.569_{\pm 0.047}$ & $\bm{50.053}_{\pm 0.163}$ \\
\cmidrule{1-10}
\multirow[c]{2}{*}{\texttt{rel-avito}} & \multirow[c]{2}{*}{\texttt{ad-ctr}} & Val & $0.048$ & $0.048$ & $0.040$ & $0.044$ & $0.044$ & $0.037_{\pm 0.000}$ & $\bm{0.037}_{\pm 0.000}$ \\
 &  & Test & $0.052$ & $0.051$ & $0.043$ & $0.046$ & $0.046$ & $\bm{0.041}_{\pm 0.000}$ & $\bm{0.041}_{\pm 0.001}$ \\
\cmidrule{1-10}
\multirow[c]{2}{*}{\texttt{rel-event}} & \multirow[c]{2}{*}{\texttt{user-attendance}} & Val & $0.262$ & $0.457$ & $0.262$ & $0.296$ & $0.268$ & $0.262_{\pm 0.000}$ & $\bm{0.255}_{\pm 0.007}$ \\
 &  & Test & $\bm{0.264}$ & $0.470$ & $\bm{0.264}$ & $0.304$ & $0.269$ & $\bm{0.264}_{\pm 0.000}$ & $\bm{0.258}_{\pm 0.006}$ \\
\cmidrule{1-10}
\multirow[c]{2}{*}{\texttt{rel-f1}} & \multirow[c]{2}{*}{\texttt{driver-position}} & Val & $11.083$ & $4.334$ & $4.136$ & $7.181$ & $7.114$ & $3.450_{\pm 0.030}$ & $\bm{3.193}_{\pm 0.024}$ \\
 &  & Test & $11.926$ & $4.513$ & $4.399$ & $8.501$ & $8.519$ & $\bm{4.170}_{\pm 0.137}$ & $\bm{4.022}_{\pm 0.119}$ \\
\cmidrule{1-10}
\multirow[c]{2}{*}{\texttt{rel-hm}} & \multirow[c]{2}{*}{\texttt{item-sales}} & Val & $0.086$ & $0.142$ & $0.086$ & $0.117$ & $0.086$ & $0.086_{\pm 0.000}$ & $\bm{0.065}_{\pm 0.000}$ \\
 &  & Test & $0.076$ & $0.134$ & $0.076$ & $0.111$ & $0.078$ & $0.076_{\pm 0.000}$ & $\bm{0.056}_{\pm 0.000}$ \\
\cmidrule{1-10}
\multirow[c]{2}{*}{\texttt{rel-stack}} & \multirow[c]{2}{*}{\texttt{post-votes}} & Val & $0.062$ & $0.146$ & $0.062$ & $0.102$ & $0.064$ & $0.062_{\pm 0.000}$ & $\bm{0.059}_{\pm 0.000}$ \\
 &  & Test & $0.068$ & $0.149$ & $0.068$ & $0.106$ & $0.069$ & $0.068_{\pm 0.000}$ & $\bm{0.065}_{\pm 0.000}$ \\
\cmidrule{1-10}
\multirow[c]{4}{*}{\texttt{rel-trial}} & \multirow[c]{2}{*}{\texttt{study-adverse}} & Val & $57.083$ & $75.008$ & $56.786$ & $57.083$ & $57.083$ & $\bm{45.774}_{\pm 1.191}$ & $\bm{46.290}_{\pm 0.304}$ \\
 &  & Test & $57.930$ & $73.781$ & $57.533$ & $57.930$ & $57.930$ & $\bm{44.011}_{\pm 0.998}$ & $\bm{44.473}_{\pm 0.209}$ \\
\cmidrule{2-10}
 & \multirow[c]{2}{*}{\texttt{site-success}} & Val & $0.475$ & $0.462$ & $0.475$ & $0.447$ & $0.450$ & $0.417_{\pm 0.003}$ & $\bm{0.401}_{\pm 0.009}$ \\
 &  & Test & $0.462$ & $0.468$ & $0.462$ & $0.448$ & $0.441$ & $0.425_{\pm 0.003}$ & $\bm{0.400}_{\pm 0.020}$ \\}
\begin{tabular}{lllrrrrrrr}
\toprule
\textbf{Dataset} & \textbf{Task} & \textbf{Split} & \makecell{\textbf{Global}\\\textbf{Zero}} & \makecell{\textbf{Global}\\\textbf{Mean}} & \makecell{\textbf{Global}\\\textbf{Median}} & \makecell{\textbf{Entity}\\\textbf{Mean}} & \makecell{\textbf{Entity}\\\textbf{Median}} &
\textbf{LightGBM} &
\textbf{RDL}\\
\midrule
\multirow[c]{9}{*}{\amazon} & \multirow[c]{3}{*}{\userItemPurchase} 
 & Train & 11,759,844 & 2.18 & - \\
 &  & Val & 802,540 & 2.28 & 0.18 \\
 &  & Test & 918,919 & 2.33 & 0.15 \\

\cmidrule{2-6}

 & \multirow[c]{3}{*}{\userItemRate} 
 & Train & 7,146,115 & 1.8 & - \\
 &  & Val & 519,496 & 2.01 & 0.19 \\
 &  & Test & 599,867 & 2.05 & 0.15 \\

 \cmidrule{2-6}

 & \multirow[c]{3}{*}{\userItemReview} 
 & Train & 5,138,184 & 2.19 & - \\
 &  & Val & 268,651 & 2.3 & 0.18 \\
 &  & Test & 305,476 & 2.4 & 0.15 \\

\cmidrule{1-6}

\multirow[c]{3}{*}{\handm} & \multirow[c]{3}{*}{\userItemPurchase} & 
Train & 13,191,321 & 3.38 & - \\
 &  & Val & 237,152 & 3.18 & 3.51 \\
 &  & Test & 207,996 & 3.10 & 3.76 \\

% \cmidrule{1-6}
% \multirow[c]{3}{*}{\fone} & \multirow[c]{3}{*}{\driverConstructorResult}
% & Train & 1900 &  & - \\
%  &  & Val & 142 & 1.95 & 9.86 \\
%  &  & Test & 105 & 1.84 & 10.48 \\

\cmidrule{1-6}

\multirow[c]{6}{*}{\stackex} & \multirow[c]{3}{*}{\userPostComment}
 & Train & 43,337 & 2.08 & - \\
 & & Val & 1,603 & 1.94 & 3.43 \\
 &  & Test & 1,517 & 2.0 & 4.09 \\

\cmidrule{2-6}

 & \multirow[c]{3}{*}{\postPostLinked}
 & Train & 7,162 & 1.2 & - \\
 &  & Val & 294 & 1.3 & 0.0 \\
 &  & Test & 359 & 1.39 & 1.39 \\

\cmidrule{1-6}

\multirow[c]{6}{*}{\trials} & \multirow[c]{3}{*}{\sponsorConditionRec}
 & Train & 503,176 & 12.51 & - \\
 &  & Val & 30,448 & 14.63 & 34.48 \\
 &  & Test & 25,694 & 12.49 & 38.37 \\

\cmidrule{2-6}

& \multirow[c]{3}{*}{\sponsorFacilityRec}
 & Train & 1,485,360 & 2.27 & - \\
 &  & Val & 80,103 & 2.16 & 20.91 \\
 &  & Test & 50,635 & 1.85 & 23.29 \\

\cmidrule{1-6}

\multirow[c]{3}{*}{\avito} & \multirow[c]{3}{*}{\userAdVisit} & 
Train & 2,738,733 & 31.53 & - \\
 &  & Val & 877,441 & 29.27 & 6.79 \\
 &  & Test & 712,985 & 19.73 & 4.73 \\
\bottomrule
\end{tabular}
\label{tab:app_regression}
\end{table}

\begin{table}
\caption{Link prediction results (MAP mean$_{\pm \text{std}}$ over $5$ runs, higher is better) on \relbench. Best values are in bold along with those not statistically different from it.
% \jure{again same}
% \jure{to all tables, can be get avg. relative improvement/difference between RDL and DT?}
}
\centering
% \setlength{\tabcolsep}{4pt}
% \scriptsize
\tiny
\newcommand{\tabledata}{\multirow[c]{6}{*}{\texttt{rel-amazon}} & \multirow[c]{2}{*}{\texttt{user-item-purchase}} & Val & $0.31$ & $0.07$ & $0.18_{\pm 0.07}$ & $\bm{1.53}_{\pm 0.05}$ & $0.13_{\pm 0.00}$ \\
 &  & Test & $0.24$ & $0.06$ & $0.16_{\pm 0.05}$ & $\bm{0.74}_{\pm 0.08}$ & $0.10_{\pm 0.00}$ \\
\cmidrule{2-8}
 & \multirow[c]{2}{*}{\texttt{user-item-rate}} & Val & $0.16$ & $0.09$ & $0.22_{\pm 0.02}$ & $\bm{1.42}_{\pm 0.06}$ & $0.15_{\pm 0.00}$ \\
 &  & Test & $0.15$ & $0.07$ & $0.17_{\pm 0.01}$ & $\bm{0.87}_{\pm 0.05}$ & $0.12_{\pm 0.00}$ \\
\cmidrule{2-8}
 & \multirow[c]{2}{*}{\texttt{user-item-review}} & Val & $0.18$ & $0.05$ & $0.14_{\pm 0.03}$ & $\bm{1.03}_{\pm 0.03}$ & $0.11_{\pm 0.00}$ \\
 &  & Test & $0.11$ & $0.04$ & $0.09_{\pm 0.01}$ & $\bm{0.47}_{\pm 0.05}$ & $0.09_{\pm 0.00}$ \\
\cmidrule{1-8}
\multirow[c]{2}{*}{\texttt{rel-avito}} & \multirow[c]{2}{*}{\texttt{user-ad-visit}} & Val & $0.01$ & $3.66$ & $0.17_{\pm 0.01}$ & $0.09_{\pm 0.01}$ & $\bm{5.40}_{\pm 0.02}$ \\
 &  & Test & $0.00$ & $1.95$ & $0.06_{\pm 0.01}$ & $0.02_{\pm 0.00}$ & $\bm{3.66}_{\pm 0.02}$ \\
\cmidrule{1-8}
\multirow[c]{2}{*}{\texttt{rel-hm}} & \multirow[c]{2}{*}{\texttt{user-item-purchase}} & Val & $0.36$ & $1.07$ & $0.44_{\pm 0.03}$ & $0.92_{\pm 0.04}$ & $\bm{2.64}_{\pm 0.00}$ \\
 &  & Test & $0.30$ & $0.89$ & $0.38_{\pm 0.02}$ & $0.80_{\pm 0.03}$ & $\bm{2.81}_{\pm 0.01}$ \\
\cmidrule{1-8}
\multirow[c]{4}{*}{\texttt{rel-stack}} & \multirow[c]{2}{*}{\texttt{user-post-comment}} & Val & $0.03$ & $2.05$ & $0.04_{\pm 0.02}$ & $0.43_{\pm 0.08}$ & $\bm{15.17}_{\pm 0.15}$ \\
 &  & Test & $0.02$ & $1.42$ & $0.04_{\pm 0.03}$ & $0.11_{\pm 0.05}$ & $\bm{12.72}_{\pm 0.22}$ \\
\cmidrule{2-8}
 & \multirow[c]{2}{*}{\texttt{post-post-related}} & Val & $0.47$ & $0.00$ & $1.62_{\pm 0.36}$ & $0.00_{\pm 0.01}$ & $\bm{7.76}_{\pm 0.20}$ \\
 &  & Test & $1.46$ & $1.74$ & $2.00_{\pm 0.43}$ & $0.07_{\pm 0.08}$ & $\bm{10.83}_{\pm 0.22}$ \\
\cmidrule{1-8}
\multirow[c]{4}{*}{\texttt{rel-trial}} & \multirow[c]{2}{*}{\texttt{condition-sponsor-run}} & Val & $2.63$ & $8.58$ & $4.88_{\pm 0.13}$ & $3.12_{\pm 0.24}$ & $\bm{11.33}_{\pm 0.04}$ \\
 &  & Test & $2.52$ & $8.42$ & $4.82_{\pm 0.20}$ & $2.89_{\pm 0.39}$ & $\bm{11.36}_{\pm 0.08}$ \\
\cmidrule{2-8}
 & \multirow[c]{2}{*}{\texttt{site-sponsor-run}} & Val & $4.91$ & $15.90$ & $10.92_{\pm 0.67}$ & $14.09_{\pm 0.77}$ & $\bm{17.43}_{\pm 0.07}$ \\
 &  & Test & $3.75$ & $17.31$ & $8.40_{\pm 0.70}$ & $10.70_{\pm 1.10}$ & $\bm{19.00}_{\pm 0.12}$ \\}
\begin{tabular}{lllrrrrr}
\toprule
\textbf{Dataset} & \textbf{Task} & \textbf{Split} &
\makecell{\textbf{Global}\\\textbf{Popularity}} & \makecell{\textbf{Past}\\\textbf{Visit}} &
\textbf{LightGBM} &
\makecell{\textbf{RDL}\\\textbf{(GraphSAGE)}} & \makecell{\textbf{RDL}\\\textbf{(ID-GNN)}} \\
\midrule
\multirow[c]{9}{*}{\amazon} & \multirow[c]{3}{*}{\userItemPurchase} 
 & Train & 11,759,844 & 2.18 & - \\
 &  & Val & 802,540 & 2.28 & 0.18 \\
 &  & Test & 918,919 & 2.33 & 0.15 \\

\cmidrule{2-6}

 & \multirow[c]{3}{*}{\userItemRate} 
 & Train & 7,146,115 & 1.8 & - \\
 &  & Val & 519,496 & 2.01 & 0.19 \\
 &  & Test & 599,867 & 2.05 & 0.15 \\

 \cmidrule{2-6}

 & \multirow[c]{3}{*}{\userItemReview} 
 & Train & 5,138,184 & 2.19 & - \\
 &  & Val & 268,651 & 2.3 & 0.18 \\
 &  & Test & 305,476 & 2.4 & 0.15 \\

\cmidrule{1-6}

\multirow[c]{3}{*}{\handm} & \multirow[c]{3}{*}{\userItemPurchase} & 
Train & 13,191,321 & 3.38 & - \\
 &  & Val & 237,152 & 3.18 & 3.51 \\
 &  & Test & 207,996 & 3.10 & 3.76 \\

% \cmidrule{1-6}
% \multirow[c]{3}{*}{\fone} & \multirow[c]{3}{*}{\driverConstructorResult}
% & Train & 1900 &  & - \\
%  &  & Val & 142 & 1.95 & 9.86 \\
%  &  & Test & 105 & 1.84 & 10.48 \\

\cmidrule{1-6}

\multirow[c]{6}{*}{\stackex} & \multirow[c]{3}{*}{\userPostComment}
 & Train & 43,337 & 2.08 & - \\
 & & Val & 1,603 & 1.94 & 3.43 \\
 &  & Test & 1,517 & 2.0 & 4.09 \\

\cmidrule{2-6}

 & \multirow[c]{3}{*}{\postPostLinked}
 & Train & 7,162 & 1.2 & - \\
 &  & Val & 294 & 1.3 & 0.0 \\
 &  & Test & 359 & 1.39 & 1.39 \\

\cmidrule{1-6}

\multirow[c]{6}{*}{\trials} & \multirow[c]{3}{*}{\sponsorConditionRec}
 & Train & 503,176 & 12.51 & - \\
 &  & Val & 30,448 & 14.63 & 34.48 \\
 &  & Test & 25,694 & 12.49 & 38.37 \\

\cmidrule{2-6}

& \multirow[c]{3}{*}{\sponsorFacilityRec}
 & Train & 1,485,360 & 2.27 & - \\
 &  & Val & 80,103 & 2.16 & 20.91 \\
 &  & Test & 50,635 & 1.85 & 23.29 \\

\cmidrule{1-6}

\multirow[c]{3}{*}{\avito} & \multirow[c]{3}{*}{\userAdVisit} & 
Train & 2,738,733 & 31.53 & - \\
 &  & Val & 877,441 & 29.27 & 6.79 \\
 &  & Test & 712,985 & 19.73 & 4.73 \\
\bottomrule
\end{tabular}
\label{tab:app_link}
\end{table}

\subsection{Detailed Results}

Tables \ref{tab:app_classif}, \ref{tab:app_regression} and \ref{tab:app_link} show mean and standard deviations over $5$ runs for the entity classification, entity regression and link prediction results respectively.

\subsection{Hyperparameter Choices}

All our RDL experiments were run based on a single set of default task-specific hyperparameters, \emph{i.e.} we did not perform exhaustive hyperparamter tuning, \emph{cf.}~Table~\ref{tab:hparams}.
This verifies the stability and robustness of RDL solutions, even against expert data scientist baselines. Specifically, all task types use a shared GNN configuration (a two-layer GNN with a hidden feature size of $128$ and ``sum'' aggregation) and sample subgraphs identically (disjoint subgraphs of $512$ seed entities with a maximum of $128$ neighbors for each foreign key).
Across task types, we only vary the learning rate and the maximum number of epochs to train for.

\begin{table}[t]
  \centering
  \caption{{Task-specific RDL default hyperparameters.}  }
  \label{tab:hparams}
  \renewcommand{\arraystretch}{1.1}
  \setlength{\tabcolsep}{3pt}
  \scriptsize
  \begin{tabular}{lrrr}
    \toprule
      \mr{2}{\textbf{Hyperparameter}} & \mc{3}{c}{\textbf{Task type}} \\
      & Node classification & Node regression & Link prediction \\
    \midrule
      Learning rate & 0.005 & 0.005 & 0.001 \\
      Maximum epochs & 10 & 10 & 20 \\
      Batch size & 512 & 512 & 512 \\
      Hidden feature size & 128 & 128 & 128 \\
      Aggregation & summation & summation & summation \\
      Number of layers & 2 & 2 & 2 \\
      Number of neighbors & 128 & 128 & 128 \\
      Temporal sampling strategy & uniform & uniform & uniform \\
   \bottomrule
  \end{tabular}
\end{table}

Notably, we found that our default set of hyperparameters heavily underperformed on the node-level tasks on the \trials dataset. On this dataset, we used a learning rate of $0.0001$, a ``mean'' neighborhood aggregation scheme, $64$ sampled neighbors, and trained for a maximum of $20$ epochs.
For the ID-GNN link-prediction experiments on \trials, it was necessary to use a four-layer deep GNN in order to ensure that destination nodes are part of source node-centric subgraphs.

\subsection{Ablations}

We also report additional results ablating parts of our relational deep learning implementation. All experiments are designed to be data-centric, aiming to validate basic properties of the chosen datasets and tasks. Examples include confirming that the graph structure, node features, and temporal-awareness all play important roles in achieving optimal performance, which also underscores the unique challenges our~\relbench dataset and tasks present.

\begin{figure}[t!]
    \centering
    \includegraphics[width=1.\linewidth]{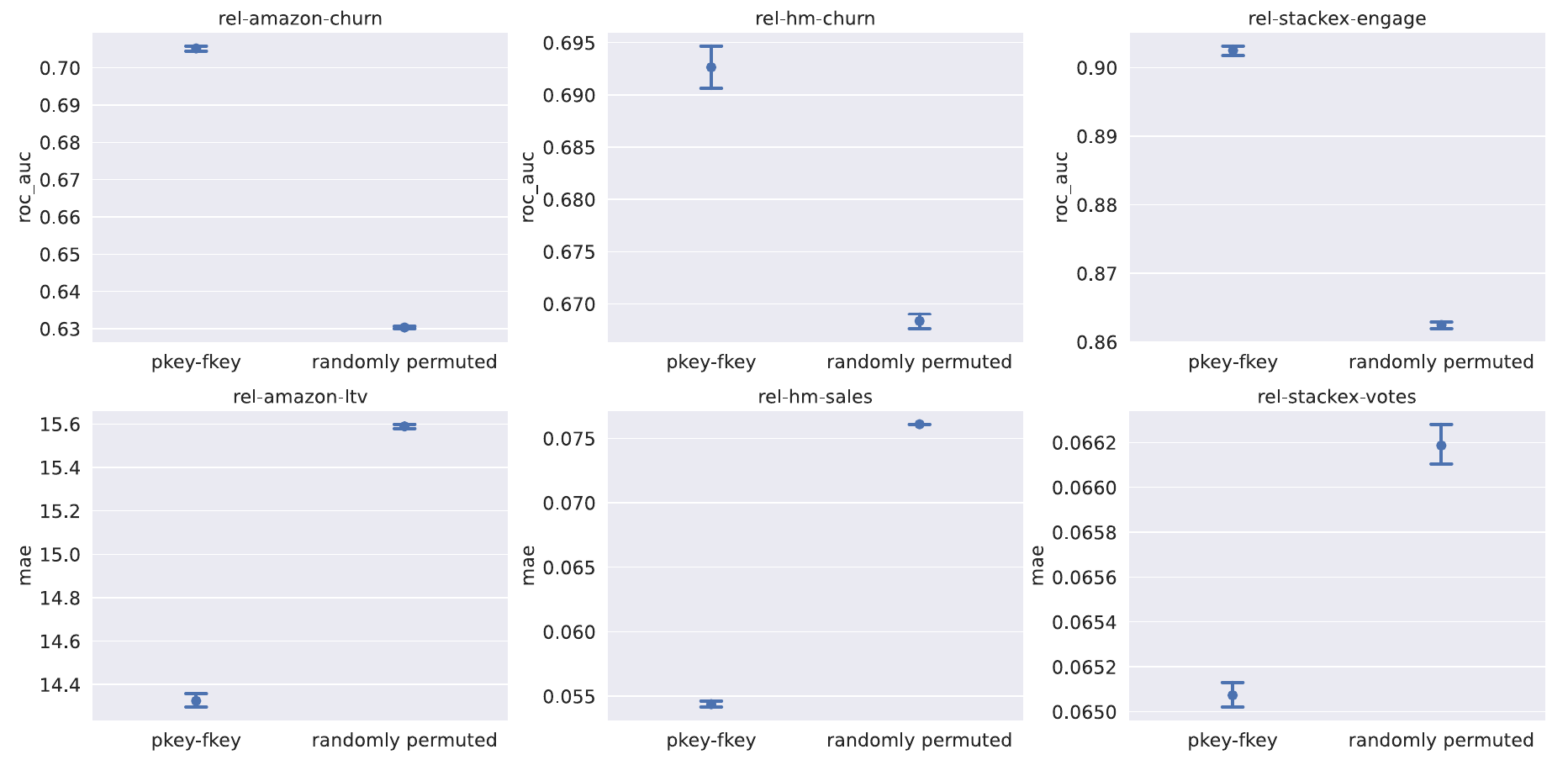}
    \caption{Investigation on the role of leveraging primary-foreign key (pkey-fkey) edges for the GNN. At the top row are three node classification tasks with metric AUROC (higher is better) while at the bottom are three node regression tasks with metric MAE (lower is better), evaluated on the test set. We find that our proposal of using pkey-fkey edges for message passing is vital for GNN to achieve desirable performance on~\relbench. Error bars correspond to 95\% confidence interval.}
    \label{fig:ablation-edge}
\end{figure}

\textbf{Graph structure.} We first investigate the role of the graph structure we adopt for GNNs on~\relbench. Specifically, we compare the following two approaches of constructing the edges: \textbf{1.} \emph{Primary-foreign key (pkey-fkey)}, where the entities from two tables that share the same primary key and foreign key are connected through an edge; \textbf{2.} \emph{Randomly permuted}, where we apply a random permutation on the destination nodes in the primary-foreign key graph for each type of the edge while keeping the source nodes untouched. From Fig.~\ref{fig:ablation-edge} we observe that with random permutation on the primary-foreign key edges the performance of the GNN becomes much worse, verifying the critical role of carefully constructing the graph structure through, \eg, primary-foreign key as proposed in~\citet{fey2023relational}.

\begin{figure}[t!]
    \centering
    \includegraphics[width=1.\linewidth]{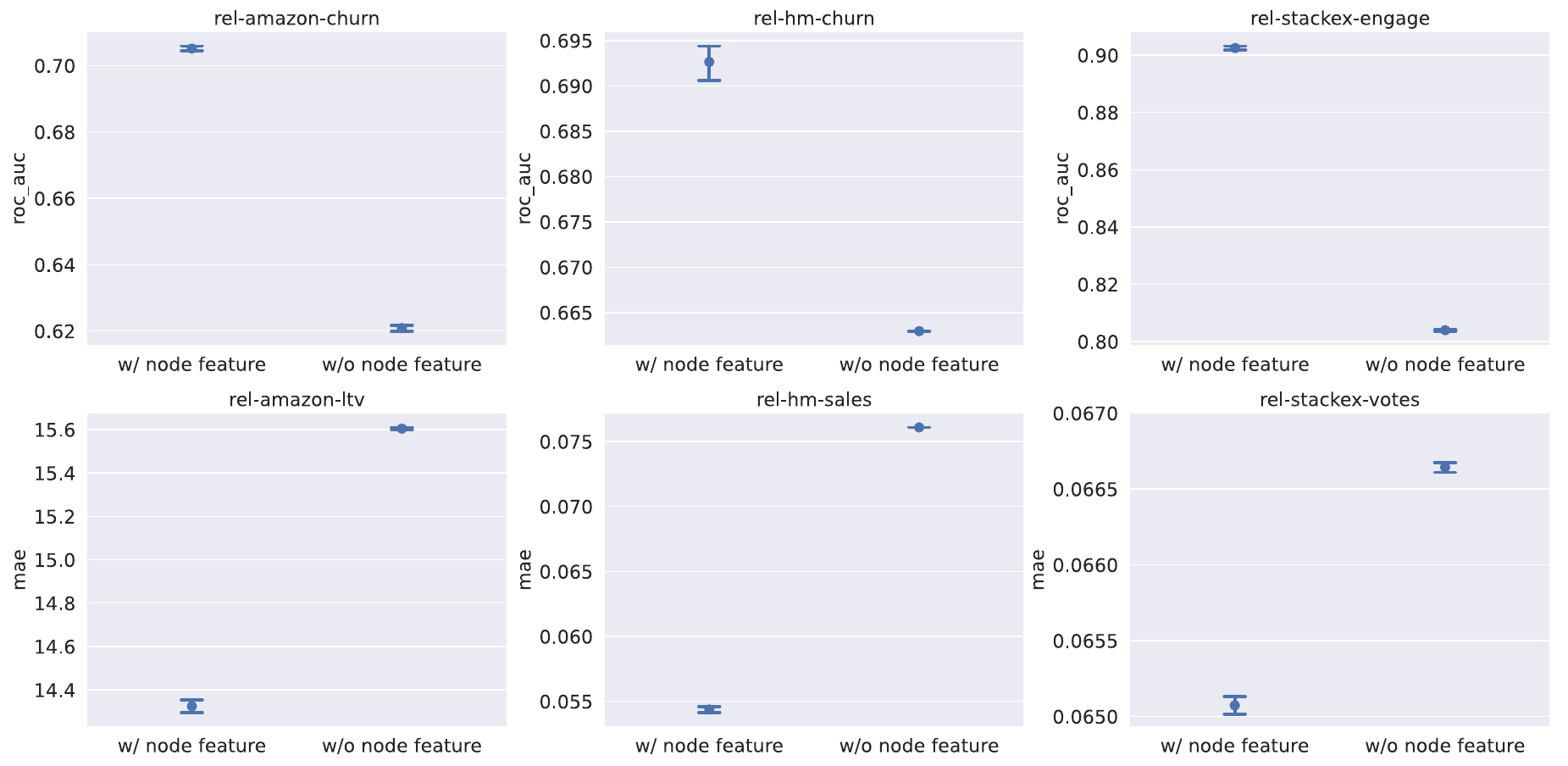}
    \caption{Investigation on the role of node features. At the top row are three node classification tasks with metric AUROC (higher is better) while at the bottom are three node regression tasks with metric MAE (lower is better), evaluated on the test set. We observe that leveraging node features is important for GNN. Error bars correspond to 95\% confidence interval.}
    \label{fig:ablation-nodefeature}
\end{figure}

\begin{figure}[t!]
    \centering
    \includegraphics[width=1.\linewidth]{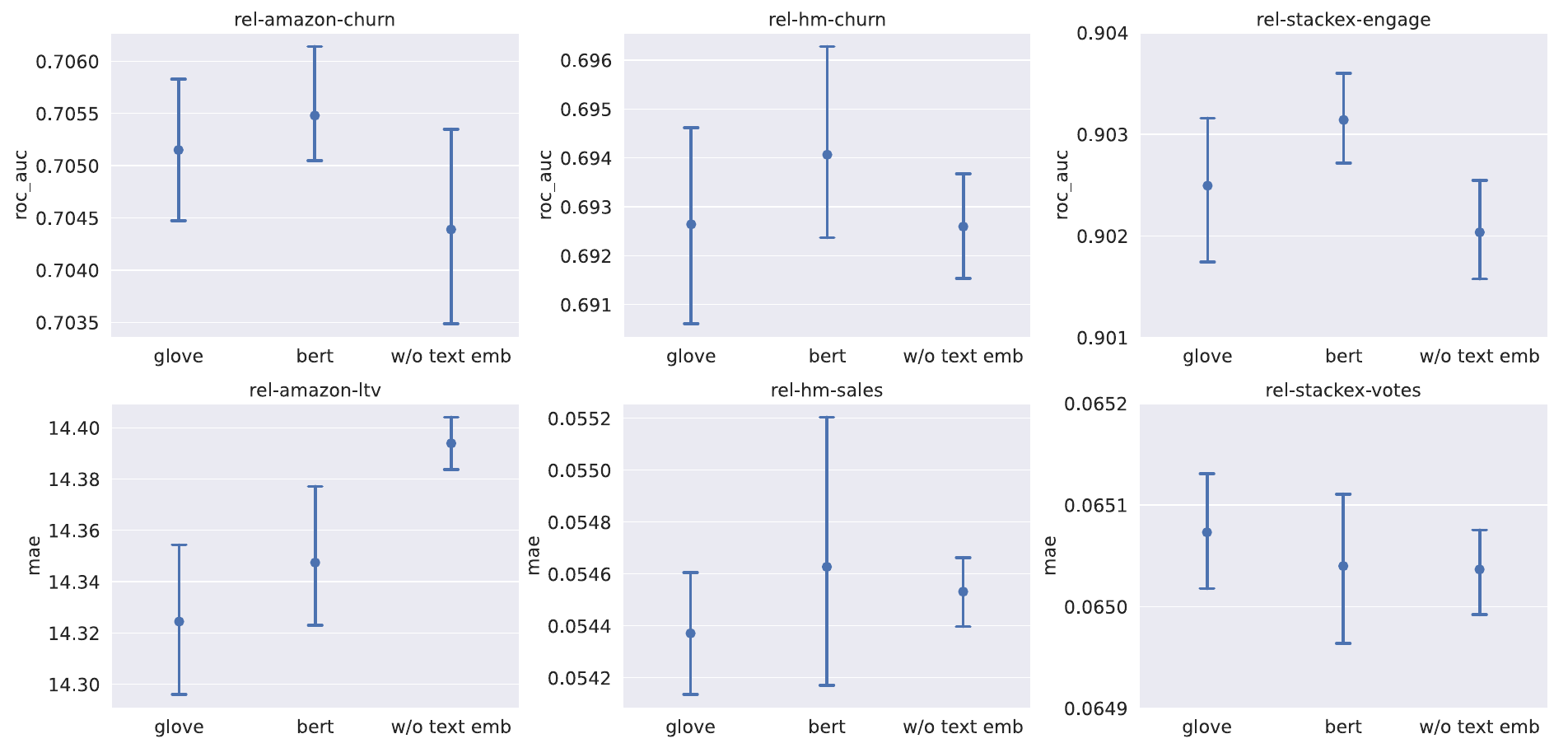}
    \caption{Investigation on the role of text embedding. At the top row are three node classification tasks with metric AUROC (higher is better) while at the bottom are three node regression tasks with metric MAE (lower is better), evaluated on the test set. We observe that adding text embedding using GloVe~\citep{pennington2014glove} or BERT~\citep{devlin2018bert} generally helps improve the performance. Error bars correspond to 95\% confidence interval.}
    \label{fig:ablation-textemb}
\end{figure}

\textbf{Node features and text embeddings.} Here we study the effect of node features used in~\relbench. In the experiments depicted in Fig.~\ref{fig:ablation-nodefeature}, we compare GNN (w/ node feature) with its variant where the node features are all masked by zeros (\ie, w/o node feature). We find that utilizing rich node features incorporated in our~\relbench dataset is crucial for GNN. Moreover, we also investigate, in particular, the approach to encode texts in the data that constitutes part of the node features. In Fig.~\ref{fig:ablation-textemb}, we compare GloVe text embedding~\citep{pennington2014glove} and BERT text embedding~\citep{devlin2018bert} with w/o text embedding, where the text embeddings are masked by zeros. We observe that encoding the rich texts in~\relbench with GloVe or BERT embedding consistently yields better performance compared with using no text features. We also find that BERT embedding is usually better than GloVe embedding especially for node classification tasks, which suggests that enhancing the quality of text embedding will potentially help achieve better performance.

\begin{figure}[t!]
    \centering
    \includegraphics[width=1.\linewidth]{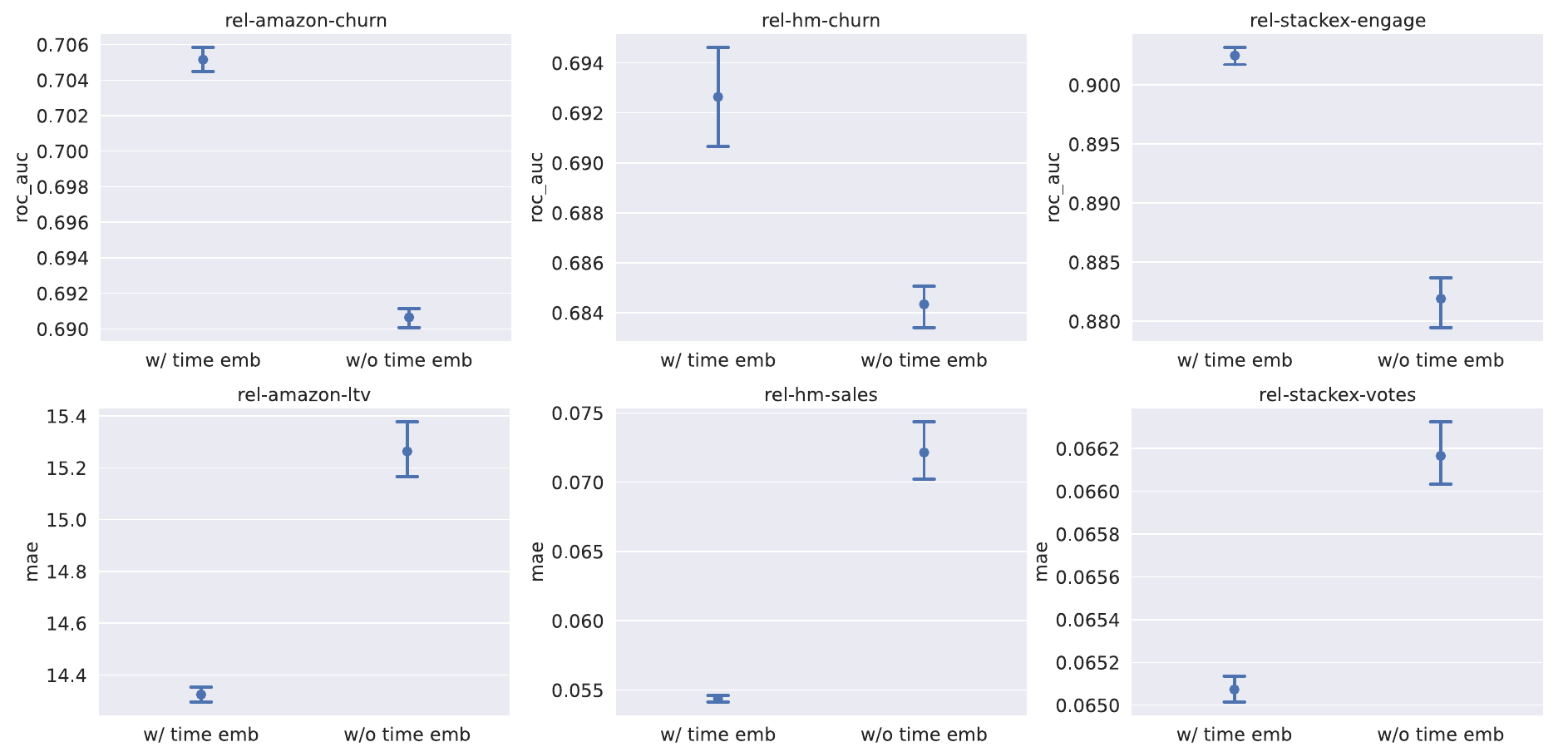}
    \caption{Investigation on the role of time embedding. At the top row are three node classification tasks with metric AUROC (higher is better) while at the bottom are three node regression tasks with metric MAE (lower is better), evaluated on the test set. We find that adding time embedding to the GNN consistently boosts the performance. Error bars correspond to 95\% confidence interval.}
    \label{fig:ablation-timeemb}
\end{figure}

\textbf{Temporal awareness.} We also investigate the importance of injecting temporal awareness into the GNN by ablating on the time embedding. To be specific, in the implementation we add a relative time embedding when deriving the node features using the relative time span between the timestamp of the entity and the querying seed time. Results are exhibited in Fig.~\ref{fig:ablation-timeemb}. We discover that adding the time embedding significantly enhance the performance across a diverse range of tasks, demonstrating the efficacy and importance of building up the temporal awareness into the model.

\section{User Study Additional Details}\label{app: user study}

\subsection{Data Scientist Example Workflow}

In this section we provide a detailed description of the data scientist workflow for the \userChurn task of the \handm dataset. The purpose of this is to exemplify the efforts undertaken by the data scientist to solve \relbench tasks. For data scientist solutions to all tasks, see \userstudy.

Recall that the main data science workflow steps are:
\begin{enumerate}
    \item Exploratory data analysis (EDA).
    \item Feature ideation.
    \item Feature enginnering. 
    \item Tabular ML.
    \item Post-hoc analysis of feature importance (optional).
\end{enumerate}

\subsubsection{Exploratory Data Analysis}

\begin{figure}[t]

\centering
\hspace{-40pt}
\includegraphics[width=.36\textwidth]{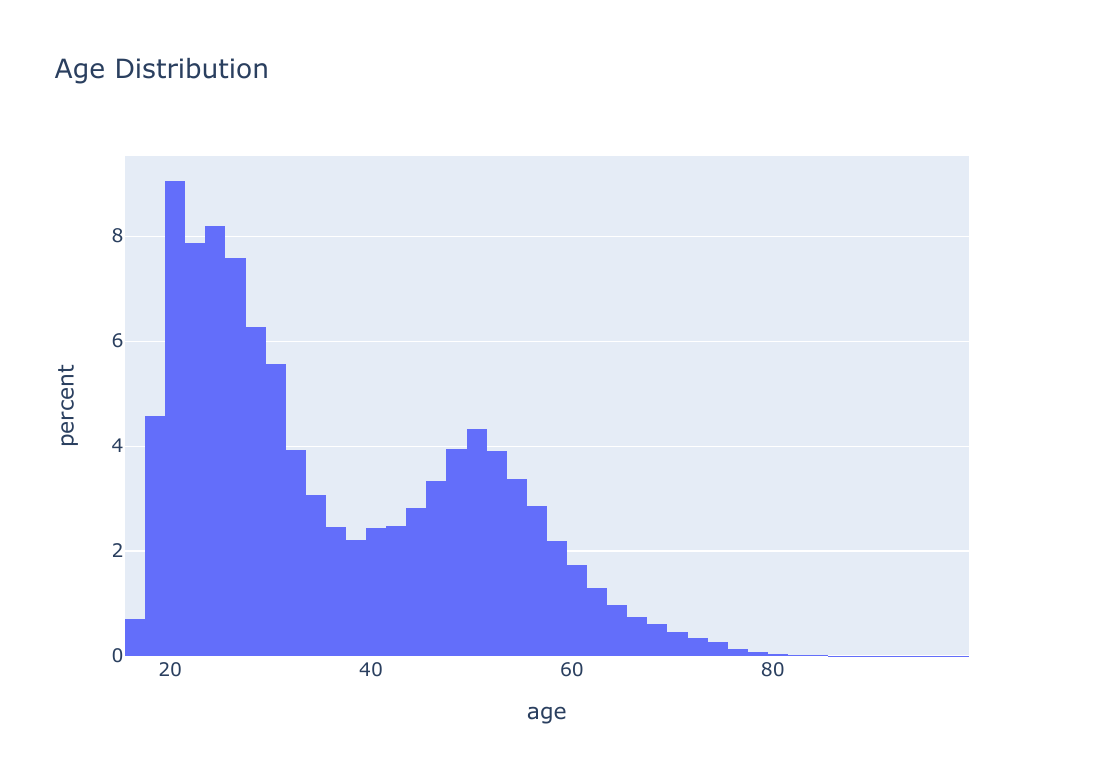}\hfill
\includegraphics[width=.36\textwidth]{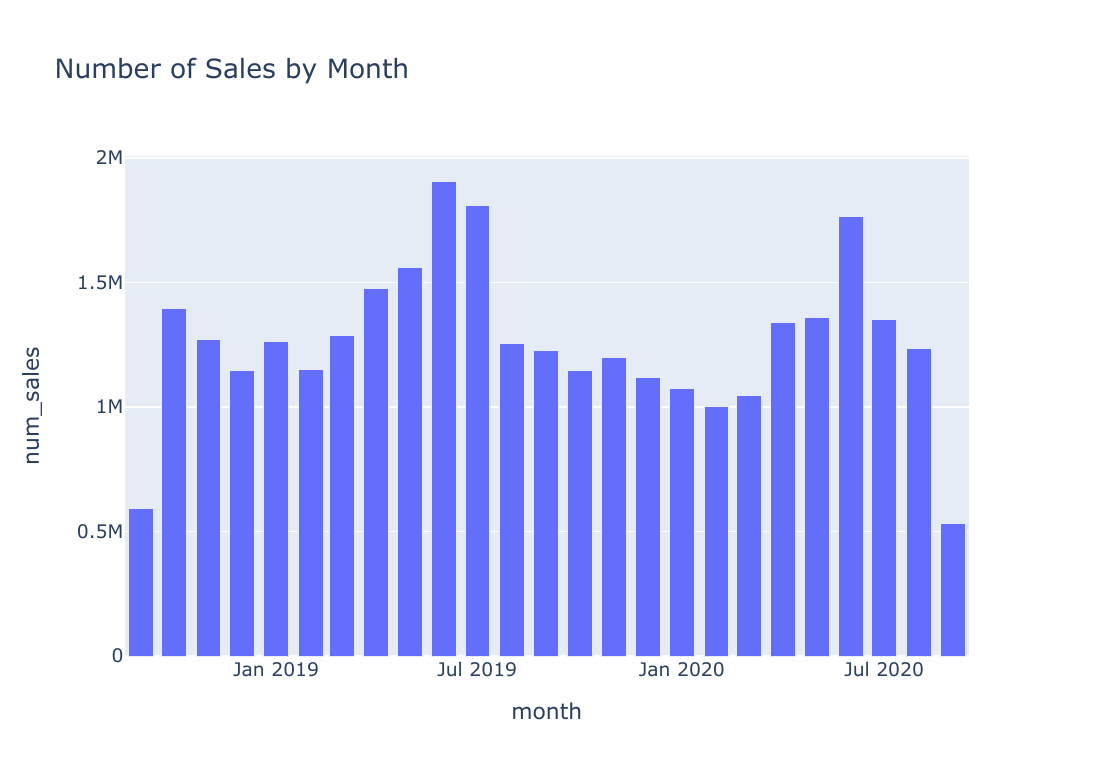}\hfill
\includegraphics[width=.36\textwidth]{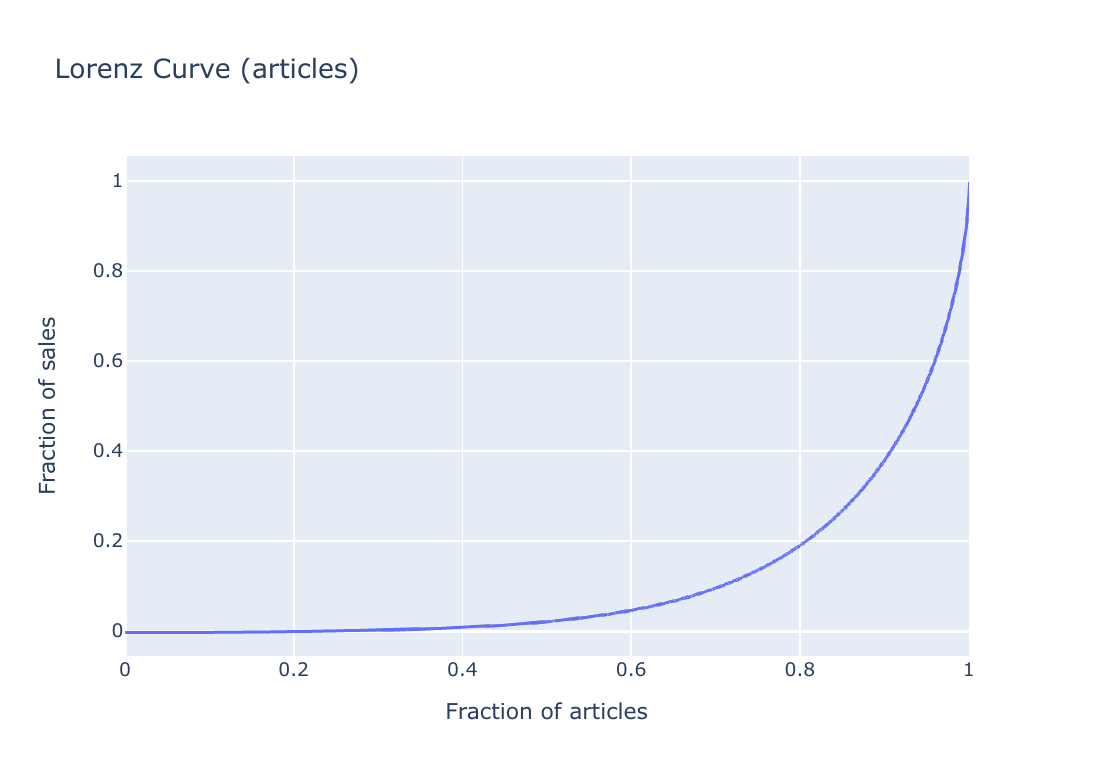}
\hspace{-40pt}

\caption{\textbf{EDA Plots.} Each plot explores different characteristics of the dataset. Understanding the data and identifying relationships between different quantities is an essential prerequisite to meaningful feature engineering.}
\label{fig:feats}

\end{figure}

During the exploratory data analysis (EDA) the data scientist familiarizes themselves with a new dataset. It is typically carried out in a Jupyter notebook, where the data scientist first loads the dataset or establishes a connection to it and then systematically explores it. The data scientist may:
\begin{itemize}
    \item Visualize the database schema, looking at the fields of different tables and the relationships between them.
    \item Closely analyze the label sets:
        \begin{itemize}
            \item Look at the relative sizes and temporal split of the training, validation and test subsets.
            \item Look at label statistics such as the mean, the standard deviation and various quantiles. 
            \item For classification tasks, understand class (im)balance: how much bigger is the modal class than the rest? For example, in the \userChurn task roughly 82\% of the samples have label $1$, so there is a good amount of imbalance but not enough to strictly require up-sampling techniques.
            \item For regression tasks, understand the label distribution: are the labels concentrated around a typical value or do they follow a power law wherein the labels span several orders of magnitude? In extreme cases, this exploration will point to a need for specialized handing of the label space for model training.
        \end{itemize}
    \item Plot distributions and aggregations of interesting columns/fields. For example, in Figure \ref{fig:feats} we can see three such plots. From left to right:
    \begin{itemize}
        \item The first plot shows the distribution of age among customers. We see two distinct peaks one in the mid-twenties and another in the mid-fifties, suggesting different customer ``archetypes'', which may have different spending patterns.
        \item The second plot shows the number of sales per month over a two year period. We can see some seasonality with summer months being particularly good for overall sales. This suggests date related features could be useful.
        \item The third plot shows a \emph{Lorenz curve} of sales per article, showcasing the canonical \emph{Pareto Principle}: 20\% of the articles account for 80\% of the sales.
    \end{itemize}
    \item Run custom queries to look at interesting quantities and/or relationships between different columns. For instance, in the EDA for \handm, an interesting quantity to look at is the variability in item prices across the year. This reveals that most of the variability is downward, representing temporary discounts.
    \item Investigate outliers or odd-looking patterns in the data. These usually will have some real-world explanation that may inform how the data scientist chooses to pre-process the data and construct features.
\end{itemize}

In all, this process takes in the order of a few hours (3-4 for most datasets in the user study).

\subsubsection{Feature Ideation}

Having explored the dataset in the EDA, the data scientist will then brainstorm features that, to their judgement, will provide valuable signal to a model for a specific learning task. In the case of the \userChurn task, a rather simple feature would be the customer's age, which is a field directly available in one of the tables. A slightly more complex feature would be the total amount spent by the customer so far. Finally, an example of a fairly complex feature is the average monthly sales volume of items purchased by the customer in the past week. A high value for this feature may indicate that the customer has been shopping trendy items lately, whereas a low value for this feature may indicate that the customer has been interested in more arcane or specific items.

In practice, the ideation phase consists of writing down all of these feature ideas in a file or a piece of paper. It is the quickest part of the whole process and in this user study took between 30 minutes and one hour.

\subsubsection{Feature Engineering}

With a list of features in hand, the data scientist then proceeds to actually write code to generate all the features for each sample in the the train, validation and test subsets. In this user study, this was carried out using DuckDB SQL\footnote{See \url{https://duckdb.org/}.} with some Jinja templating\footnote{See \url{https://jinja.palletsprojects.com/en/3.1.x/intro/}.} for convenience.

Revisiting the example features from the previous section, the conceptual complexity of the features closely tracks with the technical complexity of implementing them. For customer age all that is required is a simple \textit{join}. The total amount spent by the customer, can be calculated using a \textit{group by} clause and a couple of \textit{join}'s. Lastly, calculating the average monthly sales volume of items purchased by the customer in the past week requires multiple \textit{group by}'s, \textit{join}'s, and \textit{window functions} distributed across multiple \textit{common table expressions} (CTEs).

A key consideration during feature engineering is the prevention of \textit{leakage}. The data scientist must ensure that none of the features accidentally include information from after the sample timestamp. This is especially true for complex features like the third example above, where special care must be taken to ensure that each \textit{join} has the appropriate filters to comply with the sample timestamp. 

For some tasks, \emph{e.g.}, \studyOutcome, the initial features \emph{did} leak information from the validation set into the training set. Thanks to the \relbench testing setup, leaking test data into the training data is hard to do by accident, since test data is hidden. Leaking information from validation to train (but not test to train) led to extremely high validation performance and very low test performance (test was significantly lower than LightGBM with no feature engineering). The large discrepancy between validation and test performances alerted the data scientist to the mistake, and the features were eventually fixed. This example illustrates another complexity that feature engineering introduces, with special care needed to ensure leakage does not happen. 

Other considerations that the data scientist must keep in mind during development and implementation of the features are parsing issues, runtime constrains and memory load. For example, during the user study we identified a parsing issue arising from special characters in user posts/comments in the \stackex dataset. The \textit{backslash} character, widely used \LaTeX can trip up certain text parsers if not handled with care. Furthermore, runtime and memory constraints are important to keep in mind when working with larger datasets and computing features that require nested \textit{join}'s and aggregations. During the user study, there were some cases where we had to refactor SQL queries to make them more efficient, increasing the overall implementation time. For some tasks we had to implement sub-sampling of the training set to reduce the burden on compute resources.

Finally, once the features have been generated for each data subset, the data scientist will usually inspect the generated features looking for anomalies (e.g. an unusual prevalence of \textit{NULL} values). In this user study we also implemented some automated sanity checks to validate the generated features beyond manual inspection.

\subsubsection{Tabular Machine Learning}

The output of the Feature Engineering phase is a DuckDB table with engineered features for each data subset. There is some non-trivial amount of work required to go from those tables to the numerical arrays used for training by most Tabular ML models (LightGBM in this case). This is implemented in a Python script that loads the data, transforms it into arrays and carries out hyperparameter tuning. In this user study we ran 5 hyperparameter optimization runs, with 10 trials each, reporting the mean and standard deviation over the 5 runs. For the \userChurn task this took one to two hours.

\subsubsection{Post-hoc Analysis}

\begin{figure}[t]

\centering
\includegraphics[width=200px]{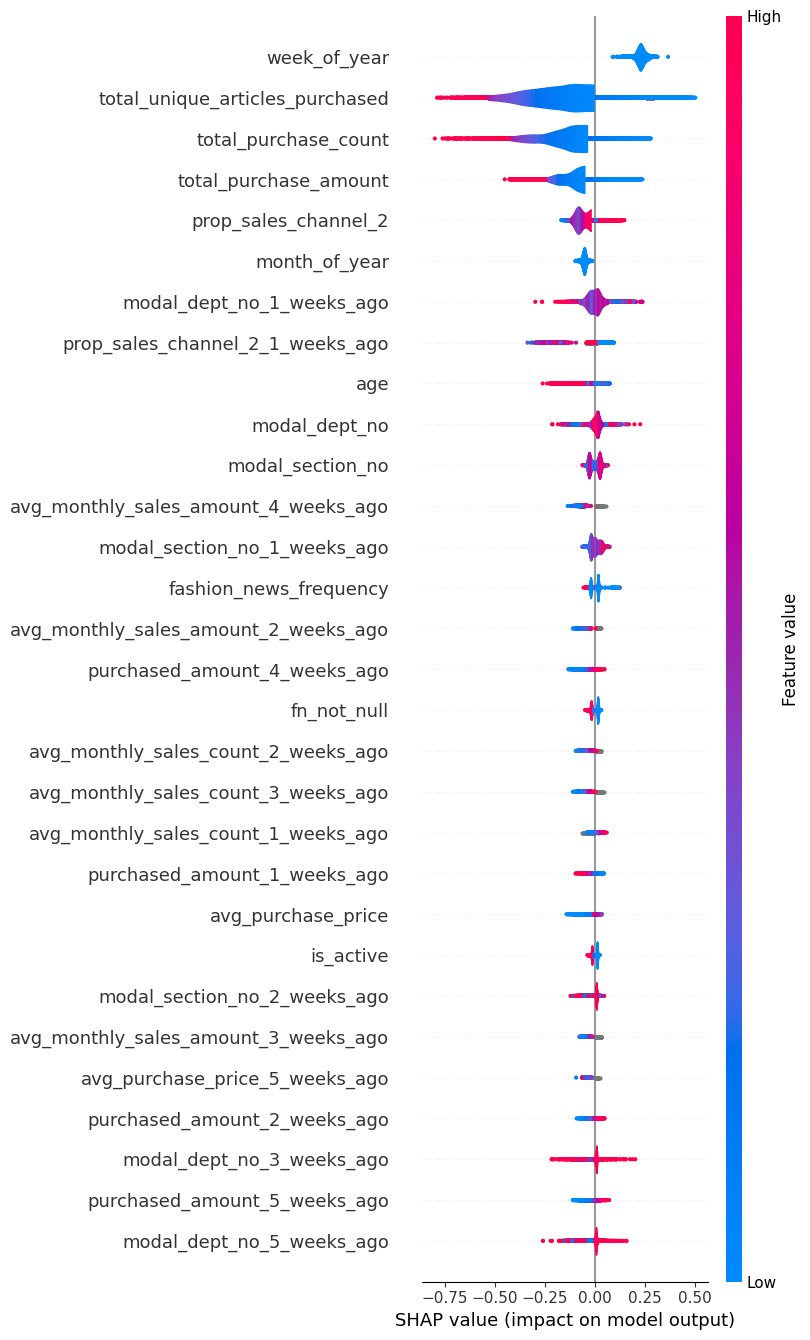}
% \vspace{-30pt}
\caption{\textbf{Feature Importances.} SHAP values of top 30 features ranked by importance. Note: \textit{week\_of\_year} feature shows little variability because the validation set is temporally concentrated in a few weeks.}
% \vspace{-5pt}
\label{fig:shap}
\end{figure}

The last step in the process is to look at a trained model and analyze its performance and feature importance. To this end we used SHAP values \citep{lundberg2017unified} and the corresponding python package\footnote{See \url{https://shap.readthedocs.io/en/latest/}.}. Figure \ref{fig:shap} shows the top 30 most important features in the \userChurn task. The individual \textit{violin plots} show the distribution of SHAP values for a subset of the validation set, the color indicates the value of the feature. For the \userChurn task, the most predictive features were primarily (1) all-time statistics of user behavior pattern, and (2) temporal information that allows the model to be aware of seasonality.

\subsection{Regression Output Head Analysis}

By default, our RDL implementation uses a simple linear output head on top of the GNN embeddings. However we found that on regression tasks this sometimes led to lower than desirable performance. We found that performance on many regression tasks could be improved by modifying this output head. Instead of a linear layer, we took the output from the GNN, and fed these embeddings into a LightGBM model, which is trained in a second separate training phase from the GNN model.

The resulting model still uses an end-to-end learned GNN for cross-table feature engineering, showing that the GNN is learning useful features. Instead we attribute the weaker performance to the linear output head. We believe that further attention to the regression output head is an interesting direction for further study, with the goal of designing an output head that is performant and can be trained jointly with the GNN (unlike our LightGMB modification).

We run three experiments to study this phenomena, and attempt to isolate the output head as a problematic component for regression tasks.
\begin{enumerate}
    \item LightGBM trained on GNN-learned entity-level features on regression tasks. We find that this model performs better than the original GNN, suggesting that the linear output head of the GNN is suboptimal.
    \item LightGBM trained on GNN-learned entity-level features on classification tasks. We find no performance improvement, and even some degradation, compared to the original GNN model, suggesting that the observed performance boost of (1) comes not from an overall better architecture but from the correction of an innate shortcoming of the linear output head vis-a-vis regression tasks. In other words, using a LightGBM on top of the GNN is only helpful insofar as it provides a more flexible output head for regression tasks.
    \item Evaluate GNN performance after converting regression tasks to binary classification tasks with label $y = \mathbf{1}\{y_\text{regression} > 0\}$. We find that the performance gap between the data scientist models and the GNN narrow. This suggests that the GNN can learn the relevant predictive signal, but performance is affected by how the task is formulated (classification vs regression).
\end{enumerate}
See Tables \ref{tab:regresion analysis expt 1}, \ref{tab:regresion analysis expt 2}, \ref{tab:regresion analysis expt 3} for the results of each of these experiments. 

In Figure \ref{fig:ds main fig}, for regression tasks we report the RDL results using GNN learned features with LightGBM output head. In Table \ref{tab: regression} we report result for the basic GNN in order to avoid creating confusion for other researchers when comparing different GNN methods. We believe that Tables \ref{tab:regresion analysis expt 1}, \ref{tab:regresion analysis expt 2}, \ref{tab:regresion analysis expt 3} provide clear evidence that there is an opportunity for improvements and simplifications, which we leave to future work.

\begin{table}[t]
    \vspace{-3mm}
    \caption{Entity regression results (MAE, lower is better) on selected \relbench datasets. Training a LightGBM model on features extracted by a trained GNN leads to performance lift. This is evidence that the linear layer output head of the base GNN is suboptimal.}
    \centering
    \setlength{\tabcolsep}{3pt}
    \scriptsize
    \newcommand{\tabledata}{\fone & \driverPosition & Test & $4.173_{\pm 0.178}$ & $\bm{4.05}_{\pm 0.09}$ \\
\cmidrule{1-5}
\stackex  & \postVotes & Test & $0.065_{\pm 0.00}$ & $\bm{0.062}_{\pm 0.00}$ \\
\cmidrule{1-5}
\amazon & \itemLtv & Test & $14.31_{\pm 0.028}$ & $\bm{14.10}_{\pm 0.02}$ \\
}
    \begin{tabular}{llrrr}
    \toprule
    \textbf{Dataset} & \textbf{Task} & \textbf{Split} & \textbf{GNN} & \textbf{GNN+LightGBM} \\
    \midrule
    \multirow[c]{9}{*}{\amazon} & \multirow[c]{3}{*}{\userItemPurchase} 
 & Train & 11,759,844 & 2.18 & - \\
 &  & Val & 802,540 & 2.28 & 0.18 \\
 &  & Test & 918,919 & 2.33 & 0.15 \\

\cmidrule{2-6}

 & \multirow[c]{3}{*}{\userItemRate} 
 & Train & 7,146,115 & 1.8 & - \\
 &  & Val & 519,496 & 2.01 & 0.19 \\
 &  & Test & 599,867 & 2.05 & 0.15 \\

 \cmidrule{2-6}

 & \multirow[c]{3}{*}{\userItemReview} 
 & Train & 5,138,184 & 2.19 & - \\
 &  & Val & 268,651 & 2.3 & 0.18 \\
 &  & Test & 305,476 & 2.4 & 0.15 \\

\cmidrule{1-6}

\multirow[c]{3}{*}{\handm} & \multirow[c]{3}{*}{\userItemPurchase} & 
Train & 13,191,321 & 3.38 & - \\
 &  & Val & 237,152 & 3.18 & 3.51 \\
 &  & Test & 207,996 & 3.10 & 3.76 \\

% \cmidrule{1-6}
% \multirow[c]{3}{*}{\fone} & \multirow[c]{3}{*}{\driverConstructorResult}
% & Train & 1900 &  & - \\
%  &  & Val & 142 & 1.95 & 9.86 \\
%  &  & Test & 105 & 1.84 & 10.48 \\

\cmidrule{1-6}

\multirow[c]{6}{*}{\stackex} & \multirow[c]{3}{*}{\userPostComment}
 & Train & 43,337 & 2.08 & - \\
 & & Val & 1,603 & 1.94 & 3.43 \\
 &  & Test & 1,517 & 2.0 & 4.09 \\

\cmidrule{2-6}

 & \multirow[c]{3}{*}{\postPostLinked}
 & Train & 7,162 & 1.2 & - \\
 &  & Val & 294 & 1.3 & 0.0 \\
 &  & Test & 359 & 1.39 & 1.39 \\

\cmidrule{1-6}

\multirow[c]{6}{*}{\trials} & \multirow[c]{3}{*}{\sponsorConditionRec}
 & Train & 503,176 & 12.51 & - \\
 &  & Val & 30,448 & 14.63 & 34.48 \\
 &  & Test & 25,694 & 12.49 & 38.37 \\

\cmidrule{2-6}

& \multirow[c]{3}{*}{\sponsorFacilityRec}
 & Train & 1,485,360 & 2.27 & - \\
 &  & Val & 80,103 & 2.16 & 20.91 \\
 &  & Test & 50,635 & 1.85 & 23.29 \\

\cmidrule{1-6}

\multirow[c]{3}{*}{\avito} & \multirow[c]{3}{*}{\userAdVisit} & 
Train & 2,738,733 & 31.53 & - \\
 &  & Val & 877,441 & 29.27 & 6.79 \\
 &  & Test & 712,985 & 19.73 & 4.73 \\
    \bottomrule
    \end{tabular}
    \label{tab:regresion analysis expt 1}
\end{table}

\begin{table}[t]
    \vspace{-3mm}
    \caption{Entity classification results (AUROC, higher is better, numbers bolded if withing standard deviation of best result) on selected \relbench tasks. Training a LightGBM model on features extracted by a trained GNN does not lead to performance lift, and can even hurt performance slightly. This is evidence that output head limitations hold for regression tasks only. Note, \studyOutcome uses default GNN parameters for simplicity, differing form the performance reported in the main paper.}
    \centering
    \setlength{\tabcolsep}{3pt}
    \scriptsize
    \newcommand{\tabledata}{\fone & \driverDNF & Test & $\bm{72.3}_{\pm 1.67}$ & $\bm{71.8}_{\pm 1.30}$  \\
\cmidrule{1-5}
\trials  & \studyOutcome & Test & $\bm{68.8}_{\pm 1.10}$ & $\bm{68.2}_{\pm 0.44}$ \\
\cmidrule{1-5}
\stackex & \userBadge & Test & $88.3_{\pm 0.04}$ & $\bm{88.4}_{\pm 0.04}$ \\
}
    \begin{tabular}{llrrr}
    \toprule
    \textbf{Dataset} & \textbf{Task} & \textbf{Split} & \textbf{GNN} & \textbf{GNN+LightGBM} \\
    \midrule
    \multirow[c]{9}{*}{\amazon} & \multirow[c]{3}{*}{\userItemPurchase} 
 & Train & 11,759,844 & 2.18 & - \\
 &  & Val & 802,540 & 2.28 & 0.18 \\
 &  & Test & 918,919 & 2.33 & 0.15 \\

\cmidrule{2-6}

 & \multirow[c]{3}{*}{\userItemRate} 
 & Train & 7,146,115 & 1.8 & - \\
 &  & Val & 519,496 & 2.01 & 0.19 \\
 &  & Test & 599,867 & 2.05 & 0.15 \\

 \cmidrule{2-6}

 & \multirow[c]{3}{*}{\userItemReview} 
 & Train & 5,138,184 & 2.19 & - \\
 &  & Val & 268,651 & 2.3 & 0.18 \\
 &  & Test & 305,476 & 2.4 & 0.15 \\

\cmidrule{1-6}

\multirow[c]{3}{*}{\handm} & \multirow[c]{3}{*}{\userItemPurchase} & 
Train & 13,191,321 & 3.38 & - \\
 &  & Val & 237,152 & 3.18 & 3.51 \\
 &  & Test & 207,996 & 3.10 & 3.76 \\

% \cmidrule{1-6}
% \multirow[c]{3}{*}{\fone} & \multirow[c]{3}{*}{\driverConstructorResult}
% & Train & 1900 &  & - \\
%  &  & Val & 142 & 1.95 & 9.86 \\
%  &  & Test & 105 & 1.84 & 10.48 \\

\cmidrule{1-6}

\multirow[c]{6}{*}{\stackex} & \multirow[c]{3}{*}{\userPostComment}
 & Train & 43,337 & 2.08 & - \\
 & & Val & 1,603 & 1.94 & 3.43 \\
 &  & Test & 1,517 & 2.0 & 4.09 \\

\cmidrule{2-6}

 & \multirow[c]{3}{*}{\postPostLinked}
 & Train & 7,162 & 1.2 & - \\
 &  & Val & 294 & 1.3 & 0.0 \\
 &  & Test & 359 & 1.39 & 1.39 \\

\cmidrule{1-6}

\multirow[c]{6}{*}{\trials} & \multirow[c]{3}{*}{\sponsorConditionRec}
 & Train & 503,176 & 12.51 & - \\
 &  & Val & 30,448 & 14.63 & 34.48 \\
 &  & Test & 25,694 & 12.49 & 38.37 \\

\cmidrule{2-6}

& \multirow[c]{3}{*}{\sponsorFacilityRec}
 & Train & 1,485,360 & 2.27 & - \\
 &  & Val & 80,103 & 2.16 & 20.91 \\
 &  & Test & 50,635 & 1.85 & 23.29 \\

\cmidrule{1-6}

\multirow[c]{3}{*}{\avito} & \multirow[c]{3}{*}{\userAdVisit} & 
Train & 2,738,733 & 31.53 & - \\
 &  & Val & 877,441 & 29.27 & 6.79 \\
 &  & Test & 712,985 & 19.73 & 4.73 \\
    \bottomrule
    \end{tabular}
    \label{tab:regresion analysis expt 2}
\end{table}

\begin{table}[th]
    \vspace{-3mm}
    \caption{Entity classification results (AUROC, higher is better) on selected \relbench regression tasks, converted into classification tasks with binary label $y = \mathbf{1}\{y_\text{regression} > 0\}$. Training a LightGBM model on features extracted by a trained GNN leads to performance lift. This is evidence that the linear layer output head of the base GNN is suboptimal.}
    \centering
    \setlength{\tabcolsep}{3pt}
    \scriptsize
    \newcommand{\tabledata}{\fone & \driverPosition & Test & $81.96_{\pm 1.18}$ & $\bm{86.63}_{\pm 0.40}$ \\
\cmidrule{1-5}
\stackex  & \postVotes & Test & $\bm{80.5}_{\pm 0.18}$ & $78.3_{\pm 0.05}$ \\
\cmidrule{1-5}
\amazon & \itemLtv & Test & $\bm{70.61}_{\pm 0.06}$ & $70.29_{\pm 0.06}$ \\

}
    \begin{tabular}{llrrr}
    \toprule
    \textbf{Dataset} & \textbf{Task} & \textbf{Split} & \textbf{GNN} & \textbf{Data Scientist} \\
    \midrule
    \multirow[c]{9}{*}{\amazon} & \multirow[c]{3}{*}{\userItemPurchase} 
 & Train & 11,759,844 & 2.18 & - \\
 &  & Val & 802,540 & 2.28 & 0.18 \\
 &  & Test & 918,919 & 2.33 & 0.15 \\

\cmidrule{2-6}

 & \multirow[c]{3}{*}{\userItemRate} 
 & Train & 7,146,115 & 1.8 & - \\
 &  & Val & 519,496 & 2.01 & 0.19 \\
 &  & Test & 599,867 & 2.05 & 0.15 \\

 \cmidrule{2-6}

 & \multirow[c]{3}{*}{\userItemReview} 
 & Train & 5,138,184 & 2.19 & - \\
 &  & Val & 268,651 & 2.3 & 0.18 \\
 &  & Test & 305,476 & 2.4 & 0.15 \\

\cmidrule{1-6}

\multirow[c]{3}{*}{\handm} & \multirow[c]{3}{*}{\userItemPurchase} & 
Train & 13,191,321 & 3.38 & - \\
 &  & Val & 237,152 & 3.18 & 3.51 \\
 &  & Test & 207,996 & 3.10 & 3.76 \\

% \cmidrule{1-6}
% \multirow[c]{3}{*}{\fone} & \multirow[c]{3}{*}{\driverConstructorResult}
% & Train & 1900 &  & - \\
%  &  & Val & 142 & 1.95 & 9.86 \\
%  &  & Test & 105 & 1.84 & 10.48 \\

\cmidrule{1-6}

\multirow[c]{6}{*}{\stackex} & \multirow[c]{3}{*}{\userPostComment}
 & Train & 43,337 & 2.08 & - \\
 & & Val & 1,603 & 1.94 & 3.43 \\
 &  & Test & 1,517 & 2.0 & 4.09 \\

\cmidrule{2-6}

 & \multirow[c]{3}{*}{\postPostLinked}
 & Train & 7,162 & 1.2 & - \\
 &  & Val & 294 & 1.3 & 0.0 \\
 &  & Test & 359 & 1.39 & 1.39 \\

\cmidrule{1-6}

\multirow[c]{6}{*}{\trials} & \multirow[c]{3}{*}{\sponsorConditionRec}
 & Train & 503,176 & 12.51 & - \\
 &  & Val & 30,448 & 14.63 & 34.48 \\
 &  & Test & 25,694 & 12.49 & 38.37 \\

\cmidrule{2-6}

& \multirow[c]{3}{*}{\sponsorFacilityRec}
 & Train & 1,485,360 & 2.27 & - \\
 &  & Val & 80,103 & 2.16 & 20.91 \\
 &  & Test & 50,635 & 1.85 & 23.29 \\

\cmidrule{1-6}

\multirow[c]{3}{*}{\avito} & \multirow[c]{3}{*}{\userAdVisit} & 
Train & 2,738,733 & 31.53 & - \\
 &  & Val & 877,441 & 29.27 & 6.79 \\
 &  & Test & 712,985 & 19.73 & 4.73 \\
    \bottomrule
    \end{tabular}
    \label{tab:regresion analysis expt 3}
\end{table}

\section{Dataset Origins and Licenes}\label{app:data}

%\subsection{Origins and Licenses}
This section details the sources for all data used in \relbench. In all cases, the data providers consent for their data to be used freely for non-commercial and research purposes. The only database with potentially personally identifiable information is \stackex, which draws from the Stack Exchange site, which sometimes has individuals' names as their username.  This information shared with consent, as all users must agree to the Stack Exchange privacy policy, see: \url{https://stackoverflow.com/legal/privacy-policy}.

\xhdr{\amazon} Data obtained from the Amazon Review Data Dump from \cite{ni2019justifying}. See the website: \url{https://cseweb.ucsd.edu/~jmcauley/datasets/amazon_v2/}. Data license is not specified.

\xhdr{\avito} Data is obtained from Kaggle \url{https://www.kaggle.com/competitions/avito-context-ad-clicks}. All \relbench users must download data from Kaggle themselves, a part of which is accepting the data usage terms. These terms include use only for non-commercial and academic purposes. Note that after data download, we further downsample the avito dataset by randomly selecting approximately 100,000 data point from user table and sample all other tables that have connections to the sampled users.

\xhdr{\stackex} Data was obtained from The Internet Archive, whose stated mission is to provide ``universal access to all knowledge. We downloaded our data from \url{https://archive.org/download/stackexchange} in Novermber 2023. Data license is not specified.

\xhdr{\fone} Data was sourced from the Ergast API (\url{https://ergast.com/mrd/}) in February 2024. The Ergast Developer API is an experimental web service which provides a historical record of motor racing data for non-commercial purposes. As far as we are able to determine the data is public and license is not specified.

\xhdr{\trials} Data was downloaded from the \url{ClinicalTrials.gov} website in January 2024. This data is provided by the NIH, an official branch of the US Government. The terms of use state that data are available to all requesters, both within and outside the United States, at no charge. Our \trials database is a snapshot from January 2024, and will not be updated with newer trials results.

\xhdr{\handm} Data is obtained from Kaggle \url{https://www.kaggle.com/competitions/h-and-m-personalized-fashion-recommendations}. All \relbench users must download data from Kaggle themselves, a part of which is accepting the data usage terms. These terms include use only for non-commercial and academic purposes.

\xhdr{\event} The dataset employed in this research was initially released on Kaggle for the Event Recommendation Engine Challenge, which can be accessed at \url{https://www.kaggle.com/c/event-recommendation-engine-challenge/data}. We have obtained explicit consent from the creators of this dataset to use it within \relbench. We extend our sincere gratitude to Allan Carroll for his support and generosity in sharing the data with the academic community.

\section{Additional Training Table Statistics}\label{app: train table stats}

We report additional training table statistics for all tasks, separated into entity classification (\emph{cf.}~Table~\ref{tab:node_classification_stats}), entity regression (\emph{cf.}~Table~\ref{tab:node_regression_stats}), and link prediction (\emph{cf.}~Table~\ref{tab:link_prediction_stats}).

\begin{table}[H]
    \caption{\relbench entity classification training table target statistics.}
    \centering
    \setlength{\tabcolsep}{3pt}
    \scriptsize
    \newcommand{\tabledata}{\multirow[c]{6}{*}{\amazon} & \multirow[c]{3}{*}{\userChurn} & Train & 2,956,658 (62.47\%) & 1,775,897 (37.53\%) \\
 &  & Val & 263,098 (64.2\%) & 146,694 (35.8\%) \\
 &  & Test & 213,400 (60.64\%) & 138,485 (39.36\%) \\
\cmidrule{2-5}
 & \multirow[c]{3}{*}{\itemChurn} & Train & 1,113,863 (43.52\%) & 1,445,401 (56.48\%) \\
 &  & Val & 73,242 (41.22\%) & 104,447 (58.78\%) \\
 &  & Test & 61,647 (36.95\%) & 105,195 (63.05\%) \\
\cmidrule{1-5}

\multirow[c]{6}{*}{\fone} & \multirow[c]{3}{*}{\driverDNF} & Train & 1,365 (11.96\%) & 10,046 (88.04\%) \\
 &  & Val & 125 (22.08\%) & 441 (77.92\%) \\
 &  & Test & 207 (29.49\%) & 495 (70.51\%) \\
\cmidrule{2-5}
 & \multirow[c]{3}{*}{\driverTopThree} & Train & 231 (17.07\%) & 1,122 (82.93\%) \\
 &  & Val & 119 (20.24\%) & 469 (79.76\%) \\
 &  & Test & 128 (17.63\%) & 598 (82.37\%) \\

\cmidrule{1-5}
\multirow[c]{3}{*}{\handm} & \multirow[c]{3}{*}{\userChurn} & 
Train & 3,170,367 (81.89\%) & 701,043 (18.11\%) \\
 &  & Val & 62,225 (81.28\%) & 14,331 (18.72\%) \\
 &  & Test & 61,609 (82.61\%) & 12,966 (17.39\%) \\
\cmidrule{1-5}
\multirow[c]{6}{*}{\stackex} & \multirow[c]{3}{*}{\userEngage}
 & Train & 68,020 (5.0\%) & 1,292,830 (95.0\%) \\
 & & Val & 2,411 (2.81\%) & 83,427 (97.19\%) \\
 &  & Test & 2,411 (2.74\%) & 85,726 (97.26\%) \\
\cmidrule{2-5}
 & \multirow[c]{3}{*}{\userBadge} & Train & 163,048 (4.81\%) & 3,223,228 (95.19\%) \\
 &  & Val & 7,301 (2.95\%) & 240,097 (97.05\%) \\
 &  & Test & 6,735 (2.64\%) & 248,625 (97.36\%) \\
\cmidrule{1-5}
\multirow[c]{3}{*}{\trials} & \multirow[c]{3}{*}{\studyOutcome}
 & Train & 7,647 (63.76\%) & 4,347 (36.24\%) \\
 &  & Val & 561 (58.44\%) & 399 (41.56\%) \\
 &  & Test & 483 (58.55\%) & 342 (41.45\%) \\
 \cmidrule{1-5}
 \multirow[c]{6}{*}{\event} & \multirow[c]{3}{*}{\userRepeat}
 & Train & 1,882 (48.98\%) & 1,960 (51.02\%) \\
 & & Val & 130 (48.51\%) & 138 (51.49\%) \\
 &  & Test & 110 (44.72\%) & 136 (55.28\%) \\
\cmidrule{2-5}
 & \multirow[c]{3}{*}{\userIgnore} & Train & 3,247 (16.88\%) & 15,992 (83.12\%) \\
 &  & Val & 441 (10.54\%) & 3,744 (89.46\%) \\
 &  & Test & 450 (11.40\%) & 3,499 (88.60\%) \\
 \cmidrule{1-5}
 \multirow[c]{6}{*}{\avito} & \multirow[c]{3}{*}{\userClick}
 & Train & 2,302 (3.87\%) & 57,152 (96.13\%) \\
 & & Val & 745 (3.52\%) & 20,438 (96.48\%) \\
 &  & Test & 740 (1.54\%) & 47,256 (98.46\%) \\
\cmidrule{2-5}
 & \multirow[c]{3}{*}{\userVisit} & Train & 78,467 (90.59\%) & 8,152 (9.41\%) \\
 &  & Val & 27,086 (90.35\%) & 2,893 (9.65\%) \\
 &  & Test & 30,731 (85.06\%) & 5,398 (14.94\%) \\}
    \begin{tabular}{llrrr}
    \toprule
    \textbf{Dataset} & \textbf{Task} & \textbf{Split} & \textbf{Positives} & \textbf{Negatives} \\
    \midrule
    \multirow[c]{9}{*}{\amazon} & \multirow[c]{3}{*}{\userItemPurchase} 
 & Train & 11,759,844 & 2.18 & - \\
 &  & Val & 802,540 & 2.28 & 0.18 \\
 &  & Test & 918,919 & 2.33 & 0.15 \\

\cmidrule{2-6}

 & \multirow[c]{3}{*}{\userItemRate} 
 & Train & 7,146,115 & 1.8 & - \\
 &  & Val & 519,496 & 2.01 & 0.19 \\
 &  & Test & 599,867 & 2.05 & 0.15 \\

 \cmidrule{2-6}

 & \multirow[c]{3}{*}{\userItemReview} 
 & Train & 5,138,184 & 2.19 & - \\
 &  & Val & 268,651 & 2.3 & 0.18 \\
 &  & Test & 305,476 & 2.4 & 0.15 \\

\cmidrule{1-6}

\multirow[c]{3}{*}{\handm} & \multirow[c]{3}{*}{\userItemPurchase} & 
Train & 13,191,321 & 3.38 & - \\
 &  & Val & 237,152 & 3.18 & 3.51 \\
 &  & Test & 207,996 & 3.10 & 3.76 \\

% \cmidrule{1-6}
% \multirow[c]{3}{*}{\fone} & \multirow[c]{3}{*}{\driverConstructorResult}
% & Train & 1900 &  & - \\
%  &  & Val & 142 & 1.95 & 9.86 \\
%  &  & Test & 105 & 1.84 & 10.48 \\

\cmidrule{1-6}

\multirow[c]{6}{*}{\stackex} & \multirow[c]{3}{*}{\userPostComment}
 & Train & 43,337 & 2.08 & - \\
 & & Val & 1,603 & 1.94 & 3.43 \\
 &  & Test & 1,517 & 2.0 & 4.09 \\

\cmidrule{2-6}

 & \multirow[c]{3}{*}{\postPostLinked}
 & Train & 7,162 & 1.2 & - \\
 &  & Val & 294 & 1.3 & 0.0 \\
 &  & Test & 359 & 1.39 & 1.39 \\

\cmidrule{1-6}

\multirow[c]{6}{*}{\trials} & \multirow[c]{3}{*}{\sponsorConditionRec}
 & Train & 503,176 & 12.51 & - \\
 &  & Val & 30,448 & 14.63 & 34.48 \\
 &  & Test & 25,694 & 12.49 & 38.37 \\

\cmidrule{2-6}

& \multirow[c]{3}{*}{\sponsorFacilityRec}
 & Train & 1,485,360 & 2.27 & - \\
 &  & Val & 80,103 & 2.16 & 20.91 \\
 &  & Test & 50,635 & 1.85 & 23.29 \\

\cmidrule{1-6}

\multirow[c]{3}{*}{\avito} & \multirow[c]{3}{*}{\userAdVisit} & 
Train & 2,738,733 & 31.53 & - \\
 &  & Val & 877,441 & 29.27 & 6.79 \\
 &  & Test & 712,985 & 19.73 & 4.73 \\
    \bottomrule
    \end{tabular}
    \label{tab:node_classification_stats}
\end{table}

\begin{table}[H] 
    \caption{\relbench entity regression training table target statistics.}
    \centering
    \setlength{\tabcolsep}{3pt}
    \scriptsize
    \newcommand{\tabledata}{\multirow[c]{8}{*}{\amazon} & \multirow[c]{4}{*}{\userLtv} & Train & 0.0 & 0.0 & 16.93 & 9,511.46 \\
 &  & Val & 0.0 & 0.0 & 14.14 & 7,259.91 \\
 &  & Test & 0.0 & 0.0 & 16.78 & 10,329.86 \\
 &  & Total & 0.0 & 0.0 & 16.71 & 10,329.86 \\
\cmidrule{2-7}
 & \multirow[c]{4}{*}{\itemLtv} & Train & 0.0 & 20.78 & 67.57 & 198,419.8 \\
 &  & Val & 0.0 & 22.44 & 72.10 & 75,901.55 \\
 &  & Test & 0.0 & 23.72 & 77.13 & 206,663.58 \\
 &  & Total & 0.0 & 20.97 & 68.38 & 206,663.58 \\
\cmidrule{1-7}

\multirow[c]{4}{*}{\fone} & \multirow[c]{4}{*}{\driverPosition} & Train & 1.0 & 13.33 & 13.90 & 39.0 \\
 &  & Val & 1.0 & 11.4 & 11.08 & 22.0 \\
 &  & Test & 1.0 & 12.18 & 11.93 & 24.0 \\
 &  & Total & 1.0 & 13.0 & 13.57 & 39.0 \\

\cmidrule{1-7}

\multirow[c]{4}{*}{\handm} & \multirow[c]{4}{*}{\itemSales} & 
    Train & 0.0 & 0.0 & 0.076 & 87.16 \\
 &  & Val & 0.0 & 0.0 & 0.086 & 40.36 \\
 &  & Test & 0.0 & 0.0 & 0.076 & 38.31 \\
 &  & Total & 0.0 & 0.0 & 0.076 & 87.16 \\

\cmidrule{1-7}

\multirow[c]{4}{*}{\stackex} & \multirow[c]{4}{*}{\postVotes}
 & Train & 0.0 & 0.0 & 0.093 & 78.0 \\
 & & Val & 0.0 & 0.0 & 0.062 & 36.0 \\
 & & Test & 0.0 & 0.0 & 0.068 & 26.0 \\
 & & Total & 0.0 & 0.0 & 0.090 & 78.0 \\

\cmidrule{1-7}

\multirow[c]{8}{*}{\trials} & \multirow[c]{4}{*}{\studyAdverse}
 & Train & 0.0 & 2.0 & 39.84 & 28,085.0 \\
 &  & Val & 0.0 & 2.0 & 57.08 & 17,245.0 \\
 &  & Test & 0.0 & 3.0 & 57.93 & 5,978.0 \\
 &  & Total & 0.0 & 2.0 & 42.20 & 28,085.0 \\

\cmidrule{2-7}

 & \multirow[c]{4}{*}{\facilitySuccess} 
  & Train & 0.0 & 0.0 & 0.44 & 1.0 \\
 &  & Val & 0.0 & 0.4 & 0.47 & 1.0 \\
 &  & Test & 0.0 & 0.17 & 0.4 & 1.06 \\
 &  & Total & 0.0 & 0.0 & 0.45 & 1.0 \\

\cmidrule{1-7}

\multirow[c]{4}{*}{\event} & \multirow[c]{4}{*}{\userAttendance} & Train & 0.0 & 0.0 & 0.37 & 16.0 \\
 &  & Val & 0.0 & 0.0 & 0.28 & 5.0 \\
 &  & Test & 0.0 & 0.0 & 0.26 & 8.0 \\
 &  & Total & 0.0 & 0.0 & 0.34 & 16.0 \\

\cmidrule{1-7}

\multirow[c]{4}{*}{\avito} & \multirow[c]{4}{*}{\adsCTR} & 
   Train & 0.00052 & 0.018 & 0.045 & 1.0\\
 &  & Val & 0.00091 & 0.018 & 0.048 & 1.0 \\
 &  & Test & 0.00085 & 0.019 & 0.052 & 1.0\\
 &  & Total & 0.00052 & 0.018 & 0.047 & 1.0\\}
    \begin{tabular}{llrrrrr}
    \toprule
    \textbf{Dataset} & \textbf{Task} & \textbf{Split} & \textbf{Minimum} & \textbf{Median} & \textbf{Mean} & \textbf{Maximum} \\
    \midrule
    \multirow[c]{9}{*}{\amazon} & \multirow[c]{3}{*}{\userItemPurchase} 
 & Train & 11,759,844 & 2.18 & - \\
 &  & Val & 802,540 & 2.28 & 0.18 \\
 &  & Test & 918,919 & 2.33 & 0.15 \\

\cmidrule{2-6}

 & \multirow[c]{3}{*}{\userItemRate} 
 & Train & 7,146,115 & 1.8 & - \\
 &  & Val & 519,496 & 2.01 & 0.19 \\
 &  & Test & 599,867 & 2.05 & 0.15 \\

 \cmidrule{2-6}

 & \multirow[c]{3}{*}{\userItemReview} 
 & Train & 5,138,184 & 2.19 & - \\
 &  & Val & 268,651 & 2.3 & 0.18 \\
 &  & Test & 305,476 & 2.4 & 0.15 \\

\cmidrule{1-6}

\multirow[c]{3}{*}{\handm} & \multirow[c]{3}{*}{\userItemPurchase} & 
Train & 13,191,321 & 3.38 & - \\
 &  & Val & 237,152 & 3.18 & 3.51 \\
 &  & Test & 207,996 & 3.10 & 3.76 \\

% \cmidrule{1-6}
% \multirow[c]{3}{*}{\fone} & \multirow[c]{3}{*}{\driverConstructorResult}
% & Train & 1900 &  & - \\
%  &  & Val & 142 & 1.95 & 9.86 \\
%  &  & Test & 105 & 1.84 & 10.48 \\

\cmidrule{1-6}

\multirow[c]{6}{*}{\stackex} & \multirow[c]{3}{*}{\userPostComment}
 & Train & 43,337 & 2.08 & - \\
 & & Val & 1,603 & 1.94 & 3.43 \\
 &  & Test & 1,517 & 2.0 & 4.09 \\

\cmidrule{2-6}

 & \multirow[c]{3}{*}{\postPostLinked}
 & Train & 7,162 & 1.2 & - \\
 &  & Val & 294 & 1.3 & 0.0 \\
 &  & Test & 359 & 1.39 & 1.39 \\

\cmidrule{1-6}

\multirow[c]{6}{*}{\trials} & \multirow[c]{3}{*}{\sponsorConditionRec}
 & Train & 503,176 & 12.51 & - \\
 &  & Val & 30,448 & 14.63 & 34.48 \\
 &  & Test & 25,694 & 12.49 & 38.37 \\

\cmidrule{2-6}

& \multirow[c]{3}{*}{\sponsorFacilityRec}
 & Train & 1,485,360 & 2.27 & - \\
 &  & Val & 80,103 & 2.16 & 20.91 \\
 &  & Test & 50,635 & 1.85 & 23.29 \\

\cmidrule{1-6}

\multirow[c]{3}{*}{\avito} & \multirow[c]{3}{*}{\userAdVisit} & 
Train & 2,738,733 & 31.53 & - \\
 &  & Val & 877,441 & 29.27 & 6.79 \\
 &  & Test & 712,985 & 19.73 & 4.73 \\
    \bottomrule
    \end{tabular}
    \label{tab:node_regression_stats}
\end{table}

\begin{table}[H]
    \caption{\relbench link prediction training table link statistics.}
    \centering
    \setlength{\tabcolsep}{3pt}
    \scriptsize
    \newcommand{\tabledata}{\multirow[c]{9}{*}{\amazon} & \multirow[c]{3}{*}{\userItemPurchase} 
 & Train & 11,759,844 & 2.18 & - \\
 &  & Val & 802,540 & 2.28 & 0.18 \\
 &  & Test & 918,919 & 2.33 & 0.15 \\

\cmidrule{2-6}

 & \multirow[c]{3}{*}{\userItemRate} 
 & Train & 7,146,115 & 1.8 & - \\
 &  & Val & 519,496 & 2.01 & 0.19 \\
 &  & Test & 599,867 & 2.05 & 0.15 \\

 \cmidrule{2-6}

 & \multirow[c]{3}{*}{\userItemReview} 
 & Train & 5,138,184 & 2.19 & - \\
 &  & Val & 268,651 & 2.3 & 0.18 \\
 &  & Test & 305,476 & 2.4 & 0.15 \\

\cmidrule{1-6}

\multirow[c]{3}{*}{\handm} & \multirow[c]{3}{*}{\userItemPurchase} & 
Train & 13,191,321 & 3.38 & - \\
 &  & Val & 237,152 & 3.18 & 3.51 \\
 &  & Test & 207,996 & 3.10 & 3.76 \\

% \cmidrule{1-6}
% \multirow[c]{3}{*}{\fone} & \multirow[c]{3}{*}{\driverConstructorResult}
% & Train & 1900 &  & - \\
%  &  & Val & 142 & 1.95 & 9.86 \\
%  &  & Test & 105 & 1.84 & 10.48 \\

\cmidrule{1-6}

\multirow[c]{6}{*}{\stackex} & \multirow[c]{3}{*}{\userPostComment}
 & Train & 43,337 & 2.08 & - \\
 & & Val & 1,603 & 1.94 & 3.43 \\
 &  & Test & 1,517 & 2.0 & 4.09 \\

\cmidrule{2-6}

 & \multirow[c]{3}{*}{\postPostLinked}
 & Train & 7,162 & 1.2 & - \\
 &  & Val & 294 & 1.3 & 0.0 \\
 &  & Test & 359 & 1.39 & 1.39 \\

\cmidrule{1-6}

\multirow[c]{6}{*}{\trials} & \multirow[c]{3}{*}{\sponsorConditionRec}
 & Train & 503,176 & 12.51 & - \\
 &  & Val & 30,448 & 14.63 & 34.48 \\
 &  & Test & 25,694 & 12.49 & 38.37 \\

\cmidrule{2-6}

& \multirow[c]{3}{*}{\sponsorFacilityRec}
 & Train & 1,485,360 & 2.27 & - \\
 &  & Val & 80,103 & 2.16 & 20.91 \\
 &  & Test & 50,635 & 1.85 & 23.29 \\

\cmidrule{1-6}

\multirow[c]{3}{*}{\avito} & \multirow[c]{3}{*}{\userAdVisit} & 
Train & 2,738,733 & 31.53 & - \\
 &  & Val & 877,441 & 29.27 & 6.79 \\
 &  & Test & 712,985 & 19.73 & 4.73 \\}
    \begin{tabular}{llrrrr}
    \toprule
    \textbf{Dataset} & \textbf{Task} & \textbf{Split} & \textbf{\#Links} & \makecell{\textbf{Avg \#links per}\\\textbf{entity/timestamp}} & \textbf{\%Repeated links} \\
    \midrule
    \multirow[c]{9}{*}{\amazon} & \multirow[c]{3}{*}{\userItemPurchase} 
 & Train & 11,759,844 & 2.18 & - \\
 &  & Val & 802,540 & 2.28 & 0.18 \\
 &  & Test & 918,919 & 2.33 & 0.15 \\

\cmidrule{2-6}

 & \multirow[c]{3}{*}{\userItemRate} 
 & Train & 7,146,115 & 1.8 & - \\
 &  & Val & 519,496 & 2.01 & 0.19 \\
 &  & Test & 599,867 & 2.05 & 0.15 \\

 \cmidrule{2-6}

 & \multirow[c]{3}{*}{\userItemReview} 
 & Train & 5,138,184 & 2.19 & - \\
 &  & Val & 268,651 & 2.3 & 0.18 \\
 &  & Test & 305,476 & 2.4 & 0.15 \\

\cmidrule{1-6}

\multirow[c]{3}{*}{\handm} & \multirow[c]{3}{*}{\userItemPurchase} & 
Train & 13,191,321 & 3.38 & - \\
 &  & Val & 237,152 & 3.18 & 3.51 \\
 &  & Test & 207,996 & 3.10 & 3.76 \\

% \cmidrule{1-6}
% \multirow[c]{3}{*}{\fone} & \multirow[c]{3}{*}{\driverConstructorResult}
% & Train & 1900 &  & - \\
%  &  & Val & 142 & 1.95 & 9.86 \\
%  &  & Test & 105 & 1.84 & 10.48 \\

\cmidrule{1-6}

\multirow[c]{6}{*}{\stackex} & \multirow[c]{3}{*}{\userPostComment}
 & Train & 43,337 & 2.08 & - \\
 & & Val & 1,603 & 1.94 & 3.43 \\
 &  & Test & 1,517 & 2.0 & 4.09 \\

\cmidrule{2-6}

 & \multirow[c]{3}{*}{\postPostLinked}
 & Train & 7,162 & 1.2 & - \\
 &  & Val & 294 & 1.3 & 0.0 \\
 &  & Test & 359 & 1.39 & 1.39 \\

\cmidrule{1-6}

\multirow[c]{6}{*}{\trials} & \multirow[c]{3}{*}{\sponsorConditionRec}
 & Train & 503,176 & 12.51 & - \\
 &  & Val & 30,448 & 14.63 & 34.48 \\
 &  & Test & 25,694 & 12.49 & 38.37 \\

\cmidrule{2-6}

& \multirow[c]{3}{*}{\sponsorFacilityRec}
 & Train & 1,485,360 & 2.27 & - \\
 &  & Val & 80,103 & 2.16 & 20.91 \\
 &  & Test & 50,635 & 1.85 & 23.29 \\

\cmidrule{1-6}

\multirow[c]{3}{*}{\avito} & \multirow[c]{3}{*}{\userAdVisit} & 
Train & 2,738,733 & 31.53 & - \\
 &  & Val & 877,441 & 29.27 & 6.79 \\
 &  & Test & 712,985 & 19.73 & 4.73 \\
    \bottomrule
    \end{tabular}
    \label{tab:link_prediction_stats}
\end{table}

\section{Dataset Schema}

\begin{figure}[H]
    \centering
    \includegraphics[width=0.6\textwidth]{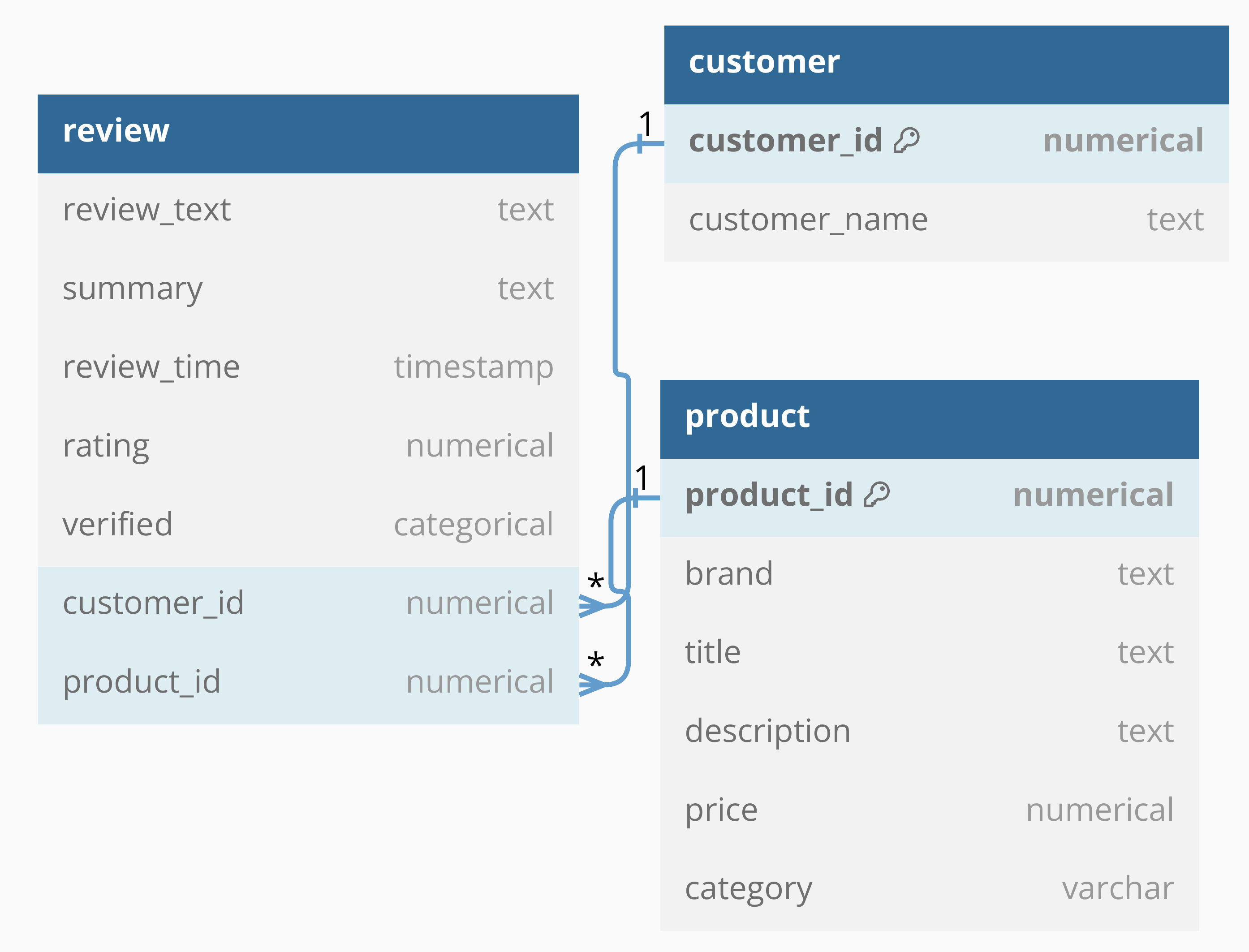}
    \caption{\amazon database diagram.}
    \label{fig:amazon-db}
\end{figure}

\begin{figure}[H]
    \centering
    \includegraphics[width=0.9\textwidth]{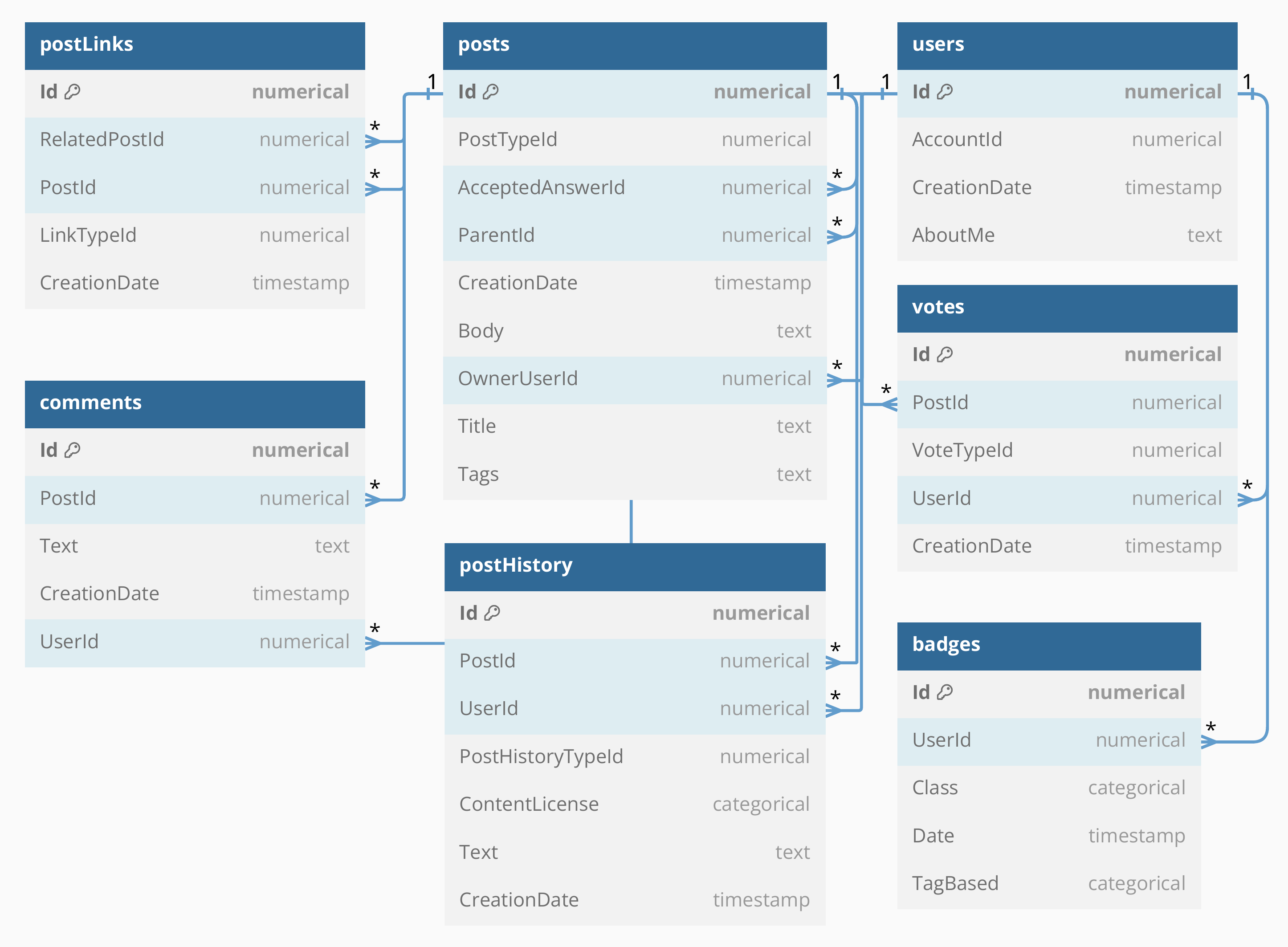}
    \caption{\stackex database diagram.}
    \label{fig:stack-db}
\end{figure}

\begin{figure}[H]
    \centering
    \includegraphics[width=0.9\textwidth]{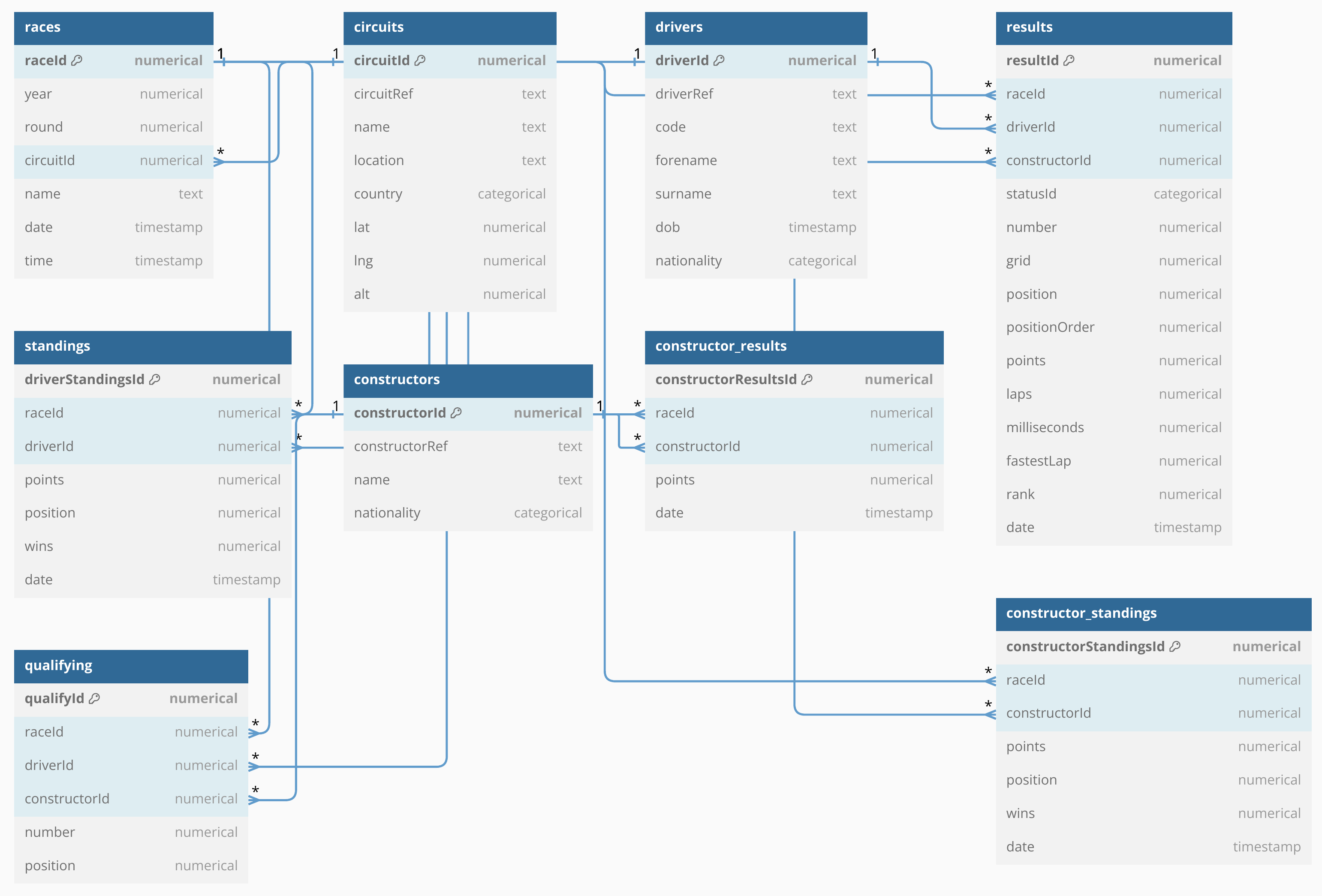}
    \caption{\fone database diagram.}
    \label{fig:f1-db}
\end{figure}

\begin{figure}[H]
    \centering
    \includegraphics[width=1\textwidth]{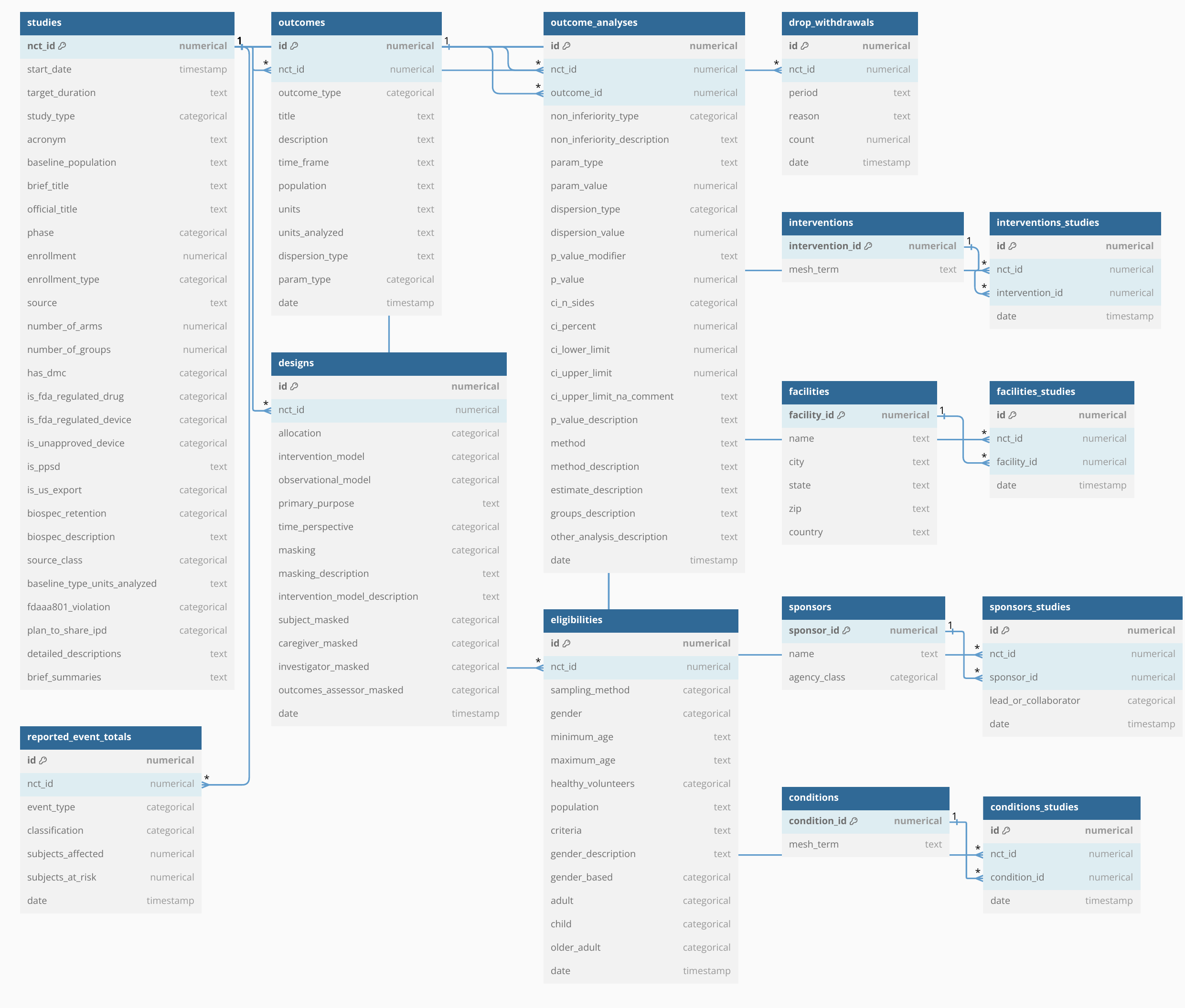}
    \caption{\trials database diagram.}
    \label{fig:trial-db}
\end{figure}

\begin{figure}[H]
    \centering
    \includegraphics[width=0.8\textwidth]{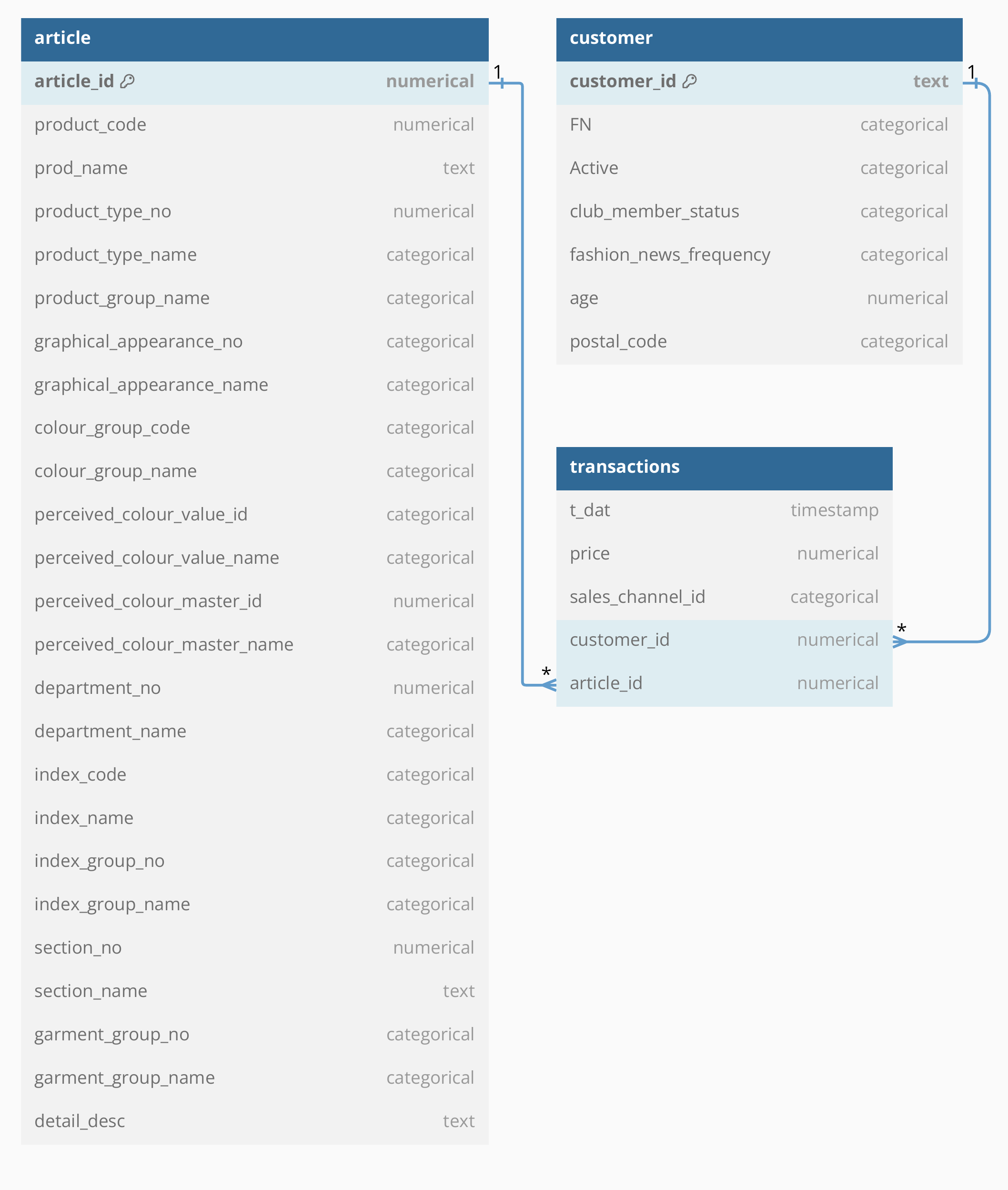}
    \caption{\handm database diagram.}
    \label{fig:hm-db}
\end{figure}

\begin{figure}[H]
    \centering
    \includegraphics[width=1\textwidth]{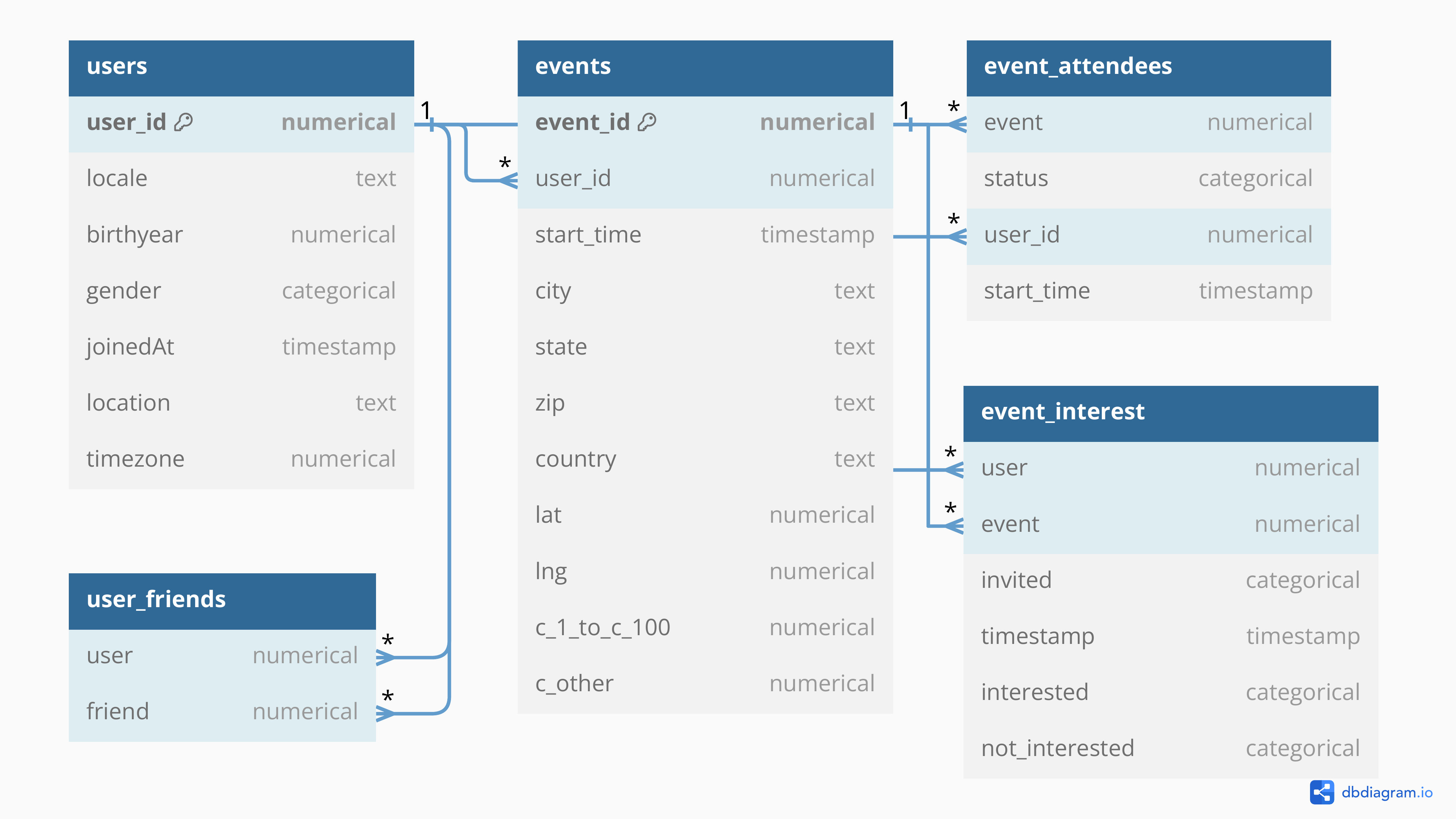}
    \caption{\event database diagram.}
    \label{fig:event-db}
\end{figure}

\begin{figure}[H]
    \centering
    \includegraphics[width=1\textwidth]{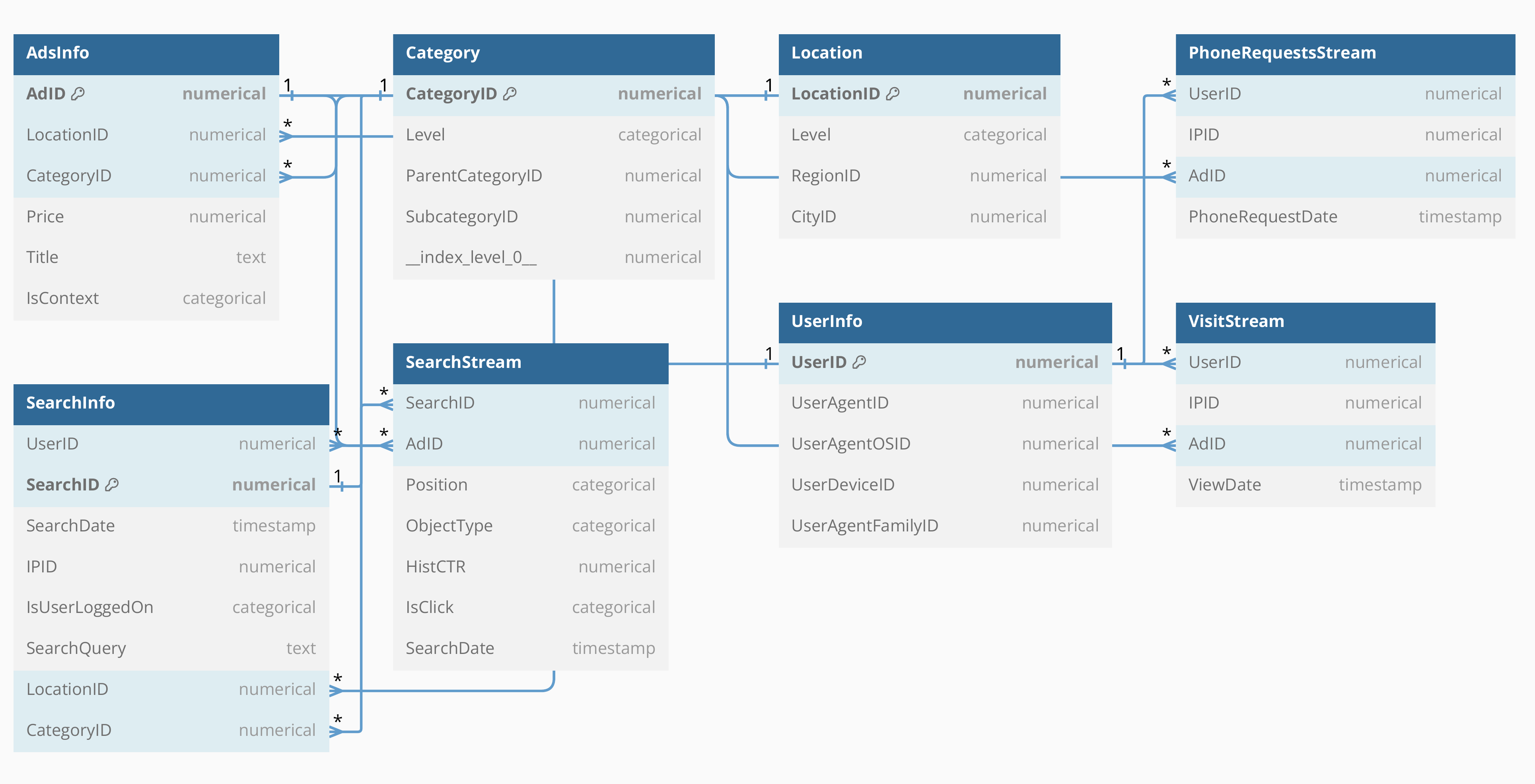}
    \caption{\avito database diagram.}
    \label{fig:avito-db}
\end{figure}
\section{Broader Impact}\label{app: broader impact}

Relational deep learning broadens the applicability of graph machine learning to include relational databases. Whilst the blueprint is general, and can be applied to a wide variety of tasks, including potentially hazardous ones, we have taken steps to focus attention of potential positive use cases. Specifically, the beta version of \relbench considers two databases, Amazon products, and Stack Exchange, that are designed to highlight the usefulness of RDL for driving online commerce and online social networks. Future releases of \relbench will continue to expand the range of databases into domains we reasonably expect to be positive, such as biomedical data and sports fixtures. We hope these concrete steps ensure the adoption of RDL for purposes broadly beneficial to society.

Whilst we strongly believe the \relbench has all the ingredients needed to be a long term benchamrk for relational deep learning, there are also possibilities for improvement and extension. Two such possibilities include: (1) RDL at scale: currently our implementation must load the entire database into working memory during training. For very large datasets this is not viable. Instead, a custom batch sampler is needed that acesses the database via queries to sample specific entities and their pkey-fkey neighbors; (2) Fully inductive link-prediction: our current link-prediction implementation supports predicting links for test time pairs (\emph{head,tail}) where \emph{head} is potentially new (unseen during training) and \emph{tail} seen in the training data. Extending this formulation to be fully inductive (\ie, \emph{tail} unseen during training) is possible, but out of the scope of this work for now. 
\end{document}